\documentclass{article}

\usepackage[T1]{fontenc}
\usepackage{grffile}

\usepackage{CJKutf8}

\usepackage{tikz-dependency}

\usepackage{hyperref}
\usepackage{longtable}
\usepackage{amsmath}
\usepackage{amsfonts}
\usepackage[numbers]{natbib}
\usepackage{tikz-dependency}

\usepackage{tipa}

\usepackage{url}

\usepackage{fullpage}

\usepackage[utf8]{inputenc}
\usepackage{graphicx}
\usepackage{xcolor}
\usepackage{multirow}

\usepackage[english]{babel}
\usepackage[utf8]{inputenc}

\newcommand{\soft}[1]{}
\newcommand{\nopreview}[1]{}

\title{Crosslinguistic word order variation reflects evolutionary pressures of dependency and information locality}

\date{}

\author{Michael Hahn$^{1,2}$, Yang Xu$^3$ \\ mhahn2@stanford.edu \\ $^1$ Department of Linguistics, Stanford University, \\ $^2$ SFB 1102, Saarland University,\\$^3$ Department of Computer Science, Cognitive Science Program, University of Toronto}

\begin{document}

\maketitle

\begin{abstract}
	Languages vary considerably in syntactic structure. About 40\% of the world's languages have subject-verb-object order, and about 40\% have subject-object-verb order. Extensive work has sought to explain this word order variation across languages.  However, the existing approaches are not able to explain coherently the frequency distribution and evolution of word order in individual languages.
	We propose that variation in word order reflects different ways of balancing competing pressures of dependency locality and information locality, whereby languages favor placing elements together when they are syntactically related or contextually informative about each other.
	Using data from 80 languages in 17 language families and phylogenetic modeling, we demonstrate that languages evolve to balance these pressures, such that
	word order change is accompanied by change in the frequency distribution of the syntactic structures which speakers communicate to maintain overall efficiency.
	Variability in word order thus reflects different ways in which languages resolve these evolutionary pressures.
	We identify relevant characteristics that result from this joint optimization, particularly the frequency with which subjects and objects are expressed together for the same verb.
	Our findings suggest that syntactic structure and usage across languages co-adapt to support efficient communication under limited cognitive resources.
\end{abstract}

\footnotetext{This is the preprint of the following perr-reviewed publication: Michael Hahn, Yang Xu (2022), \textit{Crosslinguistic word order variation reflects evolutionary pressures of dependency and information locality}, In Proceedings of the National Academy of Sciences of the United States of America, vol. 119(24). The final copyedited version is available at: \url{https://www.pnas.org/doi/10.1073/pnas.2122604119}.}

\begin{figure*}[ht!]
    \centering
    \begin{minipage}[t]{.8\textwidth}
	\vspace{0pt}\includegraphics[width=.99\textwidth]{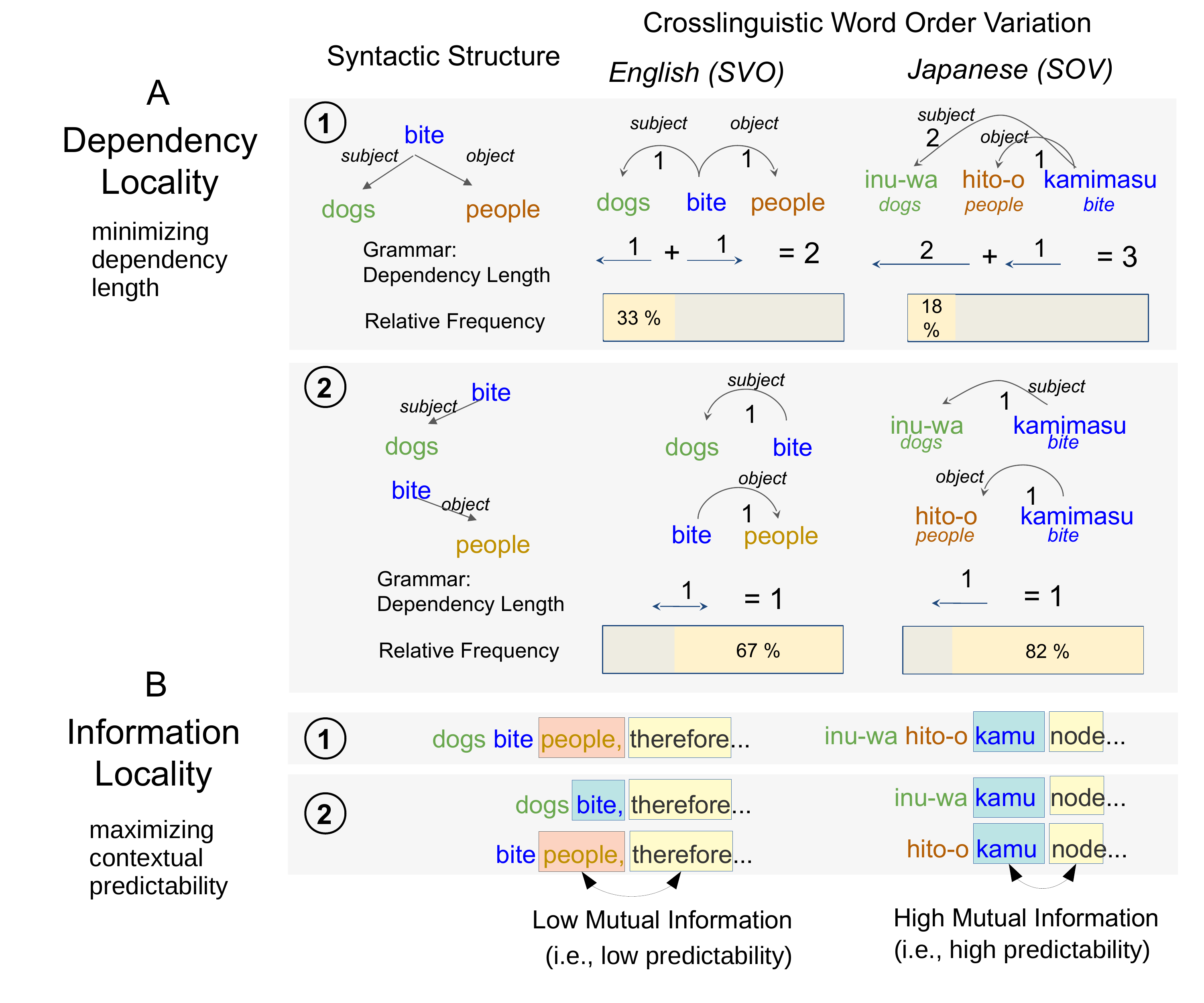}
	
	\end{minipage}
\caption{ 
Illustrations of crosslinguistic word order variation and the theoretical proposal.
A: English (SVO, center) and Japanese (SOV, right) linearize different syntactic structures into strings of words.
Numbers below arcs indicate dependency length.
Syntactic structures in (1) contain both a subject and an object; those in (2) only express one of the two.
The percentage bars show relative usage frequencies computed from large-scale English and Japanese text corpora (see SI Appendix, Figure S9).
In (1), English achieves shorter dependency length 2, compared to 3 in Japanese.
In (2), both languages achieve the same dependency length.
Therefore, structures as those in (2) are more favorable for dependency length minimization in SOV languages than those in (1).
Structures such as those in (2) are considerably more frequently expressed in Japanese.
B: SOV makes the beginning and end of clauses more predictable from the local context. For instance, in this example, the end of a clause is always a verb, increasing the mutual information with the subsequent word. Therefore, SOV tends to be more favorable for Information Locality.
}
        \label{fig:sent-dep-ab}
\end{figure*}

The world's languages show considerable variation in syntactic structure \citep{greenberg-universals-1963, baker2001atoms, croft2003typology}. A key syntactic dimension that languages vary along is word order. Linguists have long classified languages according to their basic word order, or the order in which they typically order verbs, subjects, and objects \citep{greenberg-universals-1963}.
About 40\% of the world's languages are classified as following subject-verb-object order ({SVO}, as in English, ``dogs bite people''), and 40\% are classified as following subject-object-verb order ({SOV}, as in Japanese, \textit{inu-wa hito-o kamimasu}, ``dogs people bite'') \citep{wals-81} (see Figure~\ref{fig:sent-dep-ab} for an illustration). Other orders such as verb-subject-object (VSO, as in Modern Standard Arabic, \textit{ta\textipa{Q}ad\textipa{\super Q}\textipa{:}u l-kila\textipa{:}bu n-na\textipa{:}sa} ``bite dogs people'') are much less common. Many languages have more than one ordering or exhibit historical change in their word order, although typically with one of the orderings being the most common. This ordering is considered the basic word order of a language in the typological literature \citep{greenberg-universals-1963, wals-s6}. Why do languages vary in word order the way they do, and what explains the evolution of word order? Here we present a unified theory that addresses these long-standing questions.

Different theories have been proposed for the word order variation across languages. The low frequency of object-initial orderings (e.g., OSV, ``human--dog--bites'') arguably has satisfactory explanations in terms of the meanings and functions typically associated with objects and subjects (e.g., \cite{tomlin1986basic}), but there is no consensus on the frequency distribution of SVO and SOV. Some work has argued that SOV is the default order in the history of language and that SVO has emerged later \citep{givon1979understanding,senghas1997argument, newmeyer2000evolutionary, goldin-meadow2008natural, gibson-noisy-channel-2013}, although phylogenetic simulation suggests that languages can cycle between these two orders in their historical development \citep{maurits2014tracing}. Other work suggests there is a tension between different cognitive pressures that favor SVO and SOV, respectively \citep{langus2010cognitive, ferrer-i-cancho-placement-2017}, but these accounts do not predict word order on the level of individual languages. So far, there exists no theory that coherently explains the principles underlying both crosslinguistic variation and evolution in word order.

One promising view suggests that crosslinguistic variation is constrained by the functional pressures of efficient communication under limited cognitive resources \citep{haspelmath2008parametric, jaeger2011on, kemp2018semantic, gibson2019how}.
Under this view, the structure of language in part reflects the way that language is used \citep{hopper1984the, bybee1994the, croft2000explaining, bybee2006from} and adapts to optimize informativeness and effort for human communication \citep{hawkins-performance-1994, haspelmath2006against, bybee2010language}.
Work in this paradigm has argued that many properties of language arise because they make language efficient for human language use and processing.
Several studies have shown in various domains that the grammars of human languages are more efficient than the vast majority of other logically possible grammars \cite{liu2008dependency, futrell-large-scale-2015, gildea-human-2015, Hahn2020modeling}.
Prior work has also shown that several of the known near-universal properties of languages hold in most logically possible languages that are also highly efficient, and used this fact to argue that evolution towards efficiency explains why these properties are near-universal \cite{kemp2012kinship,zaslavsky2018efficient, Mollicae2025993118,  hahn2020universals}.
However, existing efficiency-based approaches to grammatical typology leave open two key questions.
First, work in this paradigm focuses on cross-sectional studies, suggesting that languages are relatively efficient typically without answering how languages come to be efficient over time.
Second, the idea that languages are shaped by efficiency optimization does not directly explain why they vary in their grammatical structure, for instance, why both SVO and SOV are frequently attested across the world's languages.
In some domains, differences among languages have been interpreted as reflecting different optima or points along a Pareto frontier \citep{kemp2018semantic, zaslavsky2018efficient}.
However, it is currently unknown whether this perspective also applies to syntax and word order, and in particular to basic word order.

\section{Theoretical Proposal}

We address the open questions regarding word order variation and evolution using a large-scale
phylogenetic analysis of 80 languages from 17 families and contributing
the following theoretical view. We propose that crosslinguistic
variation in word order reflects a tendency for languages to trade off
competing pressures of communicative efficiency. This tendency is a product of an evolutionary process whereby different languages are functionally optimized in both their grammatical structure and the way they are used.
As such, grammar and usage should evolve together to jointly maintain efficiency, reflecting a process of coadaptation between grammar and usage.

Our proposal is grounded in prominent efficiency-based accounts of word order that are centered around various locality principles. These accounts assert that syntactic elements are ordered closer together when they are more strongly related in terms of their meaning and function \citep{behaghel1932deutsche,givon1985iconicity,rijkhoff-word-1986,hawkins-performance-1994,frazier1985syntactic}.
Recent work has specifically established two locality principles in word order typology. The first principle is \textbf{Dependency Locality} (DL), which is the observation that languages tend to order words to reduce the overall distance between syntactically related words \citep{rijkhoff-word-1986,hawkins-performance-1994,liu2008dependency,futrell-large-scale-2015, liu-dependency-2017}.
Dependency Locality can be justified in terms of parsing efficiency \citep{hawkins-performance-1994}, memory efficiency \citep{gibson-linguistic-1998} and general communicative efficiency \citep{hahn2020universals}.
The second locality principle is \textbf{Information Locality} (IL), which holds that words are close together if they contain predictive information about each other \citep{qian-cue-2012, futrell2020lossy, Culbertson2020FromTW, Hahn2020modeling}.
Information Locality is grounded in the well-established finding that words are hard to process for humans to the extent that they are hard to anticipate from preceding context \citep{hale2001probabilistic, levy-expectation-based-2008} under constraints of human memory \citep{futrell2020lossy}.
Both principles have individually received substantial support from crosslinguistic corpus studies  \citep{liu2008dependency,futrell-large-scale-2015,Hahn2020modeling}.
In a recent study, Gildea and Jaeger \cite{gildea-human-2015} even showed for five languages that they optimize a tradeoff of DL and an IL-like quantity, though this finding has not been replicated on a larger sample yet.




The two locality principles we described can make opposing predictions concerning basic word order (Figure~\ref{fig:sent-dep-ab}).
Dependency Locality should tend to favor SVO order over SOV \citep{ferrer-i-cancho-placement-2017}, because it ensures subject and object are both close to the verb.
In contrast, Information Locality may tend to favor SOV order over SVO, because uniform placement of S and O makes the beginning and end of each sentence easier to predict from local information.
We thus hypothesize that the crosslinguistic variation in basic word order emerges from an evolutionary process in which languages resolve the tension between these two pressures in different ways.

To illustrate our proposed theory, we consider a simple transitive sentence such as `dogs bite people' (Figure~\ref{fig:sent-dep-ab}A). 
Dependency Locality (DL) is defined formally in terms of Dependency Grammar~\citep{hays1964dependency,hudson1984word,melcuk1988dependency,corbett1993heads,tesniere2015elements}.
In this formalism, the syntactic structure of a sentence is drawn with directed arcs linking words -- called heads -- to those words that are syntactically subordinate to them -- called dependents.
For instance, arcs link the verb ``bite'' to its subject ``dogs'' and object ``people''.
The length of an arc is one plus the number of other words that it crosses.
The dependency length of an entire sentence is the sum of the lengths of all dependency arcs.

The grammars of languages specify how such syntactic structures are linearized into strings of words.
Figure~\ref{fig:sent-dep-ab} shows how the same syntactic structures are linearized differently by the grammars of English (SVO) and Japanese (SOV).
In a simple sentence as illustrated in Figure~\ref{fig:sent-dep-ab}A, SVO order results in overall lower dependency length than SOV (5 instead of 6).

The extent to which these general predictions are valid in a particular language will depend on the precise frequency at which speakers of a language use different syntactic structures.
For example, the difference between SVO and SOV order is neutralized for DL in sentences like (2) in Figure~\ref{fig:sent-dep-ab}A, where only a subject or an object is expressed.
Conversely, there are also syntactic structures where SOV achieves strictly shorter dependency lengths. 
Therefore, DL favors SVO more strongly when a language frequently co-expresses both subject and object of the same verb (Figure~\ref{fig:sent-dep-ab}).

Information Locality (IL) is defined in terms of the information-theoretic predictability of words from their recent prior context.
We adopt a simple formalization in terms of maximizing the mutual information between adjacent words; see SI Appendix, Section S1.1 for other operationalizations of IL.
SOV can be advantageous for IL, because a uniform verb-final ordering makes the beginning and end of each clause more predictable (Figure~\ref{fig:sent-dep-ab}B).


\begin{figure}[ht!]
    \centering
	\vspace{0pt}\includegraphics[width=.41\textwidth]{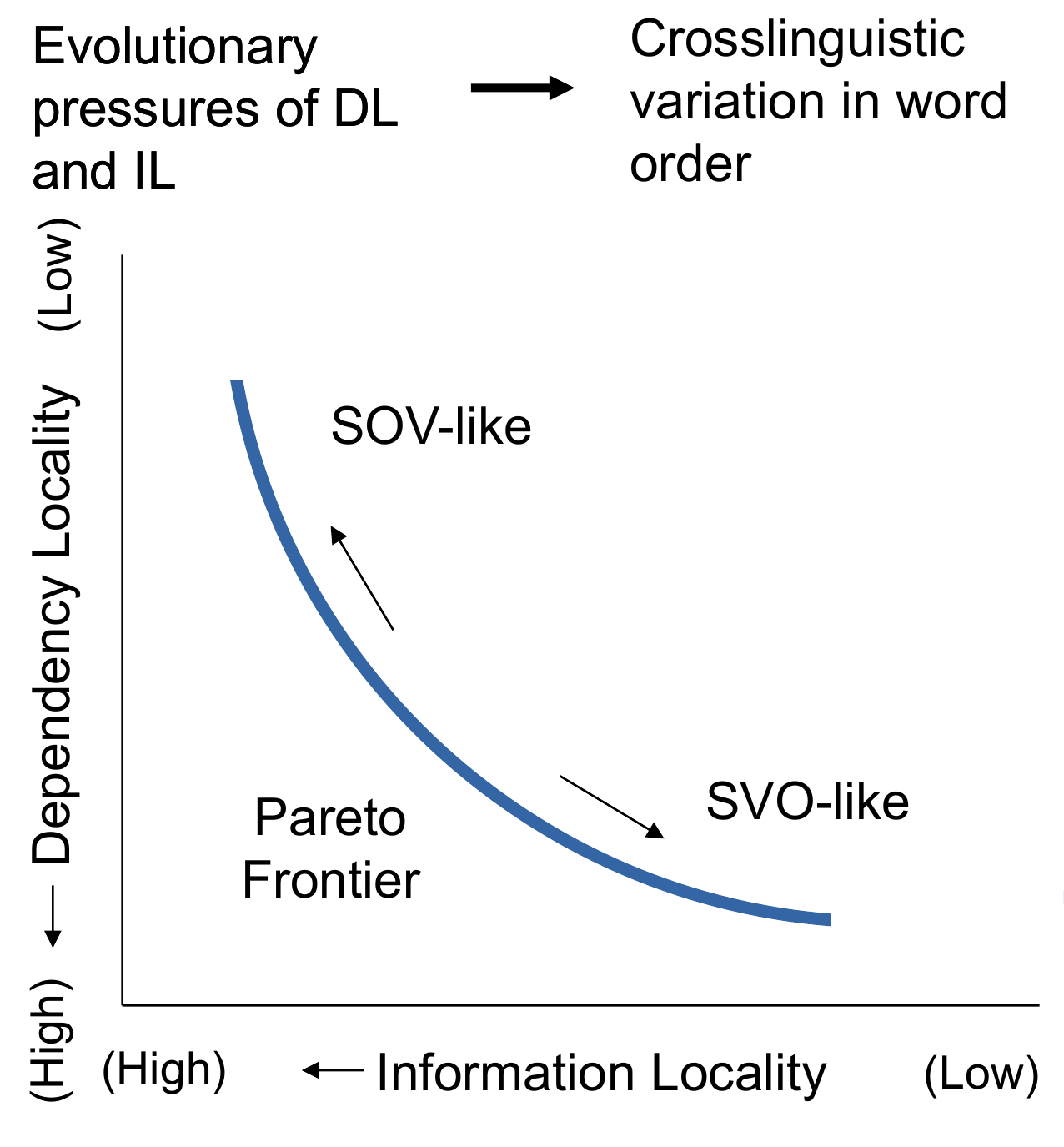}
\caption{
Illustration of out theoretical proposal:
Our theory postulates that word orders of languages trade off the competing evolutionary pressures of DL and IL. The space of logically possible word orders is bounded by a Pareto frontier of optimal orders. 
Along the Pareto frontier, SVO-like orders tend to optimize DL more strongly than SOV-like orders do.
The precise distribution of orderings along the frontier is determined by language-specific usage frequencies, which evolve together with word order in a process of coadaptation, so that the real order used in a language resembles the orderings more prevalent along the frontier.
}
        \label{fig:sent-dep-c}
\end{figure}

Which pressure will prevail in shaping a language depends not only on which pressure the language optimizes more strongly, but also on the frequency with which different syntactic structures are used.
The average values for Dependency Locality and Information Locality  achieved for a language depends on both the linearizations provided by grammars, and on the frequencies at which different syntactic structures are used.
Languages differ not only in the ways they linearize these syntactic structures, but also in the frequencies at which speakers utilize them.
For instance, syntactic structures such as those in (1) in Figure~\ref{fig:sent-dep-ab} are used at significantly higher frequencies by speakers of English than speakers of Japanese. Given a distribution over syntactic structures, we can identify which grammar orders them in such a way to achieve optimal average dependency length, optimal average information locality, or any linear combination of the two.
This gives rise to a Pareto frontier of grammars.

We summarize our proposal with the following two theoretical predictions (see Figure~\ref{fig:sent-dep-c}). First, languages evolve to maintain a relatively efficient balance between DL and IL, and second, this evolutionary process jointly affects usage frequencies (which syntactic structures are chosen by speakers) and grammar (how they are linearized).
This means that the actual word orders used by languages should be close to the most efficient possible grammar, given the distribution over syntactic structures.

\section{Results}
We first evaluate our proposed theory in a synchronic setting using data from 80 languages. We compute, for each language, which basic word orders are most efficient given its usage distribution as recorded in large-scale text corpora.
We then use a diachronic model of drift on phylogenetic trees to assess whether languages have evolved historically toward states where syntactic usage frequency and basic word order are aligned.

To begin with, we compared the efficiency of the attested orderings to both a null distribution of baseline grammars, and to the Pareto frontier of optimized grammars.
We represent grammars using the  established model of counterfactual order grammars introduced by Gildea and colleagues \cite{gildea-optimizing-2007}.
These are simple, parametric models specifying how the words in a syntactic structure are linearized depending on their syntactic relations.
They specify the orders not only of subjects and objects, but also of all other syntactic relations annotated in the syntactic structures, such as adpositions, adjectival modifiers, or relative clauses.
For instance, such a grammar may specify that subjects follow or precede verbs, that adpositions are pre- or postpositions, and that adjectival modifiers follow or precede nouns (see Materials and Methods for details).
Given a frequency distribution over syntactic structures, any grammar achieves a certain average dependency length and information locality across the syntactic structures from that distribution.

We compare two groups of word orders: SVO-like order where S and O are ordered on different sides of the verb, and SOV-like order where S and O are ordered on the same side of the verb.
Languages can fall on a spectrum between languages with entirely strict SVO order and languages with entirely strict SOV order~\citep{steele1978word}.
English is close to one end of the spectrum with dominant SVO order, with rare exceptions (e.g., stylistically marked VS order in ``then came a dog'').
Japanese falls entirely on one end of the spectrum, allowing only SOV and OSV order.
Many languages occupy intermediate positions. 
For instance, in Russian, all logically possible orderings of S, V, O can occur, though with different frequencies.

We quantify the position of an individual language on the continuum between SVO and SOV using a quantitative corpus-based metric called subject-object position congruence.
This metric indicates the chance that two randomly selected instances of S and O from a corpus -- not necessarily from the same sentence -- are on the same side of their respective verbs. This number is 1 in strict SOV languages like Japanese, close to 0.5 in languages with flexible word order, and close to 0 in English.







\subsection{Word Order Variation Reflects Competing Pressures}

\begin{figure}
    \centering

	\vspace{0pt}\includegraphics[width=.45\textwidth]{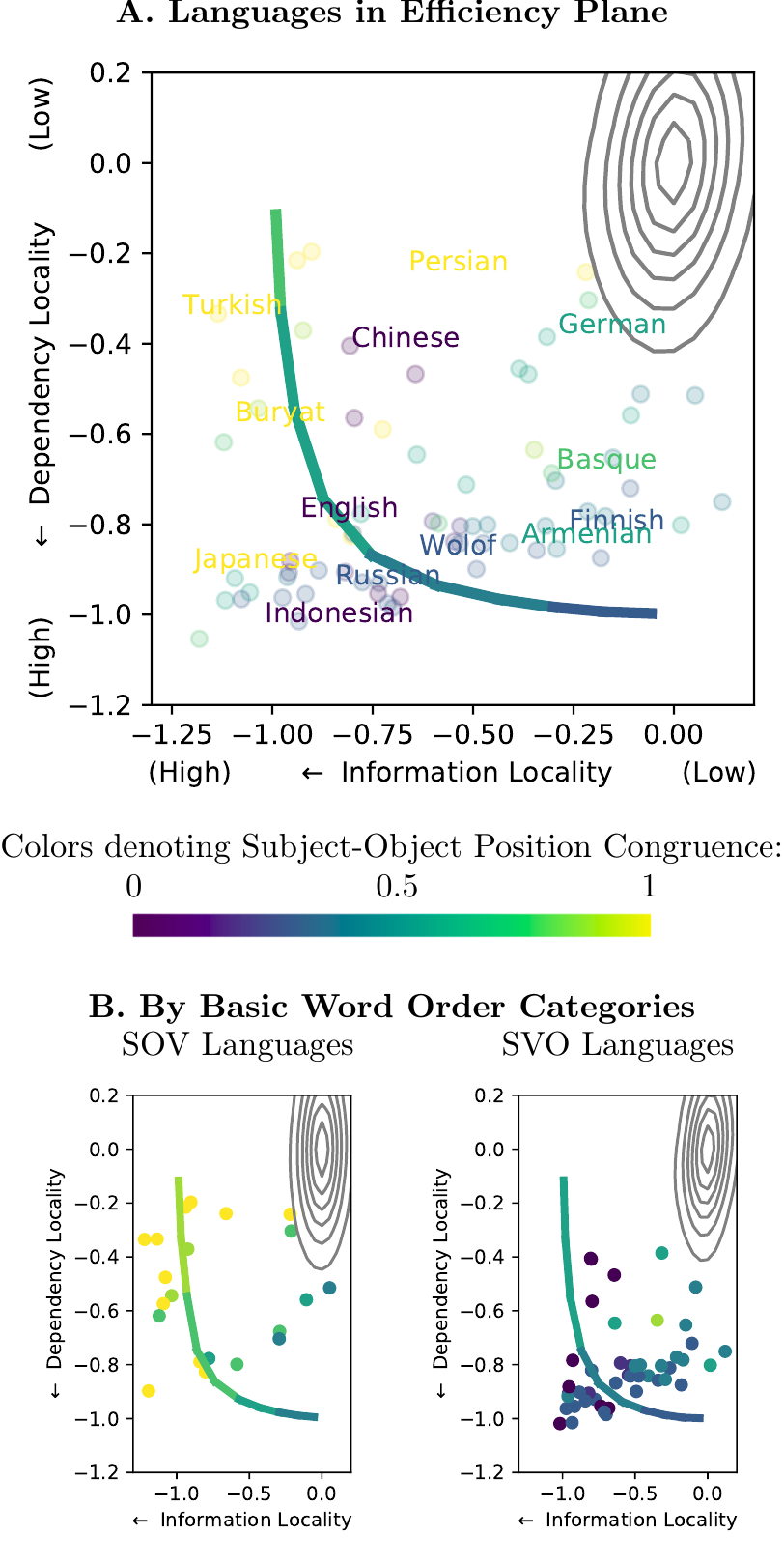}

    \caption{
    Summary of efficiency analysis of crosslinguistic word order variation. A: All 80 languages in the plane spanned by IL (x-axis) and DL (y-axis), with 12 languages annotated (see SI Appendix, Figure S16 for results with all language names). IL and DL are scaled to unit length across languages; see SI Section S22 for raw results. Positions closer to the bottom left corner indicate higher efficiency. Colors indicate subject-object position congruence as attested in corpus data. The contour lines indicate a Gaussian fit to the distribution of baseline grammars averaged across languages. The curve indicates the average of the Pareto frontier across all 80 languages. Some languages are beyond the curve, because they utilize word order flexibility to achieve higher efficiency than any fixed ordering grammar. Colors along the frontier indicate the average subject-object position congruence of counterfactual grammars along the frontier of maximally optimal grammars. B: The same information as shown in A with languages classified as SOV (left) and SVO (right) in the typological literature. The distribution of optimal grammars along the frontier over-represents SOV-like grammars in SOV languages, and SVO-like grammars in SVO languages. }
    \label{fig:pareto-planes}
\end{figure}

Figure~\ref{fig:pareto-planes}A shows the positions of the 80 languages in the efficiency plane spanned by IL and DL.
In order to make efficiency planes comparable across different corpora,  we rescaled the distance between the optimal IL (DL) value and the mean of the baseline to 1.
Contour lines indicate the density of sampled possible grammars; the vast majority of the $\approx 10^{43}$ possible grammars concentrates a in a range of IL and DL separated from the Pareto frontier, and only a vanishing proportion of grammars extends to the frontier.
Consistent with findings from prior work on IL and DL \cite{gildea-human-2015, futrell2020dependency, Hahn2020modeling}, the attested grammars occupy the range between the baseline samples and the frontier, making them more efficient than almost all other possible grammars:
Each language outperformed at least either $\geq$ 90\% of baselines on DL, or $\geq$ 90\% on IL.
All languages outperformed the median baseline on DL, and all but three outperformed the median baseline on IL. 
Some languages are beyond the average curve because their frontiers are to the left of the average curve.
There are also languages that are more efficient than all computationally optimized grammars within the formalism of word order grammars, suggesting they achieve even higher efficiency through flexibility in word order (see SI Section S28).

The attested grammars of the languages, and the possible counterfactual grammars along the Pareto frontier, are colored by their subject-object position congruence as found in corpus data.
DL was correlated with congruence, such that languages with higher subject-object-position congruence optimized DL less strongly ($R=-0.47$, 95\% CI $[-0.63, -0.28]$, $p=10^{-5}$; Spearman's $\rho=0.45$, $p=3\cdot 10^{-5}$).
This agrees with recent findings that SOV languages optimize DL less strongly \cite{futrell-large-scale-2015, Jing2021DependencyLM}.
To account for the statistical dependencies between related languages more rigorously, we grouped the 80 languages into 17 maximal families (or phyla) describing maximal units that are not genetically related to each other (see Materials and Methods), and performed a regression analysis where we entered per-family random slopes and intercepts.
This analysis confirmed a significant effect of congruence on DL ($\beta=-0.37$, 95\% CrI $[-0.62, -0.08]$, $P(\beta\geq 0) = 0.009$, Bayesian $R^2=0.45$).
Subject-object position congruence did not correlate with IL ($R=0.09$, $p>0.05$).

Figure~\ref{fig:pareto-planes} also shows an analogous result for the counterfactual grammars along the frontier:
Among those, subject-object position congruence is higher when DL is not optimized and IL is strongly optimized; it is lower when DL is most optimized.
In a mixed-effects analysis with per-family random effects, subject-object position congruence was significantly higher at the end optimizing for IL (left) than at the end optimizing for DL (bottom) (difference between the two: $\beta=0.38$, 95\% CrI $[0.25, 0.53]$, $P(\beta\leq 0) < 0.0001$).

In Figure~\ref{fig:pareto-planes}B we plot these results specifically for languages classified as SOV and SVO in the typological literature \cite{wals-82, gell-mann-origin-2011}.
SVO languages tend to optimize DL more strongly.
Note that some of the 80 languages belong to less common categories such as VSO, see SI Figure S17 for results for those categories.
We also observe that subject-object position congruence tends to be higher along the frontier for SOV languages than SVO languages.
This observation suggests coadaptation: Language users tend to produce frequency distributions of syntactic structure for which the real word order of the language is efficient.

\begin{figure*}[ht]
    \centering
    
	\vspace{0pt}\includegraphics[width=.95\textwidth]{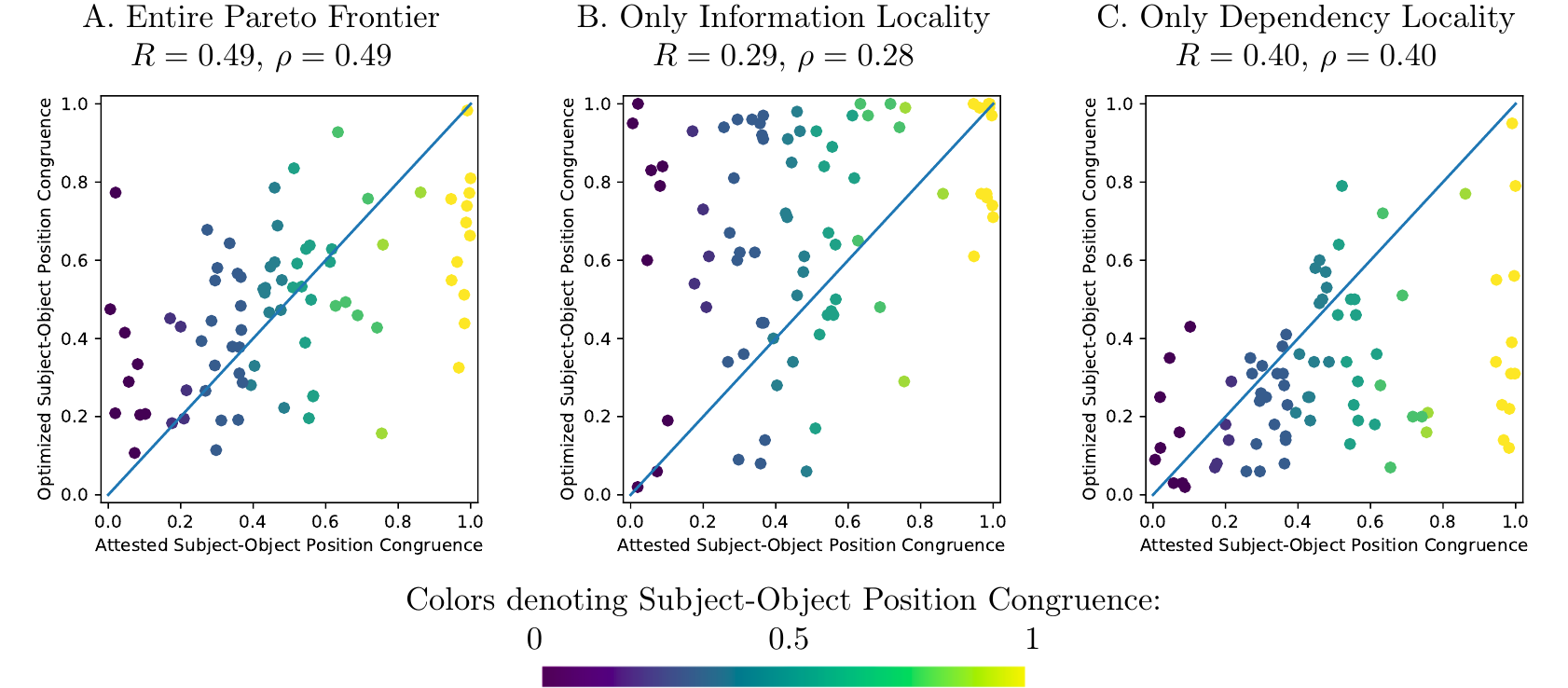}

    \caption{Coadaptation between grammar and usage (see SI Appendix, Figure S11 for further results). Real subject-object position congruence (x-axis) is compared against against the average subject-object position over the entire Pareto frontier (left), at the end optimizing only for IL (center), and at the end optimizing only for DL (right). The line indicates the diagonal. Optimizing for IL or DL tends to over- and under-predict SOV-like orderings, respectively. Considering the joint optimization of both factors results in better prediction of real orderings. 
    }
    \label{fig:congruence-real-optim}
\end{figure*}

To test this idea about coadaptation more rigorously, we computed the average subject-object position congruence along the frontier.
In Figure~\ref{fig:congruence-real-optim}A, we compare average congruence along the frontier with attested congruence.
The correlation between the average congruence and the attested congruence was $R=0.49$ (95\% CI $[0.31, 0.65]$, $p=4 \cdot 10^{-6}$; Spearman's $\rho = 0.49$, $p=4 \cdot 10^{-6}$).
The correlation was substantially lower when considering the density not along the full frontier, but at the two endpoints, where only IL or DL are optimized (Figure~\ref{fig:congruence-real-optim}B--C).
The correlation between the average density and the congruence was significant in a by-families mixed-effects regression predicting the attested congruence from the average density ($\beta=0.62$, $95\%$ CrI $[0.28, 1.0]$, $P(\beta\leq 0) = 0.0011$, Bayesian $R^2 = 0.52$).
Analogous regressions predicting the attested congruence from the density at either of the endpoints yielded inferior model fit (Bayes factor 32 compared to the IL-optimized endpoint, 101 compared to the DL-optimized endpoint).
%
A possible concern is that a majority of the 80 languages belongs to the Indo-European family.
We confirmed the presence of coadaptation in an analogous analysis excluding Indo-European ($\beta=0.72$, 95\% CrI $[0.25, 1.18]$, $P(\beta\leq 0) = 0.0026$, Bayesian $R^2=0.74$), showing that this result is not driven by this family.

\subsection{Word Order Evolves to Maintain Communicative Efficiency}

We have provided evidence that variability in basic word order reflects competing pressures of IL and DL, resolved differently across languages through coadaptation of grammar and usage.
However, this does not rule out the possibility that the observed correlations are artifacts of the common histories of languages descended from common ancestors.

To test whether the observed patterns arise from the process of language evolution, we performed a phylogenetic analysis on the evolution of efficiency and word order.
Phylogenetic analyses have previously been applied to studying the historical evolution of languages~\citep[e.g., ][]{gray2009language,greenhill2009austronesian,chang2015ancestry,sagart2019dated}, including the evolution of word order patterns \citep{dunn-evolved-2011, maurits2014tracing}.

A phylogenetic analysis allows us to construct an explicit model of language change, drawing on two sources of information:
First, in several cases, our dataset includes data from different stages of the same language (such as Ancient Greek and Modern Greek).
Such datasets provide direct evidence of historical development.
Second, using phylogenetic information, the model can also draw strength from contemporary language data:
Data from related languages may permit inferences about their (undocumented) common ancestor, and thus about possible trajectories of historical change \citep{pagel2004bayesian, dunn-evolved-2011, maurits2014tracing}.

In order understand how basic word order evolves, we used a model of drift (or random walks) on phylogenetic trees \citep{felsenstein1973maximum,pagel1997inferring, pagel2004bayesian} to model how grammar and usage frequencies of a language evolve over time.
We describe the state of a language $L$ at time $t$ as a vector $\xi_{L,t}$ encoding (i) the efficiency of the language as parameterized by IL and DL, (ii) its subject-object position congruence, (iii) the average subject-object position congruence along the frontier.
Whenever a language splits into daughter languages, the point $\xi_{L,t}$ continues to evolve independently in each daughter language.
As the components of $\xi_{L,t}$ are continuous, we model their change over time using a random walk given by an Ornstein-Uhlenbeck process (\citep{felsenstein1988phylogenies,hansen1997stabilizing, blackwell2003bayesian}, see Materials and Methods for details).
This process is parameterized by rates of change in the traits, correlations between the changes in different traits, and by the long-term averages of the traits \citep{felsenstein1973maximum,hansen1997stabilizing, freckleton2012fast}.
This model allows us to model the development of word order both in the real language and across the optimization landscape within a single model.

We obtained phylogenetic trees for the 80 languages in our sample from Glottolog~\citep{nordhoff2011glottolog} (see Materials and Methods), and inserted historical stages as inner nodes in these trees.
We fitted the parameters of the model using Hamiltonian Monte Carlo methods (see Materials and Methods for details).

The long-term behavior of the Ornstein-Uhlenbeck model is encoded in its stationary distribution, which describes the likely outcomes of long-term language evolution \cite{gardiner1983handbook}.
First, language evolution maintains relative efficiency: the mean of the stationary distribution was estimated at $\mu=0.35$ ($95\%$ CrI $[0.22, 0.45]$) of the distance between the maximum IL and the mean of the baselines for IL and $\mu=0.28$ (95\% CrI $[0.21, 0.35]$) of the distance between the maximum DL and the baseline mean for DL, with standard deviations 0.36 (95\% CrI $[0.3, 0.44]$) for IL and 0.24 (95\% CrI $[0.2, 0.3]$) for DL, well-separated from the less efficient bulk of possible grammars.
%
%
High congruence was correlated with decreased efficiency in DL ($R=-0.31$, 95\% CrI $[-0.48, -0.14]$, $P(R > 0) = 0.0004$).
Second, attested and average congruence were substantially correlated ($R=0.36$, 95\% CrI $[0.19, 0.53]$, $P(R\leq 0) < 0.0001$), confirming the presence of coadaptation between word order and usage frequencies in resolving the competing pressures of IL and DL. 
When excluding the Indo-European family, there continued to be a correlation between congruence and DL ($R=-0.33$, 95\% CrI $[-0.61, 0.00]$, $P(R\geq 0) = 0.026$), and -- more importantly -- substantial evidence for the presence of coadaptation (correlation between attested and average optimized congruence: $R=0.39$, 95\% CrI $[0.10, 0.65]$, $P(R\leq 0) = 0.005$, see SI Appendix, Section S8 for further details).

\begin{figure}[h]
    \centering
	\vspace{0pt}\includegraphics[width=.55\textwidth]{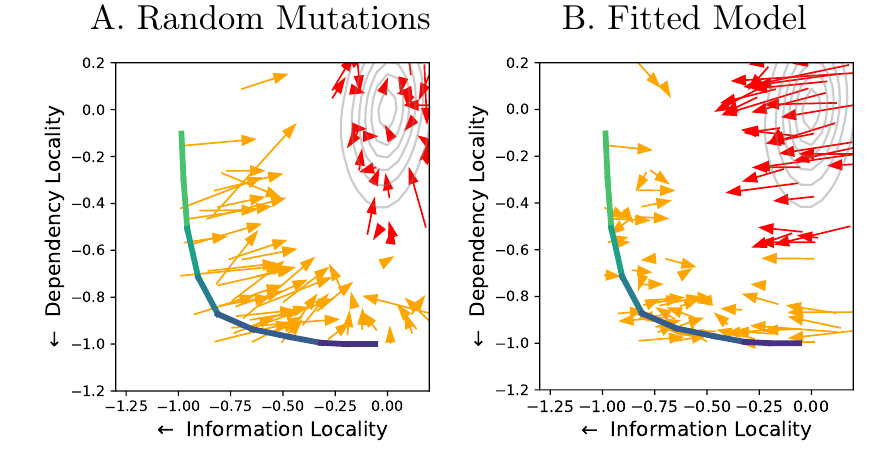}
    
    \caption{A comparison of fitted phylogenetic drift model (right) against random mutations in grammars (left). The analysis is for 40 grammars close to the frontier (orange) and 40 samples from the baseline distribution (red), each evaluated on the tree structure distribution of one of the 80 languages. Arrows denote sampled change due to 200 mutations (left) or $\approx$ 200 yr of language change (right). The fitted model predicts that grammars stay along the frontier, and that inefficient grammars move towards it. In contrast, random mutations drive grammars towards the baseline distribution.}
    \label{fig:drift}
\end{figure}

While the model shows that evolution maintains efficiency, it might be the case that, once languages are close to the frontier, most possible changes would keep languages in that area, so that apparent optimization might be simply the result of neutral drift, rather than a pressure towards increasing efficiency.
To rule out this possibility, we compared the model to neutral drift, which we simulated by iteratively flipping randomly chosen pairs of syntactic relations with minimally differing weights in the grammar.
Results are shown in Figure~\ref{fig:drift}A (see SI Appendix, Section S9 for further results).
For each grammar, we created 30 chains with 200 changes each.
Grammars close to the frontier quickly and consistently move towards the baseline distribution, with very few chains improving efficiency even temporarily.
We contrast this with sample trajectories from the fitted phylogenetic model (Figure~\ref{fig:drift}B); these stay along the frontier, and move towards it when grammars are inefficient.
This provides evidence that language evolution selects specifically for grammatical changes that maintain or increase efficiency.

\begin{figure*}
    \centering
    
    	\vspace{0pt}\includegraphics[width=.98\textwidth]{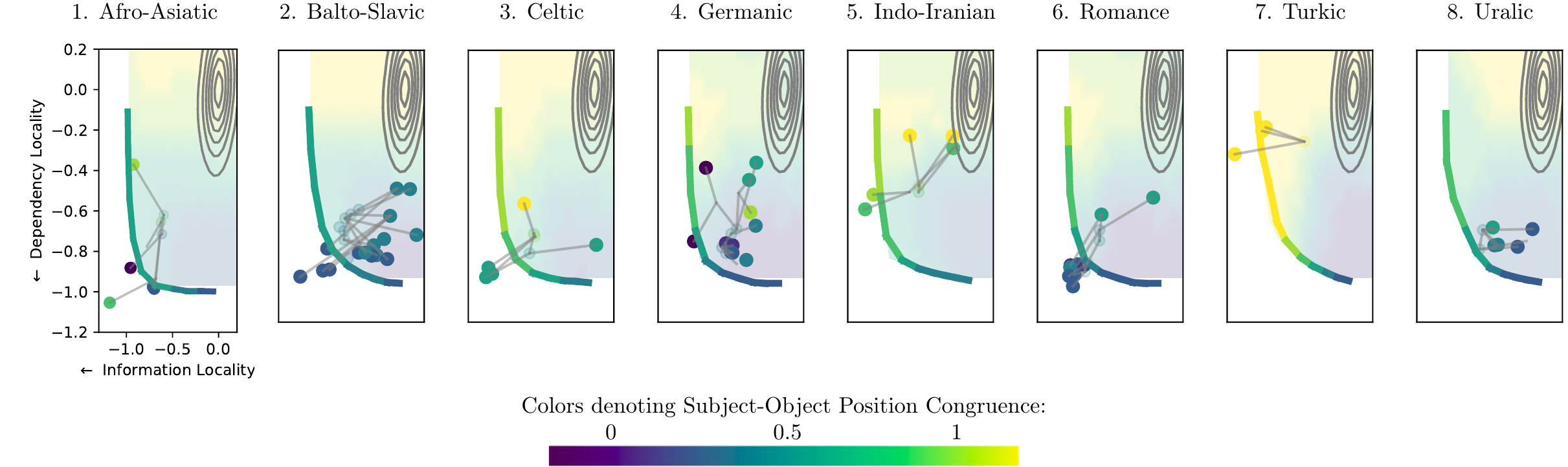}
    
    \caption{Historical evolution within language families. All families (maximal phyla, or subgroups within the well-represented Indo-European family) with at least three attested languages in the dataset are shown. Panels 2--6 show subgroups of Indo-European. For each family, we show the average of the Pareto frontier and of the average congruence on the frontier and behind it. Large full dots indicate attested languages, while small faint dots indicate positions and orderings of historical stages as inferred by the phylogenetic analysis. Languages evolve in the area between the baselines and the frontier. Basic word order tends to match the order most prevalent among optimal grammars. The prevalence of orderings along the frontier varies between families, indicating coadaptation between grammar and usage.}
    \label{fig:pareto-planes-families}
\end{figure*}

We visualize the model fit on language families in our sample in Figure~\ref{fig:pareto-planes-families}.
We plot maximal families (phyla), except within the well-represented Indo-European family, where we plot smaller units.
In addition to the attested languages, we also show historical trajectories as reconstructed by the phylogenetic analysis.
For each family, we show the average congruence across languages.

The families illustrate the findings of the analysis:
First, languages evolve in the area between the baseline samples and the frontier.
Second, evolution towards higher DL goes hand-in-hand with lower subject-object position congruence.
Third, languages with high subject-object position congruence, such as the Turkic languages, also represent such possible grammars more strongly along the Pareto frontier of optimal possible grammars.
See SI Appendix, Section S24 for analyses of historical change in individual languages, which suggest that they either change toward the optimal frontier or stay
near-optimal throughout history.

A well-known variable covarying with word order is case marking: Languages with case marking are more likely to have free word order, and loss of case marking has been argued to correlate with word order becoming more fixed or shifting towards SVO \citep{vennemann1974explanation}, and, conversely, SOV languages mostly have case marking \cite{greenberg-universals-1963}.
This has a clear functional motivation, because case marking can distinguish subjects and objects when order does not.
This raises the question whether our results can be explained away by assuming that both grammar and usage change in response to changes in case marking.
We coded the 80 languages for the presence of case marking distinguishing subjects from objects \cite{wals-49}.
We then conducted a version of the phylogenetic analysis where languages are allowed to concentrate in different regions depending on the presence of case marking.
The rates of change and the long-term averages were allowed to depend on the presence or absence of case marking, and only the correlations between short-term changes in the different traits remained independent of case marking.
We found that the phylogenetic analysis continued to predict correlations in changes of word order and DL, and in changes of usage and word order (see SI  Appendix, Section S12):
Changes were correlated in DL and congruence ($R=-0.39$, 95\% CrI $[-0.60, -0.18]$, $P(R\geq 0) < 0.0001$), and in attested and average optimized congruence ($R=0.49$, 95\% CrI $[0.29, 0.67]$, $P(R \leq 0) < 0.0001$).

\subsection{Relation between Usage and Word Order}
Our results provide evidence that variation in basic word order results from an evolutionary process trading off competing pressures of DL and IL, and that languages resolve these via coadaptation between usage and basic word order.
In what aspects of usage do languages vary or change that reflect this coadaptation?
We predict that languages favor SOV-like orders more when they do not frequently coexpress subjects and objects on a single verb.
Indeed, it has been argued, based on evidence from Turkish and Japanese, that SOV languages omit arguments or use intransitive structures more frequently than SVO languages do \citep{hiranuma1999syntactic,ueno2009does,luk2014investigating}, although this might not hold for Basque \citep{pastor2013processing}.
We tested this idea on a larger scale using the corpora in our sample.
We quantified the frequency of co-expression of subjects and objects by calculating the fraction of all verbs that realize at least a subject or an object simultaneously realize both.
In a linear mixed-effects models, with by-family intercept and slope, attested subject-object position congruence was predictive of this fraction ($\beta=-0.11$, $SE=0.04$, $95\%$ credible interval $[-0.20, -0.03]$, $\operatorname{P}(\beta>0) = 0.006$, Bayesian $R^2 = 0.18$, see Materials and Methods for details). 
Optimized subject-object position congruence was also predictive of this fraction ($\beta=-0.23$, $SE=0.05$, $95\%$ credible interval $[-0.33,  -0.13]$, $\operatorname{P}(\beta>0) < 0.0001$, Bayesian $R^2=0.32$). 
This shows that languages differ in the rate at which they coexpress subjects and objects, and that this is a factor through which frequency and word order can influence each other.

We also found an association between order and coexpression within individual languages: Languages with word order flexibility tend to order subjects differentially depending on the presence or absence of an object, in a way consistent with optimizing DL (see SI Appendix, Section S14).

\section{Discussion}

We have investigated the frequency distribution and historical evolution of word order across languages.
Languages evolve to maintain relative efficiency for Information Locality and Dependency Locality, compared to the vast majority of other logically possible grammars.
We found that variation in basic word order emerges from these two competing pressures, resolved differently across languages through a process of coadaptation  between grammar and usage.

Our results go beyond existing efficiency-based accounts of word order in two ways.
First, extending the cross-sectional synchronic comparison, we explicitly model the evolutionary process through which languages maintain efficiency.
Second, we make specific predictions for individual languages based on their distributions of syntactic structures.
\cite{maurits2010why} propose that the frequency of different basic word order patterns is predicted by a tendency to avoid peaks and troughs in the rate at which information is transmitted (though \cite{gonering-morgan-2020-processing} report a replication with diverging results).
The model of \cite{maurits2010why} accounts for the low frequency of O-initial orders, but it underpredicts SOV and predicts SVO as the most efficient order even when it is applied to Japanese (i.e., an exemplary language for SOV order).
In comparison, our account explains the language-dependence of basic word order.
For Japanese, SOV-like orderings are predominant along the Pareto frontier (see SI Appendix, Section S22), in contrast with the model of \cite{maurits2010why} that predicts SVO across languages.

Among the previous efficiency-based proposals, the one  most closely related to our study, \cite{ferrer-i-cancho-placement-2017}, agrees with our proposal that SVO is favored by DL, but proposes that SOV is favored by a pressure towards predictability of the final verb.
Despite the seeming similarity to our theoretical proposal, there are several key differences.
First, the underlying psycholinguistic theories \cite{hale2001probabilistic, levy-expectation-based-2008} refer to predictability of all words in a sentence, and do not warrant the assumption that it is specifically the verb that should be more predictable.
Second, while the arguments in \cite{ferrer-i-cancho-placement-2017} were theoretical in nature, our proposal here is grounded in large-scale empirical evidence, including both synchronic and diachronic analyses. Specifically, our work reveals strong empirical evidence for coadaptation as a key factor in resolving competing pressures.

The variation between SOV and SVO basic word orders stands in interesting contrast to some other aspects of word order, where a clear preference is observed across languages.
While the relative position of the subject relative to the object is variable, several other syntactic relations have a typologically stable position relative to the object \cite{greenberg-universals-1963, dryer-greenbergian-1992}.
These patterns, known as Greenberg's correlation universals, are predicted uniformly by both DL \cite{frazier1985syntactic, rijkhoff-word-1986, hawkins-performance-1994} and IL (see SI Appendix, Section S21).
The interaction of IL and DL thus explains both why the order of those syntactic relations show uniformity across languages, and why basic word order shows variability.
The process of coadaptation that we discover in basic word order might also operate in other aspects of grammar, and it might explain the observation that language families appear to systematically differ in which subset of Greenberg's correlations  they support \cite{dunn-evolved-2011}.

Our results on coadaptation do not speak to the causal direction between usage and grammar in changes in individual languages. The process of coadaptation as we identified is consistent with causal influence in different directions, with the possibility of a third hidden causal factor. In the future, the availability of more historical data with high temporal resolution might make it possible to explore the causal direction of change in individual languages.
This might also make it possible to determine whether further aspects of usage beyond the coexpression of subjects and objects are affected by coadaptation.

A limitation of our work is that corpus data with syntactic annotation is available primarily for European and Asian languages.
Alternative approaches to estimating usage distributions could leverage either manual annotation of text for relevant quantities such as subject-object coexpression, or automatically projecting syntactic annotation to other languages on multilingual text, such as Bible translations.
However, those approaches would either not reflect usage distributions in sufficient detail to identify optimized orderings, or risk reflecting non-idiomatic properties of translated text.

Our phylogenetic analysis confirmed that grammar and usage patterns evolve together, so that the real order distribution found in a language tends to resemble that dominating along the Pareto frontier of optimal orderings.
This process of coadaptation highlights that grammar and usage frequencies can interact in the evolution of language.
This finding has connections to findings in some other areas.
Existing work has suggested that communicative need, or how frequently a linguistic element is used by its speakers, differs across languages, and that this is responsible for some of the differences observed among the structures of different languages \cite{perkins1992deixis, lupyan2010language, gibson2017color}.
\cite{perkins1992deixis} provides evidence that differences in social structure account for differences in the complexity of deictic inflection across languages; \cite{lupyan2010language} show that languages in small-scale societies tend to have higher inflectional complexity.
Is has been argued that color naming systems are influenced by the way color names are used \cite{zaslavsky2019color}.
\cite{gibson2017color} suggest that color naming systems differ between industrialized and non-industrialized societies due to differences in the usefulness of color in a society, and \cite{regier2016languages} show that vocabulary about the environment depends on the climatic conditions in which a language is spoken.
In syntax, recent work argues that adjective use \cite{kachakeche} and comprehension \cite{RubioFernndez2020SpeakersAL} interact with word order, in a way beneficial for communicative efficiency.
The coadaptation account is also compatible with evidence that human language comprehension itself adapts to the statistics of the language \cite{vasishth2010short}. Related to our findings on the correlation between DL and subject-object position congruence, \cite{yadav2020wordOT} find that SOV languages tolerate longer head-final dependencies, attributing this to adaptation of the human language processing system.

It has been argued that there is an inherent directionality in the evolution of basic word order, and that SOV is the default or original order in the history of language.
Many historically documented word order changes have gone from SVO to SOV, and the protolanguages of several extant families are thought to have been SOV \citep{givon1979understanding, newmeyer2000evolutionary, maurits2014tracing}.
However, only a small fraction of all word order changes are directly documented through written evidence of historical languages. \cite{maurits2014tracing}, using phylogenetic modeling, found that languages can cycle between SOV and SVO over long-term development, with little bias towards either order.
The strongest evidence that SOV might be the `default' order comes from recently emerged sign languages \citep{senghas1997argument, sandler2005emergence, goldin-meadow1998spontaneous, meir2010emerging} and from gesturing tasks \citep{goldin-meadow2008natural, langus2010cognitive}.
If this is true, then our proposed theory would predict that new emerging languages tend to use those structures which maximize the IL advantages of SOV languages.
Indeed, multiple studies report high frequencies of utterances where only one argument is expressed in recently emerged sign languages and home-sign systems \citep{sandler2005emergence, goldin-meadow1998spontaneous, neveu2016sign, ergin2018development}, and in sign languages more generally \citep{napoli2014order}.

Assuming that SOV is the historically earlier order, some studies have further argued that SVO order later arises to avoid ambiguity in communicating reversible events \citep{gibson-noisy-channel-2013, hall2013cognitive}, or to communicate more complex structures \citep{langus2010cognitive, marno2015a, ferrer-i-cancho-placement-2017} and intensional predicates \citep{schouwstra-semantic-2011,napoli2017influence}.
In agreement with our proposed theory, this view also explains the distribution of SOV and SVO in terms of a tension between distinct cognitive and communicative pressures favoring SOV and SVO, respectively~\citep{langus2010cognitive}.
However, those proposals do not explain the fine-grained per-language distribution of word order patterns, since they do not explain why specific languages have SVO or SOV order.
Our theory provides a more precise account of the fine-grained distribution, because it explicitly accounts for the language-dependence of word order, providing per-language predictions of optimal word orders through the process of coadaptation between grammar and usage.

Our work combines evidence from richly annotated syntactic corpora with phylogenetic modeling. This approach can be generally useful for characterizing the fine-grained evolution of grammar in the world's languages.

\section{Materials and Methods}

\subsection*{Ordering Grammars}

The counterfactual order grammars have a weight in $[-1, 1]$ for every one of the 37 syntactic relation annotated in the Universal Dependencies 2.8 corpora (e.g., subject, object).
Dependents of a head are ordered around it in order of these weights; dependents with negative weights are placed to the left of the head, others to its right.
See SI Appendix, Section S2 for more details and examples.

\subsection*{Locality Principles}
We compute dependency length in terms of the Universal Dependencies 2.8 representation format.

Information locality can be formalized in multiple ways grounded in information-theoretic models of human language processing \cite{futrell2020lossy, Hahn2020modeling}.
We adopt a simple formalization in terms of maximizing $\operatorname{I}[X_t : X_{t+1}]$, or the mutual information between adjacent words.
See SI Section S1.1 for other formalization choices.

\subsection*{Inferring Frontier and Congruence along Frontier}
We create approximately optimal grammars using the hill-climbing method of \cite{gildea-optimizing-2007} and the gradient descent method of \cite{hahn2020universals}.
The hillclimbing method is applicable to IL and DL and any combination; the gradient descent method is fast but only applicable to DL.
See SI Section S2.2 for details.
These optimization methods result in approximately optimal grammars populating the area between the baselines and the Pareto frontier.
We ran these optimization methods at linear combinations $(1-\lambda) \cdot \text{DL} + \lambda \cdot \text{IL}$ for $\lambda = 0, 0.25, 0.5, 0.75, 1$, obtaining at least 150 approximately optimal samples per language.
Due to the difficulty of high-dimensional combinatorial optimization, determining the exact Pareto frontier is not feasible; we thus approximated it as the convex hull of the approximately optimized grammars.
For each language, we further randomly constructed at least 75 baseline ordering grammars.
For each language, we interpolated subject-object position congruence throughout the efficiency plane using a Gaussian kernel applied to the approximately optimized grammars and the baseline grammars, with scales chosen for each language using leave-one-out cross validation (see SI Section S3).
See SI Section S22 for raw results.

\subsection*{Data}
We drew on the Universal Dependencies 2.8 treebanks~\citep{DBLP:conf/lrec/NivreMGHMPSTZ20}, including every language for which data with at least 10,000 words was available, while excluding code-switched text, text produced entirely by nonnative speakers, and text only reflecting specific types of sentences (e.g. questions).
Data for one further language (Xibe) became available after completion and was included.
For the phylogenetic analysis, we included additional historical languages as follows.
In addition to the data available in Universal Dependencies 2.6, we added Old English, Medieval Spanish, and Medieval Portuguese dependency treebanks in a slightly different but comparable version of dependency grammar \citep{bech2014iswoc}.
We further split two treebanks spanning multiple centuries (Icelandic and Ancient Greek) into multiple stages based on documented word order changes (see SI Appendix, Section S25).
While there are some other historical treebanks such as the Penn Parsed Corpora of Historical English \citep{kroch2011penn} they are not in the Universal Dependencies format; calculating dependency length is highly nontrivial without a high-quality conversion.

We obtained the topology of phylogenetic trees from Glottolog~\citep{nordhoff2011glottolog}, inserted documented historical languages as inner nodes, and assigned dates for the other inner nodes based on the literature (see SI Appendix, Section S6 for details).

\subsection*{Bayesian Regression Analyses}
We conducted Bayesian inference for mixed-effects analyses using Hamiltonian Monte Carlo in Stan \citep{homan2014the,carpenter2017stan, buerkner2017brms}.
We assumed the prior $N(0,1)$ for the fixed effects slopes, $N(0.5,1)$ for the intercepts, weakly informative Student's $t$ priors ($\nu=3$ degrees of freedom, location $0$, and scale $\sigma=2.5$) for the standard deviations of the residuals and the random effects, and an LKJ(1) prior \citep{lewandowski2009generating} for the correlation matrix of random effects.
See SI Section S16 for details and for results with more strongly regularizing priors.
We computed Bayesian $R^2$ values following \cite{gelman2019r}.
We report $P(\beta \leq 0)$, the posterior probability that the coefficient is not positive, to quantify the fraction of the posterior supporting the hypothesis of a positive coefficient.
The families in the random effects structure correspond to the maximal subgroups in the phylogenetic trees described in `Data'.

\subsection*{Model of Language Change}

We model the state of a language $L$ as a tuple $\xi_L \in \mathbb{R}^4$ consisting of the position of the language in the efficiency plane spanned by DL and IL, the observed subject-object position congruence, and the average subject-object position congruence of optimized grammars.

A common choice in phylogenetic modeling of coevolving traits is correlated Brownian motion, also known as the Independent Contrasts model \citep{felsenstein1973maximum,freckleton2012fast}.
This model can quantify the correlation between pairs of traits (e.g., DL and position congruence) in evolution.
To enable the model to capture biases towards specific parts of the parameter space (e.g., towards regions of high or low efficiency), we added a drift term that can model drift into a specific region.
This leads to an Ornstein-Uhlenbeck process \citep{felsenstein1988phylogenies,hansen1997stabilizing,blackwell2003bayesian}, described by the following stochastic differential equation for the instantaneous change of the state $\xi_{L,t}$ of a language $L$ at a given time $t$:

\begin{equation*}
    \operatorname{d}\xi_{L,t} = \Gamma \cdot (\xi_{L,t}-\mu) \operatorname{d}t + \sqrt{\Sigma} \operatorname{d}B_t
\end{equation*}
where $\mu \in \mathbb{R}^4$ is a vector,  $\Sigma \in \mathbb{R}^{4\times 4}$ is a covariance matrix, $\Gamma \in \mathbb{R}^{4 \times 4}$ is diagonal with positive entries, $B_t$ is Brownian motion in four dimensions, and $\sqrt{\Sigma} \in \mathbb{R}^{4 \times 4}$ is positive-definite and symmetric such that $(\sqrt{\Sigma})^2 = \Sigma$.

The first term, $\Gamma \cdot (\xi_{L,t}-\mu) \operatorname{d}t$, encodes deterministic drift, and describes which region $\mu \in \mathbb{R}^4$ of parameter space languages tend to concentrate around in the long run.

The dynamics of stochastic change are described by the second term, $\sqrt{\Sigma} \operatorname{d}B_t$:
$\Sigma$ is the covariance matrix of instantaneous changes \citep{gardiner1983handbook};
its diagonal entries encode rates of random change in each dimension; the off-diagonals $\Sigma_{ij}$ encode correlations between the instantaneous changes in different dimensions \citep{felsenstein1973maximum,freckleton2012fast}.
Standard results \cite{gardiner1983handbook} imply that the long-term stationary distribution is a Gaussian centered around the vector $\mu$ with a covariance $\Omega$ given as (see SI Appendix, Section S7.1):
\begin{equation*}
    \Omega_{ij} = \frac{\Sigma_{ij}}{\Gamma_{ii} + \Gamma_{jj}}
\end{equation*}
The Pearson correlation coefficient $R$ between the $i$-th and $j$-th components (e.g., DL and position congruence) is then obtained by normalizing the corresponding off-diagonal entry by the individual variances of the two components:
\begin{equation*}
    R_{ij} = \frac{\Omega_{ij}}{\sqrt{\Omega_{ii}\Omega_{jj}}}
\end{equation*}

Without the first term (i.e., with $\Gamma =0$), the simpler Independent Contrasts model \citep{felsenstein1973maximum,freckleton2012fast} would result. 
See SI Appendix, Section S10 for results from that simpler model and model comparison.

In the version controlling for case marking, the parameters $\mu, \Gamma$ are allowed to depend on the presence or absence of case marking.
This model has no unique stationary covariance $\Omega$ across languages with and without case marking; hence, we computed the correlation coefficient $R$ using the covariance matrix $\Sigma$ of instantaneous changes instead of using $\Omega$.

See SI Appendix, Section S7.1 for more on the calculation of the long-term stationary distribution, and SI Appendix, Section S7.3 for the calculation of the correlations between short-term changes in different traits.

We conducted Bayesian inference using Hamiltonian Monte Carlo in Stan \citep{homan2014the,carpenter2017stan}.
See SI Appendix, Section S7.2 for implementation details. 

	\subsection*{Data and Code Availability} Data and code are publicly available at:
\url{https://gitlab.com/m-hahn/efficiency-basic-word-order/}.

\section*{Acknowledgments}
We thank Judith Degen, Richard Futrell, Edward Gibson, Vera Gribanova, Boris Harizanov, Dan Jurafsky, Charles Kemp, Terry Regier, and Guillaume Thomas for helpful discussion and feedback on previous versions of the manuscript. We also thank the editor and the reviewers for their constructive feedback that helped to improve the manuscript. YX is supported by a NSERC Discovery Grant RGPIN-2018-05872, a SSHRC Insight Grant \#435190272, and an Ontario Early Researcher Award \#ER19-15-050.



\setcounter{figure}{0} 
\setcounter{table}{0}
\setcounter{section}{0}
\renewcommand{\thefigure}{S\arabic{figure}}
\renewcommand{\thetable}{S\arabic{table}}
\renewcommand{\thesection}{S\arabic{section}}

\part{Detailed Formulations of Efficiency and Grammar}

\section{Information Locality}\label{sec:info-locality}
\subsection{Formalizing Information Locality}
Information Locality is motivated by the interaction of two prominent psycholinguistic perspectives on what determines human comprehension difficulty in processing syntactic structure.
\textbf{Memory-based theories} \citep{gibson-linguistic-1998,mcelree-memory-2003,lewis-activation-based-2005} propose that comprehension difficulty arises from the difficulty of retrieving and integrating information from preceding context.
\textbf{Expectation-based theories} \citep{hale2001probabilistic,levy2008expectation} states that difficulty arises at points in a sentence that are hard to anticipate from the preceding context.
Jointly considering both perspectives leads to the prediction that words should be easy to process when they are easy to predict from preceding context (as predicted by expectation-based accounts) unless the relevant predictive information has been affected by memory decay or interference (as predicted by memory-based accounts) \citep{demberg-computational-2009, boston2011parallel, Futrell2020LossyContextSA, Hahn2020modeling} (see also \cite{macdonald2002reassessing,frank2016cross,rasmussen2018left} for closely related proposals).
Under this perspective, word order enables efficient processing when predictive information about a word is concentrated in its recent past, so that it can be utilized before it suffers memory decay or interference.
This idea has been formalized using the term \textbf{Information Locality} by \citet{Futrell2020LossyContextSA} and \citet{Hahn2020modeling}, though it is closely related to proposals from preceding work on the role of efficiency in language \citep{qian2012cue}, of relations between usage statistics and conceptual structure in language \citep{Culbertson2020FromTW}, information-theoretic studies of language \citep{ebeling-entropy-1994}, and also to classical experimental findings about the role of contextual constraint on the occurrence of words \citep{Miller1950VerbalCA,Aborn1959SourcesOC}.

\begin{figure}
	\centering
\includegraphics[draft=false,width=0.9\textwidth]{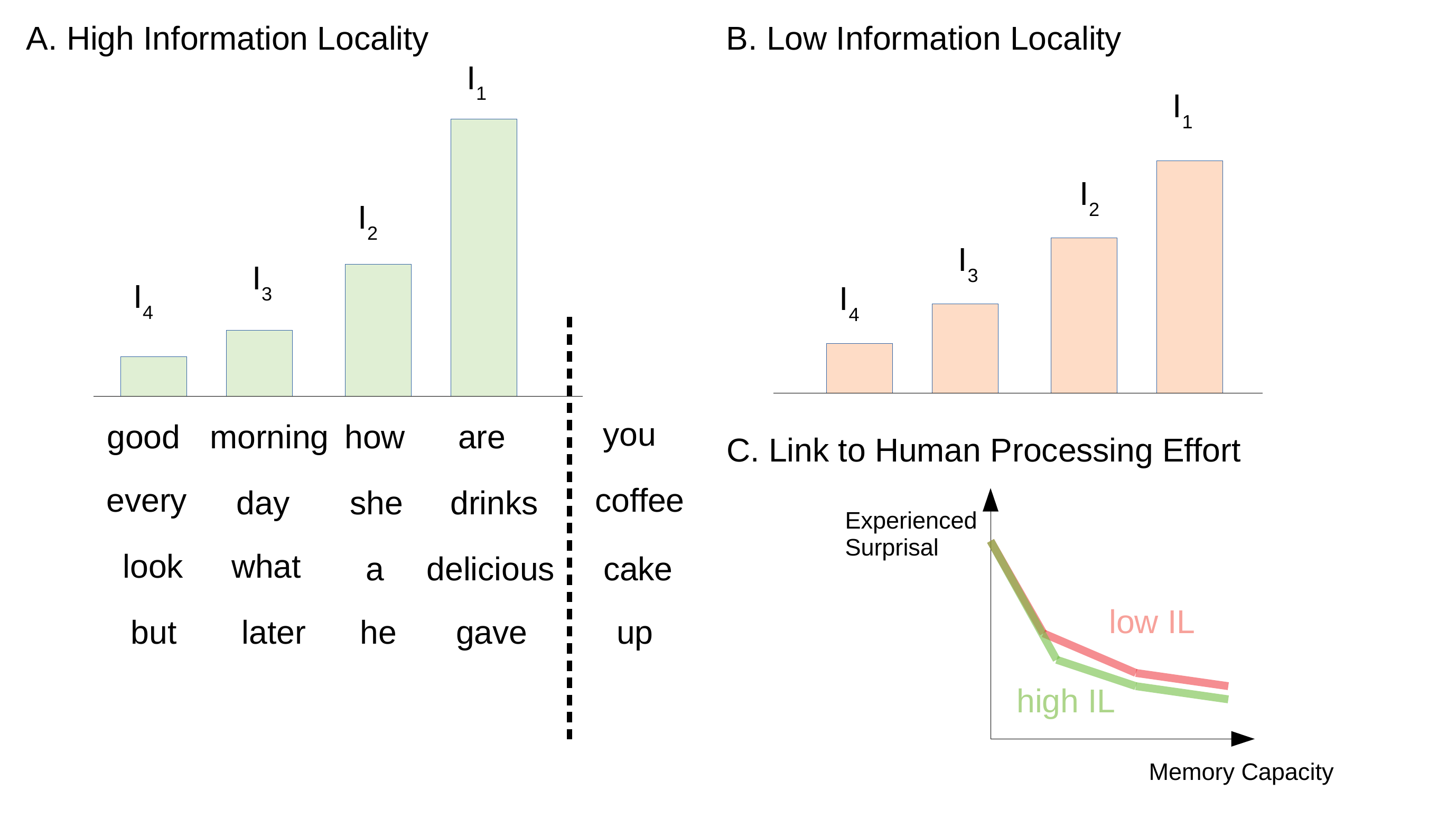} 
	\caption{Information Locality, Mutual Information, and their links to psycholinguistic processing effort. (A) The conditional mutual information $I_t$ measures how much predictive information a word $t$ words in the past provides about the next word, on average across a corpus. While we only show values up to $t=4$, $t$ runs through all integers up to the length of the longest sentence in the corpus. In human language, $I_t$ is largest at $t=1$ and quickly decays as $t$ increases. (B) Another possible situation, where the predictive information is spread out more widely over the past context. Here, $I_1$ is lower and $I_t$ decays more slowly. Such a situation corresponds to a lower degree of Information Locality than in A. (C) The decay of $I_t$ is linked to two aspects of psycholinguistic processing: memory and surprisal. For an individual comprehender, there is a tradeoff whereby a higher memory capacity lowers surprisal on average. The shape of this tradeoff depends on $I_t$ and thus on word order: If $I_t$ decays more quickly (green), a comprehender can achieve lower surprisal at the same memory budget, i.e., the tradeoff is more efficient. The efficiency of the tradeoff can be measured by its area under the curve (AUC), which is lower for the green curve.} 
	\label{fig:info-locality}
\end{figure}



The key formal notion is the \emph{conditional mutual information} between two words $X_i$, $X_{i+t}$ at a distance $t$ (Figure \ref{fig:info-locality} A):
\begin{equation}
	I_t :=	I[X_i, X_{i+t} | X_{i+1} \dots X_{i+t-1}] = \mathbb{E}_{X} \left[ \log \frac{P(X_{i+1}| X_i, X_{i+1} \dots X_{i+t-1})}{P(X_{i+t}|X_{i+1} \dots X_{i+t-1})} \right]
\end{equation}
where the expectation $X$ runs over all sequences of words in the statistics of the language.
The conditional mutual information $I_t$ measures how much predictive information words that are $t$ words apart provide about each other's identity, controlling for information that is redundant with the $t-1$ intervening words, and averaging across all such word pairs in a corpus.


Mutual information is closely related to two other well-studied information quantities \citep[e.g.][]{shannon1951entropy,bentz2017entropy,takahashi2018cross}: the \textit{entropy rate} $H[X_t|\dots, X_{t-2}, X_{t-1}]$ measuring how unpredictable words are in context on average, and the \emph{unigram entropy} $H[X_t]$ measuring the diversity of the distribution over individual words, i.e., how unpredictable a word is without context.
The difference between the two turns out to be\footnote{$H[X_t] - H[X_t|\dots, X_{t-2}, X_{t-1}] = I[X_t : \left(\dots, X_{t-2}, X_{t-1}\right)]$, which is $ \sum_{t=1}^\infty I_t$ by the chain rule of mutual information.}
\begin{equation}\label{eq:predictive-info}
H[X_t] - H[X_t|\dots, X_{t-2}, X_{t-1}] = \sum_{t=1}^\infty I_t
\end{equation}
which measures the total average amount of predictive information contained in the preceding context.

\paragraph{Formalizing Information Locality}
Broadly speaking, Information Locality asserts that language favors orderings where a higher fraction of the overall predictive information (\ref{eq:predictive-info}) is contained at words in the recent context, and only a small fraction is contained in words farther in the past.
This is equivalent to stating that $I_t$ is high for small distances $t$ and decays relatively steeply as $t$ increases (Figure~\ref{fig:info-locality} A--B).

In this paper, we choose maximization of the mutual information between adjacent words $I[X_i, X_{i+1}]$ as a particularly simple operationalization of Information Locality:
\begin{equation}\label{eq:info-adjacent}
I_1 =	I[X_i, X_{i+1}] = \mathbb{E}_{X} \left[ \log \frac{P(X_{i+1} | X_i)}{P(X_{i+1})} \right]
\end{equation}
If this quantity is high, a larger fraction of (\ref{eq:predictive-info}) is provided by the immediately preceding word.
A smaller fraction of the overall predictive information from the past is then contained in context further in the past.
Conversely, if $I_1$ is small, a larger fraction of (\ref{eq:predictive-info}) must be contained further in the past.

%
%
%
%

We next discuss how this relates to proposals from prior work.
\paragraph{Area under Memory-Surprisal Tradeoff Curve}

\citet{Hahn2020modeling} provide a mathematical derivation of information locality in terms of a memory-surprisal tradeoff, combining the expectation-based and memory-based perspectives with a general information-theoretic analysis.
This is formalized by the following theorem about comprehenders processing a stream of words using some (otherwise arbitrary) memory representations $M$ (Figure~\ref{fig:info-locality} C):
If $T \geq 0$ is an integer chosen so that the information-theoretic capacity of the listener's memory representation $M$ satisfies
\begin{equation}\label{eq:memory-bound}
	M \leq \sum_{t=1}^T t \cdot I_t 
\end{equation}
then this comprehender's average surprisal $S$ satisfies
\begin{equation}
	S \geq H[X_t|\dots X_{t-1}] + \sum_{t=T+1}^\infty  I_t 
\end{equation}
They showed that comprehenders can achieve a lower surprisal at the same memory capacity when $I_t$ decays faster.
This happens because of the factor $t$ in (\ref{eq:memory-bound}), which creates a higher memory cost due to predictive information $I_t$ at higher distances $t$.
Our chosen formalization (\ref{eq:info-adjacent}) emerges in the limit of small memory capacities:
For $T=1$, the surprisal bound precisely equals $H[X_t] - I_1$.
A higher value of $I_1$ thus guarantees a lower (i.e., more favorable) surprisal at low memory budgets.

\citet{Hahn2020modeling} proposed to quantify information locality in terms of the area under the memory-surprisal tradeoff curve (AUC): a lower AUC corresponds to a faster decay of surprisal as memory capacity increases, and thus higher IL.
%
%
In Section~\ref{sec:neural}, we compare to results obtained when quantifying IL in terms of this AUC measure as estimated by \citet{Hahn2020modeling}.

\paragraph{N-Gram Surprisal}
\cite{gildea-human-2015} showed that the word orders of five languages minimize trigram surprisal (i.e., $H[X_t|X_{t-2}, X_{t-1}]$), compared to most other possible orderings.
While they justified trigram surprisal as an approximation to surprisal as considered in expectation-based models of processing, it can also be justified as a formalization of information locality: trigram surprisal equals $H[X_t] - I_1 - I_2$; it is thus low if and only if $I_1+I_2$ is high.

\paragraph{Decay of Unconditional Mutual Information}

A line of prior work has considered the \emph{unconditional mutual information} $J_t$ \citep{ebeling-entropy-1994,Futrell2020LossyContextSA,Culbertson2020FromTW}:
\begin{equation}
	J_t :=	I[X_i, X_{i+t}] = \mathbb{E}_{X} \left[ \log \frac{P(X_{i+1} | X_i)}{P(X_{i+1})} \right]
\end{equation}
This differs from $I_t$ in that it does not factor out information redundant with intervening information.
Note that $I_1 = J_1$; thus, our formalization (\ref{eq:info-adjacent}) equivalently states that $J_t$ decays quickly as $t$ increases.






Information locality was stated in terms of unconditional mutual information by \citet{Futrell2020LossyContextSA}, who provided an approximate mathematical derivation in terms of minimizing surprisal under a certain class of  memory loss models.
While they did not provide a full operationalization of Information Locality, they proposed that language favors that words are close together when they have a high (unconditional) mutual information, i.e., $J_t$ decays quickly.

Further related to the principle of Information Locality, \citet{Culbertson2020FromTW} show that the typologically most frequent relative orderings of noun phrase modifiers are such that modifiers are closer to the noun if they have higher mutual information with the noun.
While they interpreted mutual information as reflecting statistical properties of the world that correlate with conceptual structure, their account is fully compatible with the principle of Information Locality as derived from theories of psycholinguistic processing effort.


\paragraph{Decaying Cue Effectiveness}
Relatedly, \citet{qian2012cue} argue that the effectiveness of past predictive information in language production decreases over distance.
They studied the overall predictive information (\ref{eq:predictive-info}) (their ``cumulative discourse informativity'', Formula (4) in their paper) and the decay of $I_t$ (their ``cue effectiveness'', Formula (3) in their paper), proposing that $I_t$ decays over distances $t$ due to, among other factors, limitations of human memory.
They fitted a power law to the decay of cue effectiveness; in this framework, a steep decay is reflected in the coefficients of the power law.
Information locality can also be linked to classical findings that most predictive information about a word, at least as utilized by humans, comes from a few preceding words \citep{Miller1950VerbalCA,Aborn1959SourcesOC}.

%
%
%
%
%
%
%
%
%
%
%
%
%
%

\subsection{Estimating Mutual Information}\label{sec:estimating-mi}
Mutual information is defined in terms of an idealized statistical distribution over all possible sentences; it is thus necessary to approximate it using the available finite corpus data.
We follow the approach of \citet{gildea-human-2015} and Study 3 of \citet{Hahn2020modeling}, drawing on long-standing techniques in natural language processing (see Section~\ref{sec:neural} for a second estimation method, used in Study 2 of \citet{Hahn2020modeling}).

We split each dataset into a training set and a held-out set.
While the UD datasets have predefined splits, those vary substantially in the train/held-out ratio across languages.
We therefore, for each language, randomly sampled a subset whose size was the greater of 100 sentences and 5\% of all sentences, and used those as held-out data, and the remainder as training data.
We estimate the probabilities $p(x_t|x_{t-1})$ using counts from the training set, and estimate the entropies $H[X_t]$, $H[X_t|X_{t-1}]$  as cross-entropies on the held-out data:
\begin{align}
	H[X_t] &\approx - \sum_{i=1}^{|HeldOut|} \log p(x_i) \\
	H[X_t|X_{t-1}] &\approx - \sum_{i=1}^{|HeldOut|} \log p(x_i|x_{i-1})
\end{align}
$I_1$ is then estimated as the difference of these cross-entropies.
This approach of estimating mutual information as a difference of cross-entropies is a well-established method with theoretical guarantees \citep{McAllester2020FormalLO}, avoiding an overestimation bias that would result from naively applying the definition of mutual information to the full dataset.

The method for estimating probabilities $p(x_t|x_{t-1})$ exactly follows Study 3 of \citet{Hahn2020modeling} and is based on Kneser-Ney Smoothing \citep{kneser-improved-1995}, which we describe here for completeness.
First, the unigram probabilities are estimated using Laplace smoothing as
\begin{equation}
	p(w_t) :=   \frac{N(w_t) + 1}{|Train| + |V| \cdot 1}
\end{equation}
where $N(w_t)$ is the number of occurrences of $w_t$ in the training data.
Here $|Train|$ is the number of tokens in the training set, $|V|$ is the number of types occurring in train or held-out data.

Then, conditional probabilities $p_2(w_t|w_{t-1})$ are estimated as follows.
For a sequence $w_1 w_2$, let $N(w_1 w_2)$ be the number of times $w_1 w_2$ occurs in the training set.
If $N(w_{t-1} w_t) = 0$, set
\begin{equation}
	p(w_{t}|w_{t-1}) := p(w_t)
\end{equation}
Otherwise, we interpolate between second-order and first-order estimates:
\begin{equation}
	p(w_t|w_{t-1}) :=  \frac{\operatorname{max}(N(w_{t-1} w_{t}) - 1, 0.0) + \#\{w : N(w_{t-1}w) > 0\} \cdot p(w_t)}{N(w_{t-1})}
\end{equation}
\citet{kneser-improved-1995} show that this definition results in a well-defined probability distribution, i.e., $\sum_{w \in V} p(w|w_{t-1}) = 1$.
This method can be justified as approximate Bayesian inference assuming a Zipfian-like distribution over words \citep{Teh2006ABI}.

%
%
%
%
%
%
%

\section{Ordering Grammars}

\subsection{Ordering Grammar Formalism}
We adopt the word order grammar formalism of \cite{gildea-optimizing-2007,gildea-grammars-2010,gildea-human-2015} to Universal Dependencies.
The original grammar formlism of \cite{gildea-optimizing-2007} is defined for constituency treebanks; it defines weights for each combination of parent and child constituent category (e.g., ``NP$\rightarrow$JJ'' for the position of the adjective within the noun phrase).
We adapt this to Universal Dependencies by defining weights for dependency relation labels (e.g. \textit{amod} for the noun-adjective dependency).

Dependents of a head are ordered in ascending order by their weights, so that dependents with negative weights appear before the head and dependents with positive weights appear after the head.

For instance, a grammar might define the weights (among others)
\begin{center}
nsubj : -0.8
    
obj : 0.3
\end{center}
Applying this to a simple transitive sentence would result in SVO order:

\begin{center}
\begin{dependency}[theme = simple]
   \begin{deptext}[column sep=1em]
          dogs \& bite \& people  \\
          NOUN \& VERB \& NOUN \\
   \end{deptext}
   \depedge{2}{1}{subj}
   \depedge{2}{3}{obj}
\end{dependency}
\end{center}
In contrast, the following grammar, where both weights are negative, results in SOV order:
\begin{center}
nsubj : -0.8
 
obj : -0.3
\end{center}
as in the following example:

\begin{center}
\begin{dependency}[theme = simple]
   \begin{deptext}[column sep=1em]
          dogs \& people \& bite  \\
          NOUN \& NOUN \& VERB \\
   \end{deptext}
   \depedge{3}{1}{subj}
   \depedge{3}{2}{obj}
\end{dependency}
\end{center}


\subsection{Optimization Methods}

\paragraph{Hill-Climbing Method}
The hill-climbing method is adopted from the method of \citet{gildea-optimizing-2007}.
It first randomly initializes the weights of the grammar.
In every iteration, it then randomly chooses one relation and changes the grammar by moving this relation to a randomly selected new position.
If the objective function (a linear combination of IL and DL) improves, the new grammar is adopted, else it is discarded.
We iterate this until the grammar remains stable for 2K iterations, for at most 10K iterations.

\paragraph{Gradient Descent Method}
We further use the gradient-based optimization method of \cite{hahn2020universals} to optimize DL, which converges more quickly than the hill-climbing method, in particular on larger datasets.\footnote{While this method is also applicable to optimizing mutual information (IL), it does not offer an efficiency advantage over the hill-climbing method there.}
This method considers a probabilistic extension of the grammar formalism where each grammar defines a distribution over possible linearizations of a tree; grammars as defined above correspond to the special case where the distribution is always concentrated on one linearization (i.e., it is deterministic).
We refer to \cite{hahn2020universals} for the precise definition of this extension.
This extension makes the average dependency length a \emph{differentiable} function of the grammar parameters, opening the door to the use of gradient-based optimization algorithms for ordering grammars.
The optimization method then applies stochastic gradient descent using the REINFORCE estimator~\citep{williams-simple-1992} to optimize the average dependency length across the trees in the corpus and the possible linearizations of each tree.
Over the course of optimization, the probabilistic grammars converge to essentially deterministic ones that approximately minimize average dependency length across the trees in the corpus.

\section{Interpolating Efficiency Plane and Pareto Frontier}

Here, we describe how we interpolated subject-object position congruence throughout the efficiency plane, and how we approximated the Pareto frontier.
We made all choices before evaluating the hypotheses tested in the paper.
Results do not depend on the smoothing method: See Figure~\ref{fig:raw-coadaptation} for an analysis of coadaptation based on the raw samples that do not depend on the smoothing method, showing equivalent results.

\paragraph{Distribution of Subject-Object Position Congruence}
Given the set of grammar samples  (obtained through approximate optimization or random generation) $\xi_i = (x_i, y_i)$ ($x_i=$ IL, $y_i=$ DL) with associated subject-object position congruences $z_i$,
for each point $\xi = (x,y)$ in the plane spanned by DL and IL, we predict the average subject-object position congruence of grammars at this point with a normalized Gaussian kernel as
\begin{equation}
	f(x,y) := \sum_{i=1}^N w_i z_i
\end{equation}
where
\begin{equation}
	w_i \propto \ L_1 (x_i - x)^2 + L_2 (y_i - y)^2
\end{equation}
and $\sum_i w_i =1$, and $L_1, L_2 > 0$ are chosen to minimize the regularized leave-one-out objective:
\begin{equation}
	\frac{1}{N} \sum_{i=1}^N |f_i(x,y) - z_i|^2 + \lambda \cdot (L_1^2 + L_2^2)
\end{equation}
where $f_i$ arises by leaving out $\xi_i$ from the dataset in the definition of $f$.
We determined a small regularization weight $\lambda = 0.00001$ to prevent smoothing artifacts arising due to excessively large weights $L_i$.
Optimization uses 5K iterations of random search over $L_1, L_2 \in [0,100] \times [0,100]$.

\paragraph{Pareto Frontier}
We fit the approximate Pareto frontier as a spline covering the convex hull of all grammar samples.
The definition is very similar to standard cubic splines \citep{Stoer2002IntroductionTN}, except that we constrained the spline to be convex and monotonic.
More precisely, we selected all sampled grammar points $\xi_i = (x_i, y_i)$ that were not Pareto-dominated by any other point in the convex hull.
For each segment between adjacent points $x_i, x_{i+1}$, we defined a cubic polynomial $g_i(x)$, and determined the coefficients of these cubic polynomials to maximize the area under the curve $\sum_{i=1}^N \int_{x_i}^{x_{i+1}} g_i$ (thus, making the spline fit as closely to the convex hull as possible), suject to the constraints of (i) lower-bounding the convex hull: $g_i(x_i) \leq y_i$, (ii) continuity: $g_i(x_{i+1}) = g_{i+1}(x_{i+1})$, (iii) continuity of the slope $g_i'(x_{i+1}) = g_{i+1}'(x_{i+1})$, (iv) convexity: $g_i'' \geq 0$, (v) monotonicity: $g'_i \leq 0$.
This is a standard linear program, which we solved using cvxpy \citep{Diamond2016CVXPYAP}.

\part{Languages and Datasets}

\section{Corpora and Corpus Sizes}

As described in Methods, we included all UD 2.8 languages with at least 10,000 available words, plus Xibe (new in UD 2.9, published after the other experiments were finished).
We however excluded corpora of code-switched text (Hindi English and Turkish German).
Table~\ref{tab:sizes} shows the corpus sizes for the included UD languages.
Table~\ref{tab:excluded-corpora} shows excluded treebanks from languages otherwise included.

The hillclimbing algorithm is computationally very costly when corpora are very large.
We thus had to focus on subcorpora for three languages: we focused on German-GSD (292K words) for German, Japanese-GSD (193K words) for Japanese, and Czech-PDT (1,509 words) for Czech.
We used all available corpora for the gradient descent method.

\begin{table}
\begin{tabular}{lrrr}
Language & Number of  & Nunber of \\ 
	& Sentences & Words \\\hline
Afrikaans  & 1,934 & 49,260\\
Akkadian  & 2,008 & 25,434\\
Amharic  & 1,074 & 10,010\\
Ancient Greek  & 30,999 & 416,988\\
Arabic  & 28,402 & 1,042,024\\
Armenian  & 2,502 & 52,630\\
Bambara  & 1,026 & 13,823\\
Basque  & 8,993 & 121,443\\
Belarusian  & 25,231 & 305,099\\
Breton  & 888 & 10,054\\
Bulgarian  & 11,138 & 156,149\\
Buryat  & 927 & 10,185\\
Cantonese  & 1,004 & 13,918\\
Catalan  & 16,678 & 546,638\\
Chinese  & 11,998 & 277,871\\
Classical Chinese  & 55,514 & 269,002\\
Coptic  & 1,873 & 48,632\\
Croatian  & 9,010 & 199,409\\
Czech  & 127,507 & 2,223,222\\
Danish  & 5,512 & 100,733\\
Dutch  & 20,944 & 306,720\\
English  & 33,251 & 570,631\\
Erzya  & 1,690 & 17,147\\
Estonian  & 36,508 & 506,637\\
Faroese  & 2,829 & 50,486\\
Finnish  & 36,981 & 397,001\\
French  & 42,832 & 1,132,460\\
Galician  & 4,993 & 164,385\\
German  & 208,440 & 3,753,947\\
Gothic  & 5,401 & 55,336\\
Greek  & 2,521 & 63,441\\
Hebrew  & 6,216 & 161,411\\
Hindi  & 17,647 & 375,533\\
Hungarian  & 1,800 & 42,032\\
Icelandic  & 51,957 & 1,162,040\\
Indonesian  & 7,623 & 168,286\\
Irish  & 5,776 & 131,423\\
Italian  & 35,879 & 818,562\\
Japanese  & 67,031 & 1,490,840\\
Kazakh  & 1,078 & 10,536\\
\end{tabular}
\begin{tabular}{lrrr}
Language & Number of  & Nunber of \\ 
	& Sentences & Words \\\hline
Kiche  & 1,435 & 10,013\\
Komi Zyrian  & 872 & 10,321\\
Korean  & 34,702 & 446,996\\
Kurmanji  & 754 & 10,260\\
Latin  & 22,405 & 284,794\\
Latvian  & 15,351 & 252,334\\
Lithuanian  & 3,905 & 75,403\\
Maltese  & 2,074 & 44,162\\
Manx  & 2,319 & 20,630\\
Mbya Guarani  & 1,144 & 13,089\\
Naija  & 9,242 & 140,859\\
North Sami  & 3,122 & 26,845\\
Norwegian  & 42,869 & 666,984\\
Old Church Slavonic  & 6,338 & 57,563\\
Old East Slavic  & 17,901 & 180,110\\
Old French  & 17,678 & 170,740\\
Persian  & 35,104 & 654,696\\
Polish  & 40,398 & 499,392\\
Portuguese  & 22,442 & 571,085\\
Romanian  & 40,480 & 937,540\\
Russian  & 85,789 & 1,420,647\\
Sanskrit  & 4,227 & 28,960\\
Scottish Gaelic  & 3,798 & 72,422\\
Serbian  & 4,384 & 97,673\\
Slovak  & 10,604 & 106,097\\
Slovenian  & 11,188 & 170,158\\
Spanish  & 34,693 & 1,015,119\\
Swedish  & 12,269 & 206,856\\
Tamil  & 1,134 & 12,165\\
Thai  & 1,000 & 22,322\\
Turkish  & 72,151 & 628,938\\
Ukrainian  & 7,060 & 122,091\\
Upper Sorbian  & 646 & 11,196\\
Urdu  & 5,130 & 138,077\\
Uyghur  & 3,456 & 40,236\\
Vietnamese  & 3,000 & 43,754\\
Welsh  & 1,833 & 36,837\\
Western Armenian  & 1,780 & 35,926\\
Wolof  & 2,107 & 44,258\\
Xibe (UD 2.9) & 810 & 15,401\\
       &            & \\
\end{tabular}
	\caption{Corpus sizes of the included UD languages. Experiments used UD 2.8, except in Xibe (UD 2.9), which was published after the other experiments were finished.}\label{tab:sizes}
\end{table}

\begin{table}
\begin{longtable}{lllll}
Treebank & Rationale \\ \hline
Chinese-CFL & Text written by non-native speakers\\
English-ESL & Text written by non-native speakers \\
English-Pronouns & Specifically targets pronouns \\
French-FQB & Consists entirely of questions \\
Latin-ITTB & Consists of Medieval Latin text \\
Latin-LLCT & Consists of Medieval Latin text  \\ 
\end{longtable}
	\caption{UD corpora excluded, from languages otherwise included.}\label{tab:excluded-corpora}
\end{table}

\section{Historical Languages}

Table~\ref{tab:historical} shows the historical languages in our dataset, with approximate dating assigned.

\begin{table}
\begin{longtable}{llp{10cm}llll}
Language & Time & Rationale \\ \hline
Classical Chinese & 300 BC & Life of Mengzi (died around 300 BC); the treebank contains his teachings as collected by his followers. \\
Ancient Greek & 400 BC & Approximate mean age of texts used \\
Coptic & 400 AD & Dating of the Apophthegmata Patrum texts used in the UD treebank\\
Gothic & 350 AD & Life of bible translator Ulfilas (311--383)\\
Latin & 0 AD & Approximate mean age of texts used \\
	Medieval Spanish & 1400 AD & Approximate mean age of texts used (not from Universal Dependencies, see Section~\ref{sec:other-historical}). \\
	Medieval Portuguese & 1400 AD & Approximate mean age of texts used (not from Universal Dependencies, see Section~\ref{sec:other-historical}). \\
Old Church Slavonic & 850 AD &  Bible translation after invention of Glagolitic alphabet around 850 AD. \\
Old English & 900 AD & Approximate mean age of texts used (not from Universal Dependencies, see Section~\ref{sec:other-historical}). \\
Old East Slavic & 1200 AD & Approximate mean age of texts used\\
Old French & 1200 AD  & Approximate mean age of texts used\\
Sanskrit & 900 BC & Approximate mean age of texts used \\
\end{longtable}
	\caption{Historical languages in our dataset.}\label{tab:historical}
\end{table}

\section{Phylogenetic Tree}

\subsection{Tree Topology}

\begin{figure}
    \centering
	\rotatebox{270}{\includegraphics[draft=false,width=1.3\textwidth]{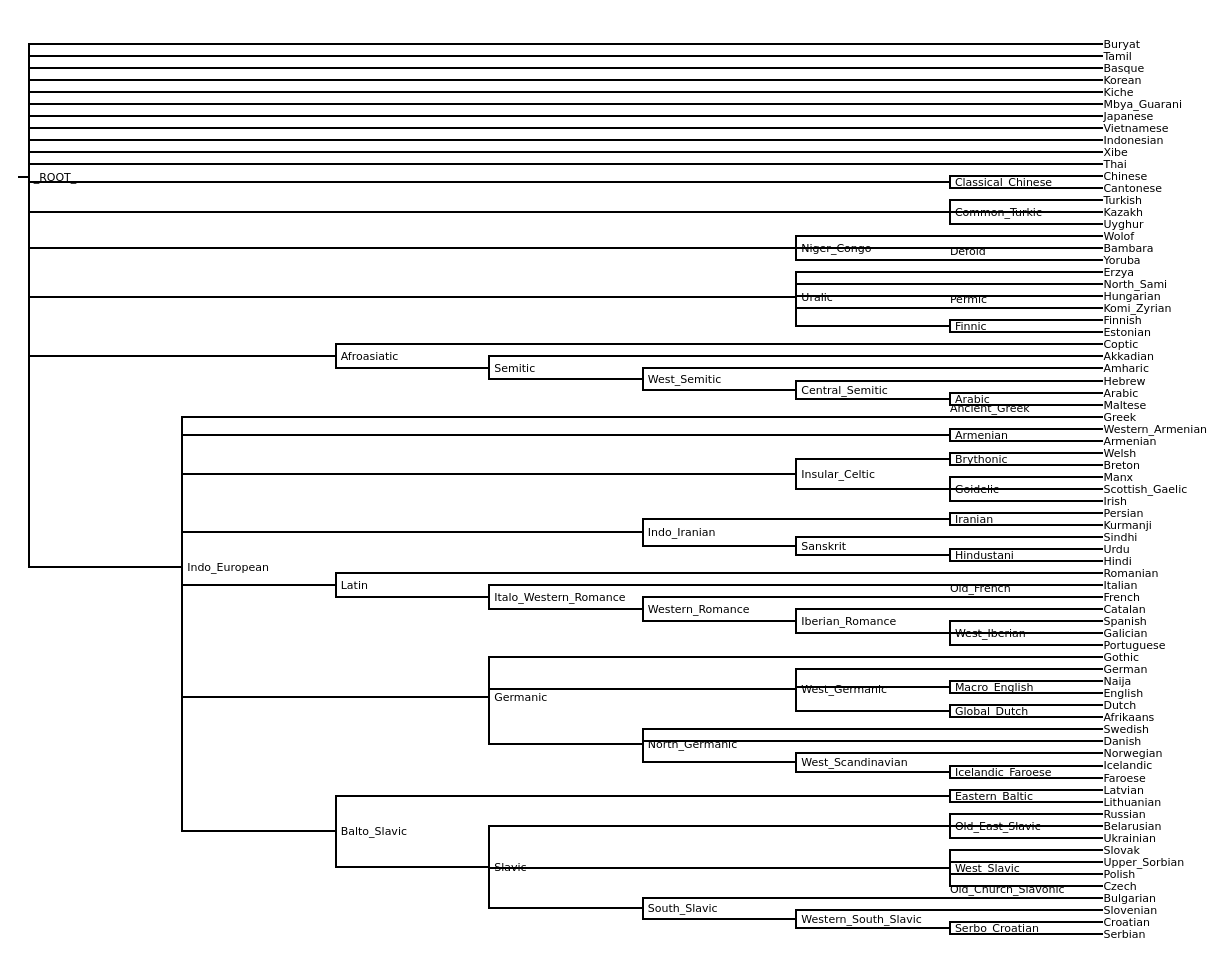}}
       \caption{Phylogenetic tree topology of the languages in our sample. Compare Figure~\ref{fig:tree-times} for a version indicating the time depth of different families.}
    \label{fig:tree}
\end{figure}

We obtained tree topologies from Glottolog~\citep{nordhoff2011glottolog}.
We only retained interior nodes when more than one of their daughter nodes had languages in our dataset.
The resulting tree topology is displayed in Figure~\ref{fig:tree}.\footnote{Tree obtained with https://icytree.org/.}

\begin{figure}
    \centering
	\rotatebox{270}{\includegraphics[draft=false,width=1.3\textwidth]{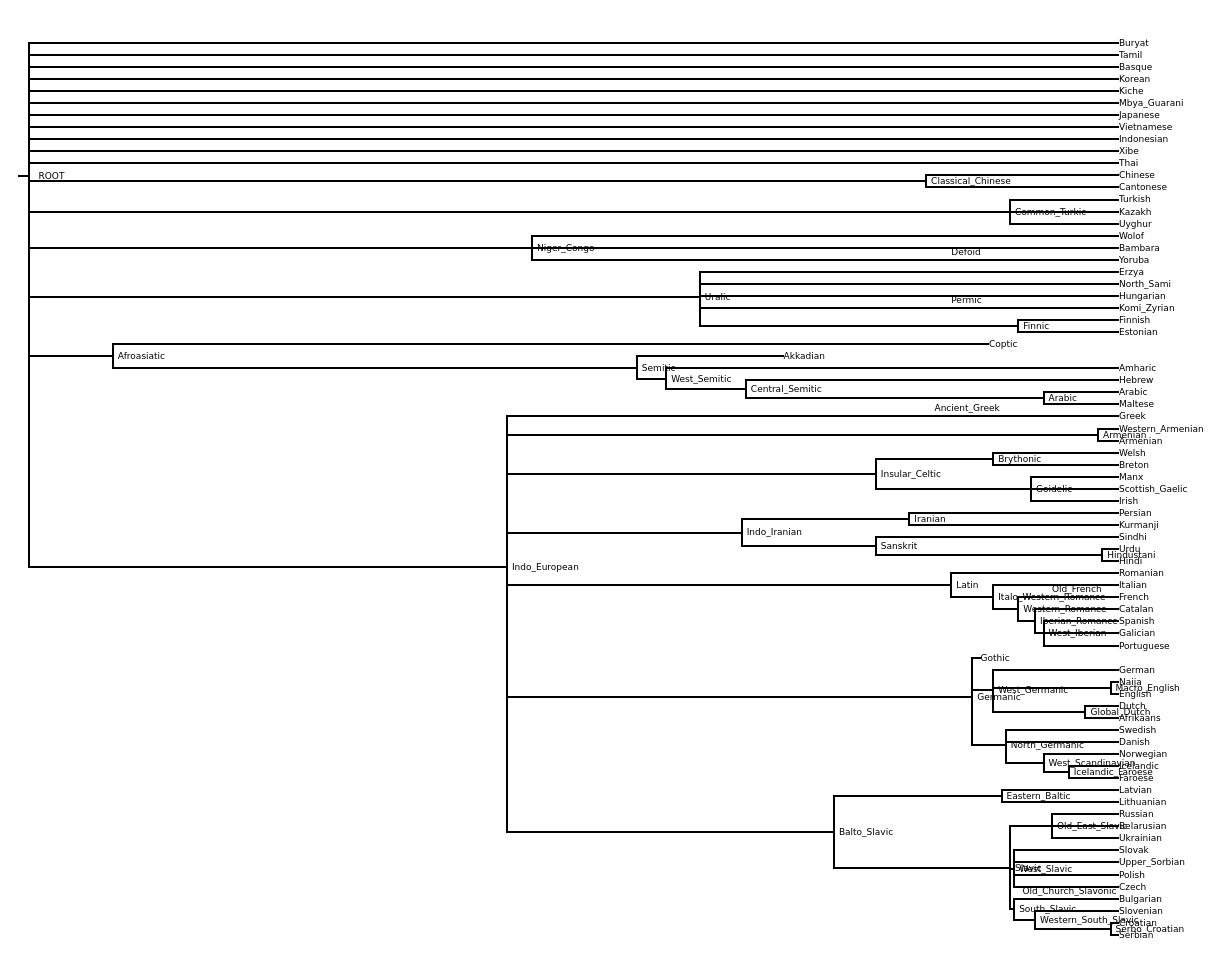}}
	\caption{Phylogenetic tree of the languages in our sample. The length of branches reflects distance in time. Compare Figure~\ref{fig:tree} for a version indicating the raw topology without time depths.}
    \label{fig:tree-times}
\end{figure}

\subsection{Dating Inner Nodes}\label{sec:inner-nodes}
We labeled interior nodes for the time at which they split into descendants, using estimates based on historical evidence and the linguistic literature:

\begin{longtable}{llp{10cm}lll}
Group & Split & Source or Rationale \\ \hline
Afroasiatic & 10,000 BC & \citet{diakonoff1998the} \\
Arabic & 1,100 AD & Calibration from \citet{holman2011automated} based on end of Arabic domination of Malta. \\
	Armenian &  1,750 AD & Separate development of Eastern and Western standards \citep[p. 1]{DumTragut2009ArmenianME} \\
Balto-Slavic & 1,400 BC & \citet{gray2003language} \\
	Brythonic & 500 AD & Migrations from Britain to Brittany \citep{holman2011automated}\\ 
Central-Semitic & 2,450 BC & \citet{kitchen2009bayesian}   \\
Common Turkic & 700AD & \citet[p. 49]{savelyev2020bayesian} estimate Common Turkic to have split around 474 AD. However, in their model, Old Turkic split off around 650 AD, earlier than the languages in our dataset, with uncertainty about the time of split of the remaining Common Turkic languages. It should predate the earliest documentation of Karluk Middle Turkic after 900AD. We thus put the divergence of the other Common Turkic languages at 700AD. \\
Eastern Baltic & 600 AD & Split between Latvian and Lithuanian \citep[p. 209]{novotna2011glottochronology}\\
Finnic & 800 AD & \citet[Section 4.1]{maurits2020best} \\
Germanic & 250AD & \cite{gray2003language} \\
Global Dutch & 1,600 AD & Dutch colony in South Africa \\
Goidelic & 950 AD & Migrations from Ireland to Scotland. \citet{holman2011automated}, citing \citet{jackson1951gaelic}, calibrates the divergence between Irish and Scottish Gaelic to 950 AD. \\
Hindustani & 1,800 AD & Standardization of Hindi and Urdu\\
Iberian Romance & 1,000 AD & Expansion of Christian kingdoms in Iberia, earliest Iberian Romance texts \\
Icelandic-Faroese & 1,400 AD & Sound shifts specific to Faroese\\
Indo-European & 5,300 BC & \citet{gray2003language} (excluding Hittite and Tocharian, for which we have no corpus data). \\
Indo-Iranian & 2,500 BC & \citet[p. 138]{parpola2013formation} \\ 
Insular Celtic & 900BC & \citet{gray2003language} estimate 900BC. \\ 
Iranian & 500 BC & \citet{gray2003language}. \\ 
Italo-Western-Romance & 500 AD & End of the Western Roman empire \citep{holman2011automated}.  \\
	Macro-English & 1900AD & In our dataset, this is the common ancestor of contemporary English and Naija (Nigerian Pidgin). \\
Niger-Congo & 5000BC & \citet{holman2011automated} estimate an age of 6227 years, but the family has to be older than Atlantic-Congo, which they estimate at 6525 years. We thus place Niger-Congo at 5000BC.\\
North-Germanic & 650 AD & Split of Old Norse into regional variants, such as assimilation of nasals to following stops in Western Norse in the 7th century \citep[p. 1856, 1859]{sandoy2017202}. Similarly \citet{holman2011automated} calibrates this to 900 AD. \\
Semitic & 3,750 BC & \citet{kitchen2009bayesian} \\
Serbo-Croatian & 1,900 AD & Standardization of Serbian and Croatian\\
Slavic       & 700AD & \citet{gray2003language}. \citet[p. 209]{novotna2011glottochronology} date the split of East Slavic to the 6th century, \citet{holman2011automated} calibrates it to 550AD. \\
South-Slavic & 750 BC & Expansion of Slavic into Balkan. Postdates Slavic and antedates Old Church Slavonic (attested after 800AD) \\
Uralic & 3,000 BC & \citet[Section 4.7]{maurits2020best}, cf \citep[p. 144]{parpola2013formation} for references \\
West Iberian & 1,100AD & Independence of Portugal \\
West-Germanic & 500 AD & Migrations into Britain and southern central Europe\\
West-Scandinavian & 1,100 AD & Sound shifts specific to Norwegian\\
West-Semitic & 3,400 BC & \citet{kitchen2009bayesian}  \\
West-Slavic & 750 BC & Expansion of Slavic. \\ 
Western Romance & 800 AD & Expansion of Christian kingdoms into Iberia \\
Western South Slavic & 1,000 AD & Antedates earliest Slovenian and Serbo-Croatian texts\\
\end{longtable}

\part{Phylogenetic Analyses}

\section{Details for Phylogenetic Models}

\subsection{Calculating the Likelihood}
For completeness, we describe how to calculate the likelihood of a multidimensional Ornstein-Uhlenbeck model on phylogenetic trees \citep{felsenstein1988phylogenies,hansen1997stabilizing, blackwell2003bayesian}.
As described in the Methods section, it is described by the following stochastic differential equation for the instantaneous change of the state $\xi_{L,t} \in \mathbb{R}^4$ of a language $L$ at a given time $t$:
\begin{equation}\label{eq:sde}
    \operatorname{d}\xi_{L,t} = \Gamma \cdot (\xi_{L,t}-\mu) \operatorname{d}t + \sqrt{\Sigma} \operatorname{d}B_t
\end{equation}
where $\mu \in \mathbb{R}^4$,  $\Gamma$ is non-degenerate, and $\Sigma \in \mathbb{R}^{4\times 4}$ is a covariance matrix, and $B_t$ is multidimensional Brownian motion.
In our model, $\Gamma$ is diagonal with positive entries.

The conditional distribution of a future observation at time $t+\Delta$ given an earlier one at time $t$ is given by the following equation \citep[Theorem 3.3]{schach1971weak}, \citep{gardiner1983handbook}, \citep[p. 156, eq. 6.124]{risken1989fokker}:
\begin{equation}\label{eq:propagator}
\xi_{L,t+\Delta} | \xi_{L,t} \sim N\left(\mu + e^{-\Delta \Gamma} (\xi_{L,t}-\mu),\ \Omega - e^{-\Delta \Gamma} \Omega e^{-\Delta \Gamma^T}\right)
\end{equation}
where the matrix $\Omega \in \mathbb{R}^{4\times 4}$ is obtained as the solution of the equation \citep[p. 110, eq. 4.4.51]{gardiner1983handbook} \citep[p. 156, eq. 6.126]{risken1989fokker}:
\begin{equation}
    \Gamma\Omega+\Omega\Gamma^T = \Sigma
\end{equation}
This can be solved as follows (recall that $\Gamma$ is diagonal in our model):
\footnote{If $\Gamma$ is not diagonal, and $\xi$ has two dimensions: 
\begin{equation}\label{eq:sigma-omega}
\left(\begin{matrix} \Omega_{11} \\ \Omega_{12} \\ \Omega_{22} \end{matrix}\right)=    \left(\begin{matrix}
    2\Gamma_{11} & 2\Gamma_{12} & 0 \\
    \Gamma_{21} & \Gamma_{11}+\Gamma_{22} & \Gamma_{12} \\
    0 & 2\Gamma_{21} & 2\Gamma_{22}
    \end{matrix}\right)^{-1}  \left(\begin{matrix} \Sigma_{11} \\ \Sigma_{12} \\ \Sigma_{22} \end{matrix}\right)
\end{equation}
}
\begin{equation}\label{eq:sigma-omega}
	\Omega_{ij} = \frac{\Sigma_{ij}}{\Gamma_{ii} + \Gamma_{jj}}
\end{equation}
One can compute the stationary distribution that solves the differential equation as follows.
The stationary distribution of an individual observation is
\begin{equation}\label{eq:ornuhl-var}
\xi_{t} \sim N\left(\mu, \Omega \right)
\end{equation}
The stationary cross-covariance between the states of two languages $L_1, L_2$, possibly on different branches of the phylogenetic tree, is given by
\begin{equation}\label{eq:ornuhl-covar}
Cov(\xi_{L_1}, \xi_{L_2}) = e^{-\Delta_1 \Gamma} \Omega e^{-\Delta_2 \Gamma^T}
\end{equation}
where $\Delta_1, \Delta_2$ are the times of evolution from their last common ancestor to $L_1$ and $L_2$, respectively.
\footnote{This can be shown as follows:
If $\xi_A$ is the last common ancestor, then (we set $\mu=0$ without loss of generality, as it does not affect the covariance):
\begin{align*}
Cov(\xi_{L_1}, \xi_{L_2}) &= \mathbb{E} \left[\xi_{L_1} \xi_{L_2}^T\right] - \mathbb{E}\xi_{L_1} \mathbb{E}\xi_{L_2}^T   &= \mathbb{E}\left[\mathbb{E} \left[\xi_{L_1} \xi_{L_2}^T | \xi_A\right]\right] - 0 \cdot 0  &= \mathbb{E}\left[\mathbb{E} \left[\xi_{L_1}|\xi_A\right] \mathbb{E} \left[\xi_{L_2}^T | \xi_A\right]\right]  
 &= \mathbb{E}\left[   e^{-\Delta_1\Gamma} \xi_A    \xi_A^T e^{-\Delta_2\Gamma^T} \right]  \\
 &= e^{-\Delta_1\Gamma} \Omega e^{-\Delta_2\Gamma^T}   \\
\end{align*}}
If $L_1, L_2$ do not share a common ancestor (the root in Figure~\ref{fig:tree} does not count as an ancestor), the covariance is zero.\footnote{As $\lim_{\Delta \rightarrow \infty} e^{-\Delta B} = 0$, this is practically equivalent to assuming a very large time-depth of the last common ancestor, which would be the case under the assumption of macrofamilies with very large time depth.}

Since any Ornstein-Uhlenbeck process is Gaussian \citep{schach1971weak}, the joint distribution of any set of observations $\xi_{L, t}$ is determined by (\ref{eq:ornuhl-var}-\ref{eq:ornuhl-covar}).

\subsection{Implementation}\label{sec:prior-lkj}

We defined the following priors on the parameters.
We parameterized $\Sigma$ as the combination of a correlation matrix and a vector of standard deviations \citep{barnard2000modeling}.
Stated differently, we parameterized $\Lambda := \sqrt{\Sigma}$ as $D U$, where $U$ is a lower-diagonal matrix, and $D$ is a diagonal matrix.
We directly parameterized $\Gamma$ using its diagonal entries.

To define a prior over $\Sigma$, we modeled $U$ as the lower Cholesky factor of a correlation matrix subject to an LKJ(1) prior (\citet{lewandowski2009generating}, i.e., the uniform distribution over $4\times 4$-correlation matrices).
We placed a standard normal prior $N(0,1)$ on the entries of $\mu$ and on the non-zero entries of $D$ and $\Gamma$.

We rescaled times so that 1000 years corresponded to one unit. 
We further rescaled the four components to range from -1 to 1 (instead of [-1,0] for IL/DL, and [0,1] for position congruence).

We implemented the models in Stan~\citep{carpenter2017stan} and obtained posterior samples using the No-U-Turn sampler.
We ran four chains with 10,000 iterations each, of which the first half each were discarded as warmup samples.

The model can be implemented either using the analytical formula for the cross-covariance (\ref{eq:ornuhl-covar}), or by explicitly modeling the inner nodes of the tree using (\ref{eq:propagator}).
The first approach makes posterior inference more efficient, but we specifically used the second approach in the analysis where $\mu, \Gamma$ depended on geography (Section~\ref{sec:areal}) or case marking (Section~\ref{sec:case}), as the cross-covariance is hard to compute explicitly when parameters vary.

We computed marginal likelihoods using Stepping Stone Sampling \citep{xie2011improving} with $K=10$ stones.
We verified stability of the estimates by running the procedure ten times for each model, and averaging the obtained marginal likelihoods.

\subsection{Correlation Component of $\Sigma$}\label{sec:instant-corr}
In the main analysis, we reported the correlation between two dimensions in the stationary distribution $\Omega$.
In some analyses, there are multiple stationary distributions (depending on geography in Section~\ref{sec:areal} and case marking in the main paper and Section~\ref{sec:case}).
In these cases, we therefore report correlations for the instantaneous changes at any point in time:
The matrix $\Sigma$ indicates the variance-covariance structure of the instantaneous changes at any time $t$ \citep{felsenstein1973maximum, freckleton2012fast}.
The main quantity of interest is the correlation between changes in two dimensions (e.g., attested and average optimized subject-object position congruence) \citep[cf.][]{felsenstein1973maximum,freckleton2012fast}, which is given by
\begin{equation}
	R_{ij} := \frac{\Sigma_{i,j}}{\sqrt{\Sigma_{i,i}\Sigma_{j,j}}}
\end{equation}
A positive value indicates that changes in both directions are positively correlated.

\section{Detailed Results for Phylogenetic Model}\label{sec:phylo-results}

\begin{figure}
	\centering
	\begin{tabular}{cccc}
		&		Fitted on Full UD Dataset & Excluding Indo-European \\
		&		\includegraphics[draft=false,width=0.35\textwidth]{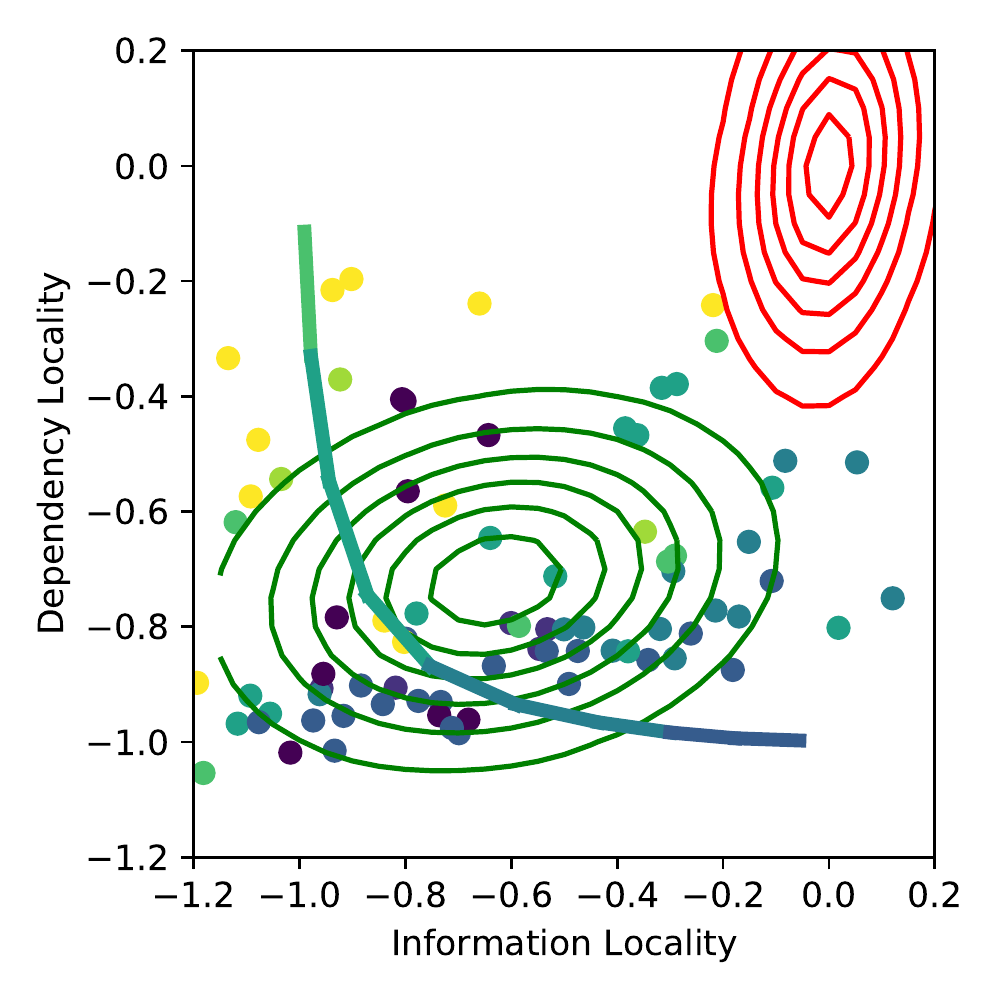} &
		\includegraphics[draft=false,width=0.35\textwidth]{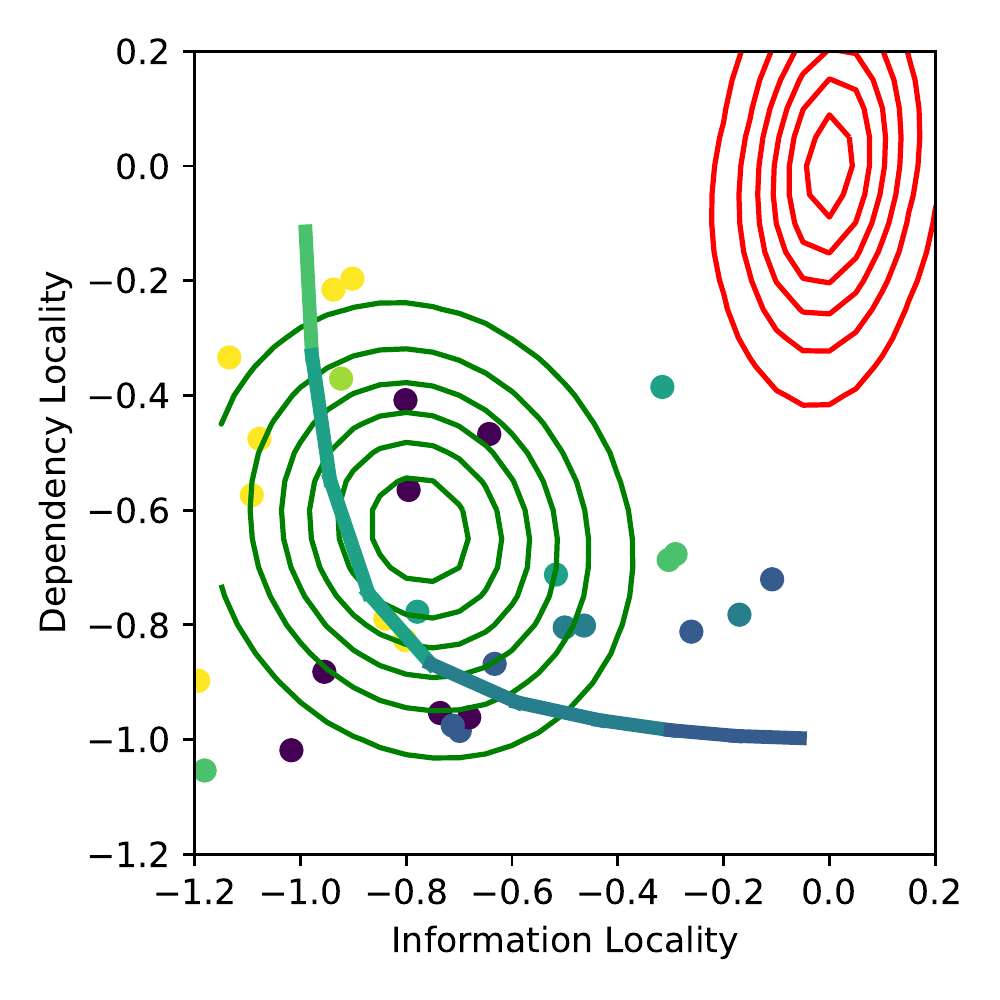}  \\
		\\
	\end{tabular}
	\caption{Stationary distribution in the plane spanned by optimized and attested subject-object position congruence; this indicates the region in which languages tend to move over the course of long-term evolution. The left column shows results on the entire dataset, the right column shows results excluding the Indo-European family.}\label{sec:stationary-il-dl}
\end{figure}

\begin{figure}
	\centering
	\begin{tabular}{cccc}
		Fitted on Full UD Dataset & Excluding Indo-European \\
		\includegraphics[draft=false,width=0.45\textwidth]{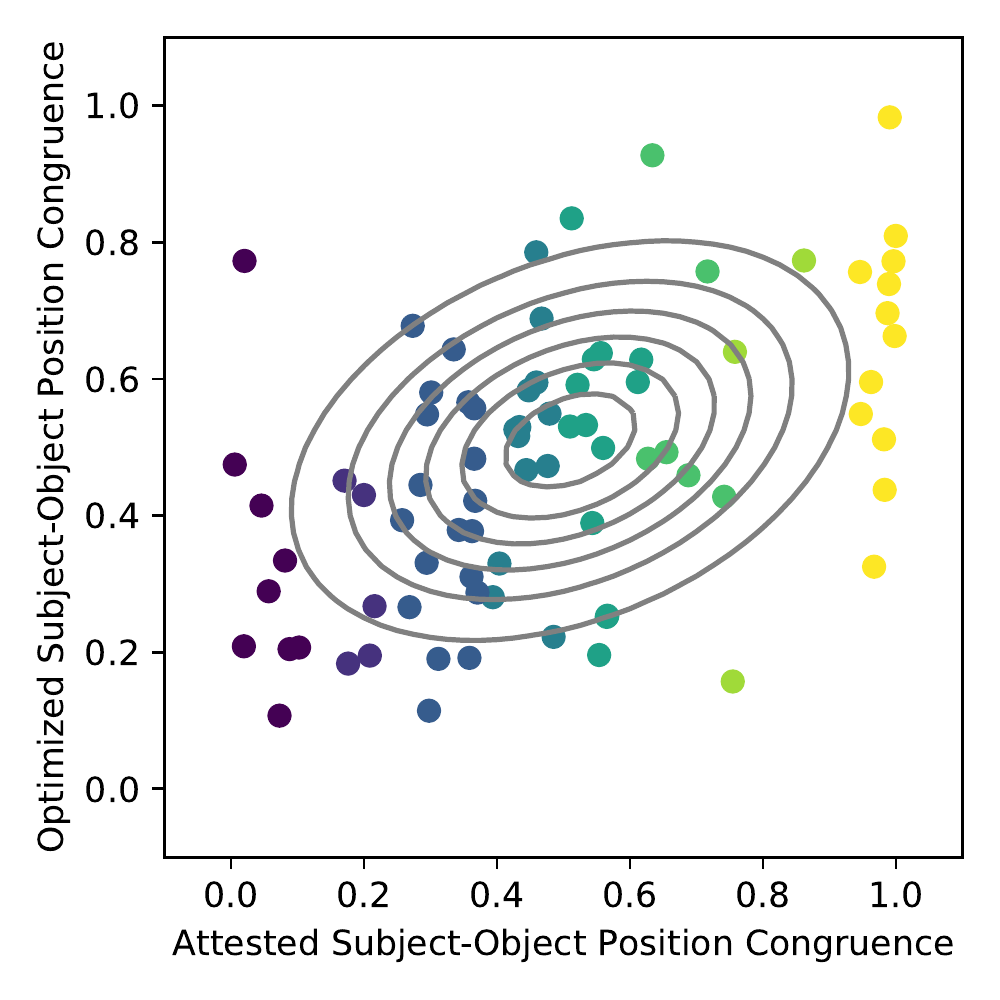} &
    \includegraphics[draft=false,width=0.45\textwidth]{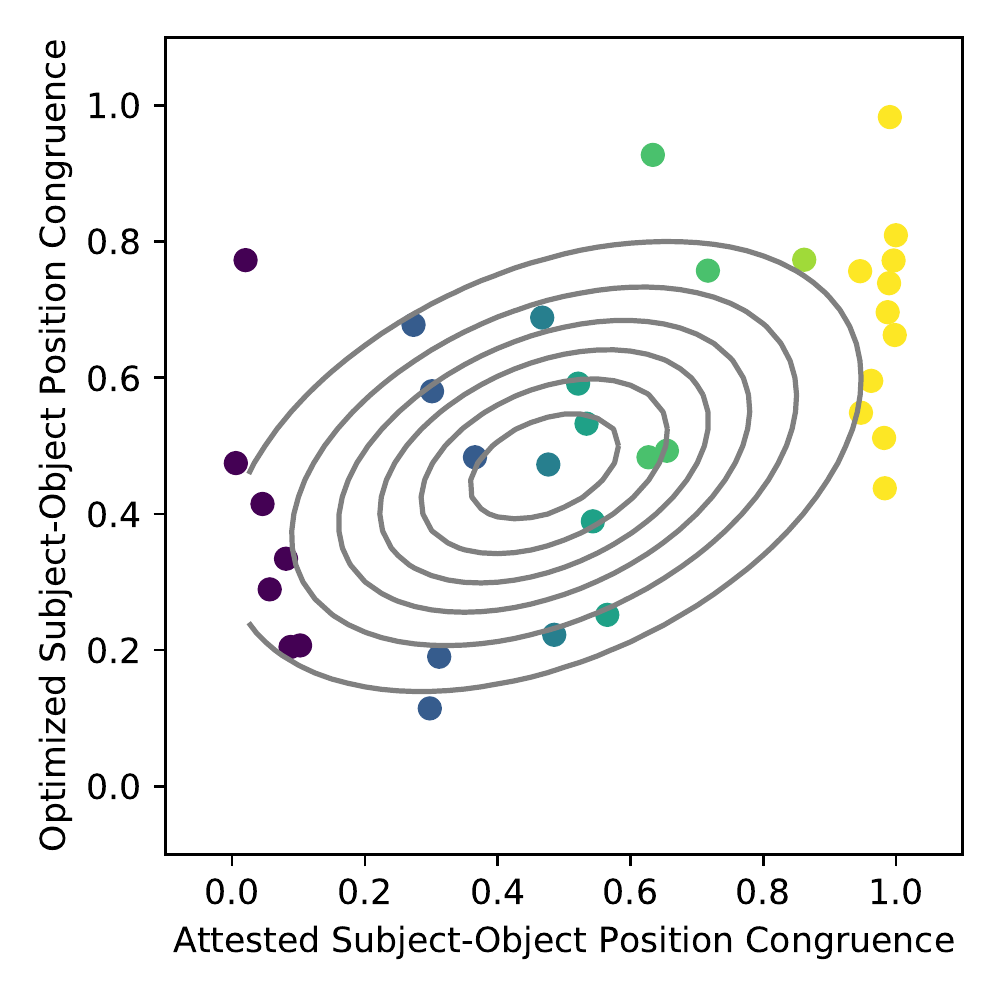} \\
	\end{tabular}
	\caption{Stationary distribution in the plane spanned by optimized and attested subject-object position congruence.}\label{fig:stationary-congruence}
\end{figure}

Figures~\ref{sec:stationary-il-dl} and \ref{fig:stationary-congruence} visualize the stationary distribution, both for all languages and when excluding the Indo-European phylum.

\paragraph{Further Model Versions}
We also considered a version of the model where we explicitly accounted for imprecise measurements due to limitations in corpus data by assuming Gaussian observation noise, i.e., observations are modeled as $\widehat{\xi}_L = \xi_L + \epsilon$, with $\epsilon_i \sim \mathcal{N}(0,\sigma_i)$. 
\footnote{In terms of implementation, this leads to the addition of a diagonal matrix $diag([\sigma_1, \dots, \sigma_4])$ to $Cov(\xi_L, \xi_L)$, and does not affect the other terms of the covariance.}
While the added model complexity did not improve model fit\footnote{The marginal log-likelihood for a model applied to optimized and attested subject-object position congruence without noise is -94; the analogous model with noise has a less favorable marginal likelihood of -96.}, it did not alter the conclusions: Indeed, with this model, the correlations were estimated to be even somewhat larger than without assuming observation noise ($R=-0.54$, 95\% CrI $[-0.79, -0.30]$, $P(R>0) < 0.0001$ for the correlation between DL and congruence; $R=0.61$, 95\% CrI $[0.32, 0.87]$, $P(R<0) < 0.0001$ for the correlation between attested and average congruence).
We also conducted a version of the model where the noise in different dimensions was allowed to be correlated, $\epsilon \sim N(0, T)$ where $T$ has the same prior as the instantaneous covariance matrix $\Sigma$ (see Section~\ref{sec:prior-lkj}).
This is a particularly conservative model, because it allows the correlation in the noise to potentially explain some of the observed correlations.
Nonetheless, correlations continued to be estimated similarly to before ($R=0.42$, 95\% CrI $[0.10, 0.72]$, $P(R<0) = 0.00815$ for attested and optimized subject-object position congruence; $R=-0.52$, 95\% CrI $[-0.73, -0.26]$, $P(R>0) = 0.00005$ for DL and attested subject-object position congruence).

\section{Details for Model and Random Mutations}

We sampled 40 random grammars and 40 approximately optimized grammars, each from one of the 80 languages.
For each grammar, we ran 30 chains, either of 200 random mutations, or of $\approx$ 200 years of evolution under the fitted model.
In Main Paper, Figure 5, we show for each sampled grammar, an arrow from the original point to the mean position at the end of the 30 chains.

For an optimized grammar, random mutations usually deteriorated efficiency: chains of 200 mutations had a 12\% chance of improving IL, and 17\% of improving DL.
In contrast, under the fitted model, change was neutral: 55\% of chains improved IL, 51\% of chains improved DL.
For a baseline grammar, the pattern was in the opposite directions, random mutations were mostly neutral (39 \% chance of improving IL, and 50\% chance of improving DL).
In contrast, under the fitted model, 71\% of chains improved IL, 81\% of chains improved DL.

\section{Comparison with Simple Brownian Model}

The simple Brownian model leaves out the drift term, leading to the stochastic differential equation:
\begin{equation*}
	\operatorname{d}\xi_t = \sqrt{\Sigma} \operatorname{d}B_t
\end{equation*}
This is known as the Independent Contrasts model \citep{felsenstein1973maximum, freckleton2012fast}, and underlies standard phylogenetic regression models \citep{Paradis2004APEAO}.

Brownian motion differs from the Ornstein-Uhlenbeck process in that it does not have a long-term stationary solution.
Instead, trajectories $\xi_t$ tend to move arbitrarily far away from the origin over time $t$.
This is clearly unrealistic in our setting, as subject-object position congruence is bounded between 0 and 1.
As there is no stationary solution, there is no straightforward way to jointly apply the model to data from languages that do not share a common ancestor.
For modelling purposes, we assumed that all families had a common ancestor at some large time $T_0$ in the past.
This modelling assumption corresponds to the assumption of macro-families of very large time-depth.
We considered $T_0$ to be 15,000 BC, 20,000 BC, 50,000 BC, and measured the instantaneous correlation of changes $R$ for each fit.\footnote{The maximum possible $T_0$ can be no later than Proto-Afroasiatic, which we calibrated at 10,000 BC (see Section~\ref{sec:inner-nodes}).}
To evaluate model fit, we compared marginal likelihood of the Brownian model with the Ornstein-Uhlenbeck model, computed using using Stepping Stone Sampling \citep{xie2011improving} with $K=10$ stones.
Note that an assumption of a specific time depth is not necessary for the Ornstein-Uhlenbeck model, as unrelated languages can be modeled as draws from the stationary distribution for that model.

In the absence of a stationary distribution, the Brownian model cannot make statements about whether languages evolve to maintain efficiency.
We therefore only applied this to the attested and optimized subject-object position congruence, not to IL/DL.

\paragraph{Results}
Model fit as measured by marginal likelihood is much weaker than in the Ornstein-Uhlenbeck model, across choices of $T_0$ (Table~\ref{tab:marg-brown}), corresponding to a Bayes factor of about $10^{24}$ in favor of the Ornstein-Uhlenbeck model.
Nonetheless, across different choices of $T_0$, the Brownian model strongly supports a positive correlation $R$ between attested and optimized subject-object position congruence, very similar to the Ornstein-Uhlenbeck analysis; the posterior probability of $R\leq 0$ is $0.00425$ at the best-fitting setup, $T_0=-15,000$.

\begin{table}
	\begin{center}
	
	\begin{tabular}{llllllll}
	Model & Log-Likelihood \\ \hline
	Ornstein-Uhlenbeck & -94 \\
		Lesioned Ornstein-Uhlenbeck (No Coadaptation) & -119 \\ \hline
		Brownian ($T_0 = -100,000$) & -140 \\
	Brownian ($T_0 = -50,000$) & -129 \\
	Brownian ($T_0 = -20,000$) & -117 \\
	Brownian ($T_0 = -15,000$) & -114 \\
\end{tabular}
	\end{center}
	
	\caption{Marginal log-likelihoods for Ornstein-Uhlenbeck and simple Brownian models. Values closer to $0$ indicate better model fit. We ran the Brownian model at different time depths, because it does not have a stationary distribution, necessitating the assumption of a common root node. The lesioned Ornstein-Uhlenbeck model without coadaptation is obtained by constraining the matrix $\Sigma$ (Equation~\ref{eq:sde}) to be diagonal.}
	\label{tab:marg-brown}
\end{table}

\section{Accounting for Areal Convergence}\label{sec:areal}
The model of random walks on phylogenetic trees assume that languages evolve independently once they have split~\citep[e.g.][]{dunn-evolved-2011, maurits2014tracing}.
However, linguistic evolution can include borrowing between geographically neighboring languages \citep[e.g.][]{dryer1989large, bisang1996areal, heine2003on, aikhenvald2007grammars,  kalyan2019problems}.
Fully integrating such borrowing within phylogenetic modeling is an open problem for computational modeling.
Here, we describe a possible modeling approach that explicitly models convergence in linguistic areas, geographic regions in which languages tend to show convergent evolution due to borrowing  \citep[e.g.][]{campbell1986meso, nichols1992linguistic, haspelmath2001the, gijn2017linguistic}, with a proof-of-concept implementation.
We note that there may be other possible approaches, and have to leave a complete investigation of models fully integrating both phylogeny and borrowing to future research.

We propose to model linguistic areas as latent variables defining time- and location-dependent values $\mu(x,t)$ (where $x$ is a point on the surface of the earth and $t$ is a point in time) that languages at time $t$ and place $x$ drift towards.
These values are inferred from the data together with the other parameters of the Ornstein-Uhlenbeck process.
By placing a suitable Gaussian process prior on $\mu(x,t)$, we encourage parameters that smoothly vary over space and time, reflecting the idea that areal convergence between languages depends on their geographic distance.
This approach is related to the model described by \citep{nuismer2015predicting}, who propose to model convergence between species by assuming correlations between the means $\mu$ of different species.
For other approaches to model interactions between species from the bioinformatics literature, see \cite{manceau2016a,drury2016estimating, bartoszek2017using, drury2018an}.

\paragraph{Model}
We model the grammar and usage components of $\mu$ as depending on the language's geographic position and the time a language was spoken.
This models the impact of linguistic areas, and allows this impact to change over time.

We assume that a language $L$ observed at time $t+\Delta$ (e.g. French) developed from a prior state at time $t$ (e.g., Old French) during time $[t, t+\Delta]$ according to the Ornstein-Uhlenbeck SDE
\begin{equation}
    \operatorname{d}\xi_{L,t} = \Gamma \cdot (\xi_{L,t}-\mu_L) \operatorname{d}t + \sqrt{\Sigma} \operatorname{d}B_t
\end{equation}
where $\mu_L$ is defined by the temporal and geographical location of the language $L$.

We placed a Gaussian process prior with a Laplace kernel on $\mu$.
That is, the covariance between $\mu$ at points $x, y$ on the surface of the earth at times $T_1, T_2$ is taken to be
\begin{equation}\label{eq:kernel}
    Cov(\mu_x, \mu_y) = \alpha \cdot \exp\left(-\frac{1}{\rho^2_1} d(x,y) - \frac{1}{\rho_2^2} |T_1-T_2|\right)
\end{equation}
where $d(x,y)$ is the great circle (geodesic) distance between points $x, y$, and $\alpha, \rho>0$ are hyperparameters.
The Laplace kernel is positive-definite with the great-circle distance $d(\cdot, \cdot)$ \citep{feragen2015geodesic} and thus provides a valid covariance for this distance; many other popular kernels like the RBF kernel are not valid for this distance \citep{feragen2015geodesic}.
This prior favors values of $\mu_L$ that vary smoothly over space and time, encoding the idea of linguistic areas.
We placed Gaussian priors with mean $0$ and variance $1$, truncated to positive values, on the hyperparameters $\alpha, \frac{1}{\rho^2}$ of the kernel~(\ref{eq:kernel}).

We extracted locations of languages from the World Atlas of Linguistic Structures \citep{haspelmath2005the}.
For ancestors, we recursively defined their location as the mean of the locations of their immediate children.

Due to substantial computational cost of this model, we applied it only to the main correlation of interest, i.e., the correlation between attested and average optimized subject-object position congruence.
As convergence is slow compared to our other models, we ran MCMC for 40,000 iterations, again discarding the first half as warmup samples.
We used the $\widehat{R}$ statistic and visual inspection of chains to assess model convergence.

\paragraph{Results}
As the mean $\mu$ depends on the geographic position, there is no single stationary distribution.
As described in Section~\ref{sec:instant-corr}, we thus instead consider the correlation component of $\Sigma$, the covariance matrix of instantaneous changes.
The correlation between changes in attested and average optimized subject-object position congruence was estimated at $R=0.44$, 95\% CrI $[0.2, 0.65]$, $P(R<0) = 0.0004$, suggesting that coadaptation is found even when accounting for areal convergence in addition to phylogenetic relations.

\section{The Role of Case Marking}\label{sec:case}

Here, we report details on the analysis of coadaptation when controlling for the presence of case marking.
We do this by fitting an extension of the model that can model different directions of change in languages with and without case marking, and checking whether the analysis continues to provide evidence for coevolution between word order and usage \emph{even beyond} what is captured by correlations of usage and word order with the presence of case marking.

\paragraph{Coding Languages for Case Marking}
We coded languages from our sample for the presence or absence of case marking on the basis of \citet{wals-49}, supplemented with information from the grammatical literature where no information was provided.
We amended the annotation from \citet{wals-49} to include only case marking that distinguishes between subjects and objects; this concerns several modern Celtic and Germanic languages, which have some nominal case marking but do not distinguish subjects and objects (e.g., Swedish and English use -\textit{s} to mark possessives, but do not distinguish nominal subjects and objects.).

We furthermore coded all interior nodes of the phylogenetic tree for case marking based on the linguistic literature.
In some cases, this annotation was unambiguous due to available historical documentation even though no treebank data was available (e.g., Proto-West-Scandinavian was a late form of Old Norse and had case markers).
In many other cases, cognate case markers are unambiguously attested both within and without a group, showing that they were present in the protolanguage (e.g., Proto-Germanic, Proto-Indo-Iranian).
Furthermore, in many protolanguages, case markers are commonly reconstructed based on their presence in different descendant branches (e.g., Proto-Indo-European, Proto-Afroasiatic, Proto-Common-Turkic, Proto-Uralic and Proto-Ugric). 
Case is not unambiguously reconstructed for Proto-Niger-Congo; we verified that both possible parameter settings lead to qualitatively equivalent results (we report results under the assumption that it did not have case, with essentially indistinguishable results for the other cases).

\paragraph{Model of Change conditioned on Case Marking}
Based on the prior literature, we expect that languages without case marking will be biased towards low subject-object position congruence~\citep{vennemann1974explanation}.
To take this into account, we modified the model by conditioning the mean vector $\mu$ on the presence or absence of case in the language $L$.
\begin{equation*}
    \operatorname{d}\xi_{L,t} = \Gamma_{C(L)} \cdot (\xi_{L,t}-\mu_{C(L)}) \operatorname{d}t + \sqrt{\Sigma} \operatorname{d}B_t
\end{equation*}
where $C(L)$ is $1$ if $L$ has case and $0$ else.
We set priors $\mu_{C(L)} \sim N(0,1)$ for both $C(L) = 0$, and $1$.

\begin{figure}

    \centering

	\begin{tabular}{ccccc}
		\textbf{(A)} & \textbf{(B)}\\
		\includegraphics[draft=false,width=0.45\textwidth]{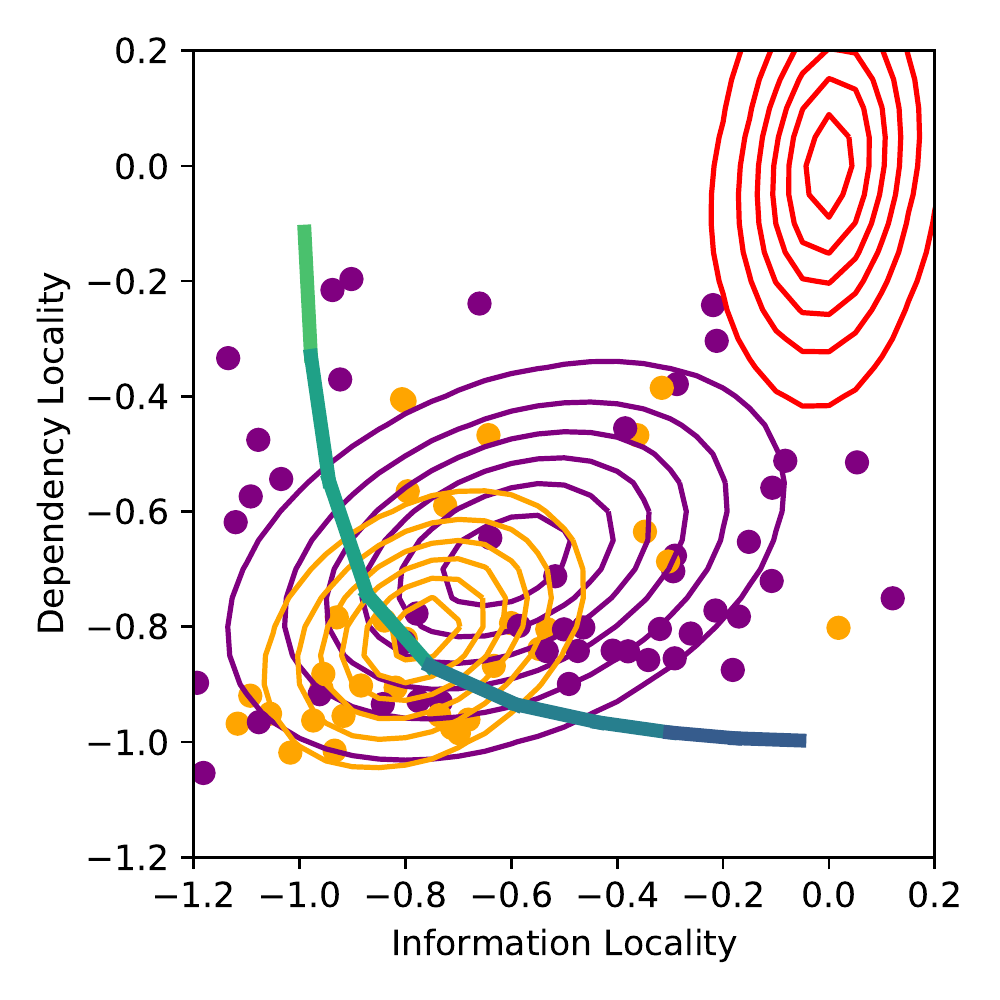}
		 &		\includegraphics[draft=false,width=0.45\textwidth]{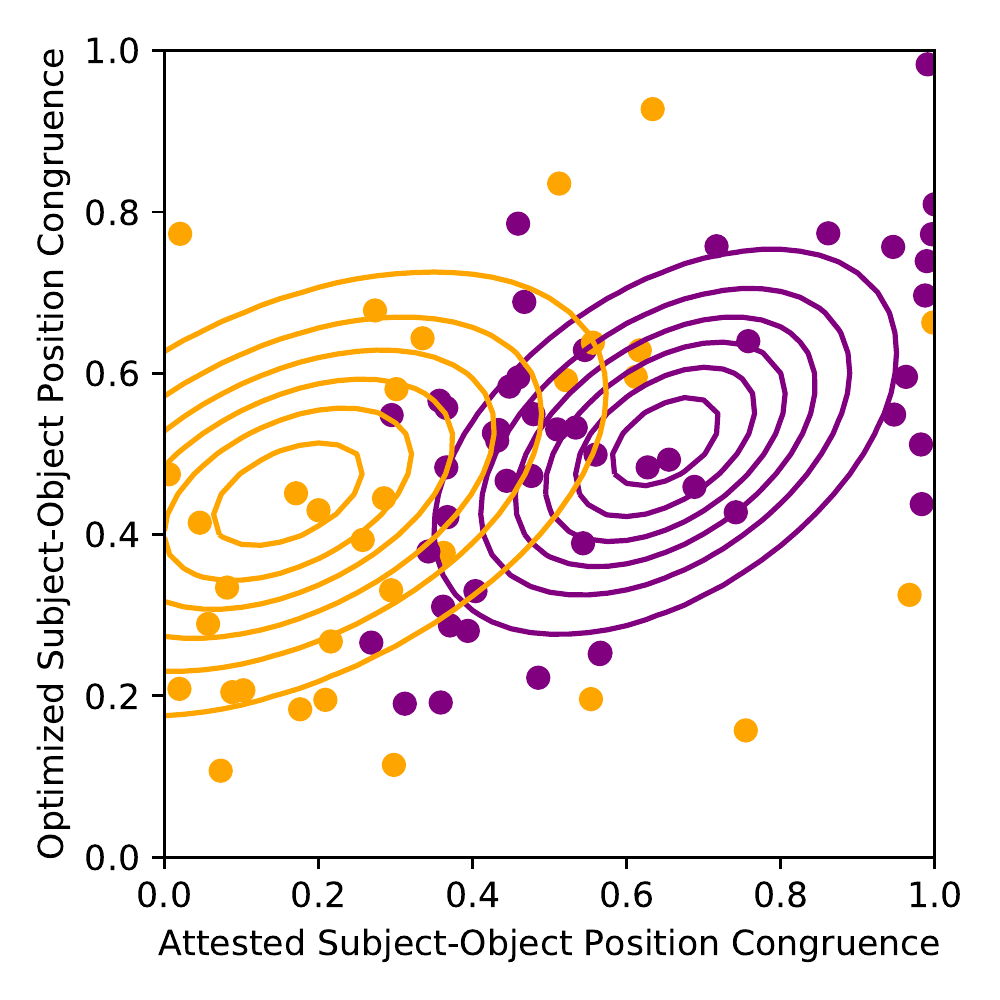} \\
 	\end{tabular}

{\huge{\textcolor[rgb]{1,0.65,0}{---}}} Without Case Marking \phantom{...........} {\huge{\textcolor[rgb]{0.5,0,0.5}{---}}} With Case Marking

    \caption{
	    Fitted stationary distribution, conditioned on case marking.
	    (A) Languages with and without case marking similarly concentrate in the region between the baseline distribution and the Pareto frontier. The difference between the mean values of IL and DL of the two stationary distributions is not statistically meaningful.
	    (B)
	    While the presence of case marking is associated with higher subject-object position congruence ($\mu = 0.65$, 95\% CrI $[0.56, 0.76]$ with case, $\mu=0.16$, 95\% CrI $[-0.07, 0.36]$ without case), coadaptation is predicted even beyond this association, as evidenced by the shape of the two stationary distributions.
	    }
    \label{fig:langs-case}
\end{figure}

\paragraph{Results}
We plot the distribution of languages and the fitted stationary distributions, conditioned on $C(L)$, in Figure~\ref{fig:langs-case}. 
In accordance with the prior literature, the model indicated that languages without case marking favor regions with low observed subject-object position congruence.
For languages with case marking, there was evidence for a bias towards higher subject-object position congruence.
We quantified correlations using the correlation component of the instantaneous changes $\Sigma$ (see Section~\ref{sec:instant-corr}), i.e., the correlation between short-term stochastic changes in different dimensions.
Similarly to the results in the main analysis, there was a negative correlation between DL and subject-object position congruence ($R=-0.39$, 95\% CrI $[-0.60, -0.18]$, $P(R\geq 0) < 0.0001$), and a positive correlation between attested congruence and average optimized congruence along the Pareto frontier ($R=0.49$, 95\% CrI $[0.29, 0.67]$, $P(R\leq 0) < 0.0001$).
This shows that languages show coadaptation between usage and grammar in basic word order, even beyond an association with case marking.

\paragraph{Conclusion}
We found that, while case marking has a robust impact on subject-object position congruence, coadaptation continues to hold when controlling for this.

\part{Additional Analyses}

\section{Optimized and Baseline Grammars}\label{sec:per-lang-rand}
\begin{figure}
\begin{longtable}{ccccccccccccccc}
{\tiny Afrikaans} & {\tiny Akkadian} & {\tiny Amharic} & {\tiny Ancient Greek} & {\tiny Arabic} & {\tiny Armenian}
\\
\includegraphics[draft=false,width=0.07\textwidth]{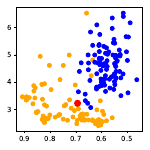} &
\includegraphics[draft=false,width=0.07\textwidth]{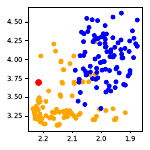} &
\includegraphics[draft=false,width=0.07\textwidth]{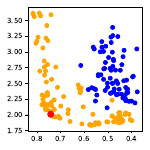} &
\includegraphics[draft=false,width=0.07\textwidth]{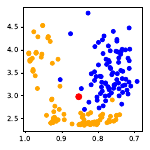} &
\includegraphics[draft=false,width=0.07\textwidth]{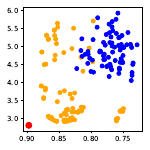} &
\includegraphics[draft=false,width=0.07\textwidth]{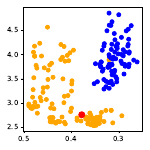} &
\\
{\tiny Bambara} & {\tiny Basque} & {\tiny Belarusian} & {\tiny Breton} & {\tiny Bulgarian} & {\tiny Buryat}
\\
\includegraphics[draft=false,width=0.07\textwidth]{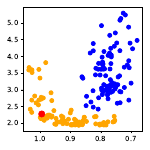} &
\includegraphics[draft=false,width=0.07\textwidth]{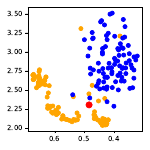} &
\includegraphics[draft=false,width=0.07\textwidth]{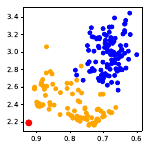} &
\includegraphics[draft=false,width=0.07\textwidth]{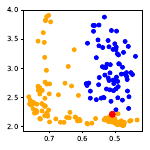} &
\includegraphics[draft=false,width=0.07\textwidth]{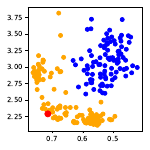} &
\includegraphics[draft=false,width=0.07\textwidth]{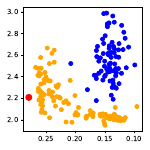} &
\\
{\tiny Cantonese} & {\tiny Catalan} & {\tiny Chinese} & {\tiny Classical Chinese} & {\tiny Coptic} & {\tiny Croatian}
\\
\includegraphics[draft=false,width=0.07\textwidth]{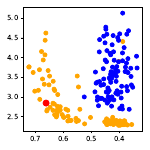} &
\includegraphics[draft=false,width=0.07\textwidth]{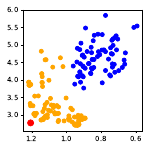} &
\includegraphics[draft=false,width=0.07\textwidth]{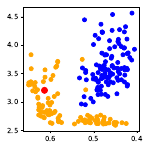} &
\includegraphics[draft=false,width=0.07\textwidth]{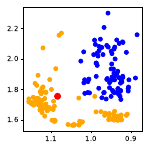} &
\includegraphics[draft=false,width=0.07\textwidth]{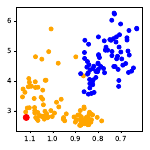} &
\includegraphics[draft=false,width=0.07\textwidth]{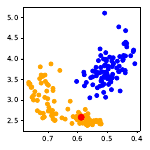} &
\\
{\tiny Czech} & {\tiny Danish} & {\tiny Dutch} & {\tiny English} & {\tiny Erzya} & {\tiny Estonian}
\\
\includegraphics[draft=false,width=0.07\textwidth]{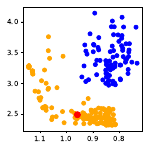} &
\includegraphics[draft=false,width=0.07\textwidth]{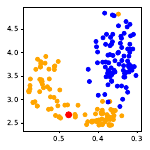} &
\includegraphics[draft=false,width=0.07\textwidth]{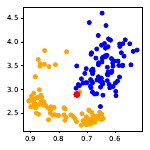} &
\includegraphics[draft=false,width=0.07\textwidth]{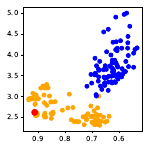} &
\includegraphics[draft=false,width=0.07\textwidth]{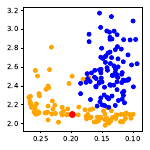} &
\includegraphics[draft=false,width=0.07\textwidth]{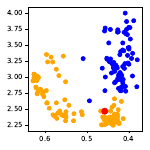} &
\\
{\tiny Faroese} & {\tiny Finnish} & {\tiny French} & {\tiny Galician} & {\tiny German} & {\tiny Gothic}
\\
\includegraphics[draft=false,width=0.07\textwidth]{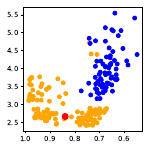} &
\includegraphics[draft=false,width=0.07\textwidth]{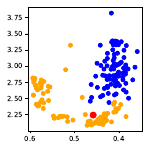} &
\includegraphics[draft=false,width=0.07\textwidth]{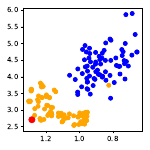} &
\includegraphics[draft=false,width=0.07\textwidth]{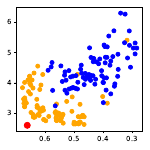} &
\includegraphics[draft=false,width=0.07\textwidth]{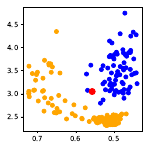} &
\includegraphics[draft=false,width=0.07\textwidth]{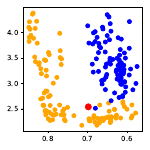} &
\\
{\tiny Greek} & {\tiny Hebrew} & {\tiny Hindi} & {\tiny Hungarian} & {\tiny Icelandic} & {\tiny Indonesian}
\\
\includegraphics[draft=false,width=0.07\textwidth]{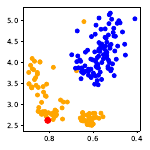} &
\includegraphics[draft=false,width=0.07\textwidth]{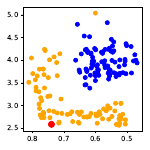} &
\includegraphics[draft=false,width=0.07\textwidth]{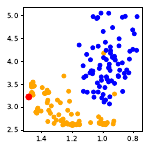} &
\includegraphics[draft=false,width=0.07\textwidth]{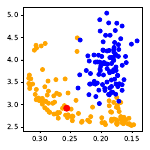} &
\includegraphics[draft=false,width=0.07\textwidth]{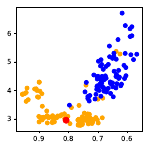} &
\includegraphics[draft=false,width=0.07\textwidth]{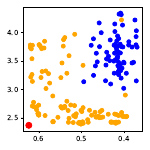} &
\\
{\tiny Irish} & {\tiny Italian} & {\tiny Japanese} & {\tiny Kazakh} & {\tiny Kiche} & {\tiny Komi Zyrian}
\\
\includegraphics[draft=false,width=0.07\textwidth]{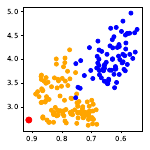} &
\includegraphics[draft=false,width=0.07\textwidth]{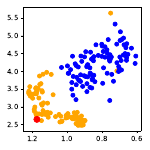} &
\includegraphics[draft=false,width=0.07\textwidth]{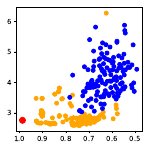} &
\includegraphics[draft=false,width=0.07\textwidth]{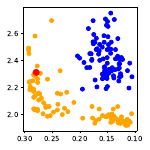} &
\includegraphics[draft=false,width=0.07\textwidth]{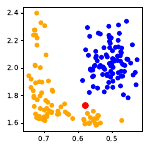} &
\includegraphics[draft=false,width=0.07\textwidth]{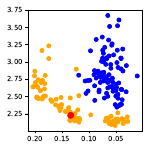} &
\\
{\tiny Korean} & {\tiny Kurmanji} & {\tiny Latin} & {\tiny Latvian} & {\tiny Lithuanian} & {\tiny Maltese}
\\
\includegraphics[draft=false,width=0.07\textwidth]{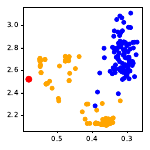} &
\includegraphics[draft=false,width=0.07\textwidth]{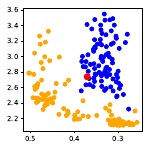} &
\includegraphics[draft=false,width=0.07\textwidth]{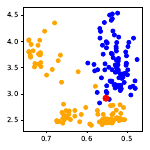} &
\includegraphics[draft=false,width=0.07\textwidth]{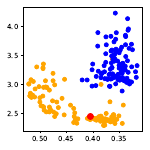} &
\includegraphics[draft=false,width=0.07\textwidth]{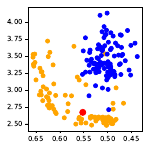} &
\includegraphics[draft=false,width=0.07\textwidth]{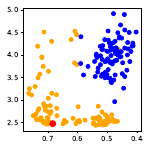} &
\\
{\tiny Manx} & {\tiny Mbya Guarani} & {\tiny Naija} & {\tiny North Sami} & {\tiny Norwegian} & {\tiny Old Church Slavonic}
\\
\includegraphics[draft=false,width=0.07\textwidth]{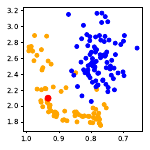} &
\includegraphics[draft=false,width=0.07\textwidth]{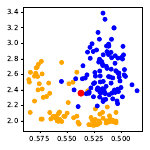} &
\includegraphics[draft=false,width=0.07\textwidth]{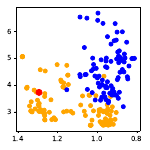} &
\includegraphics[draft=false,width=0.07\textwidth]{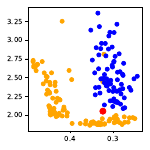} &
\includegraphics[draft=false,width=0.07\textwidth]{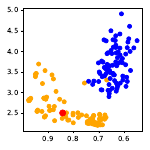} &
\includegraphics[draft=false,width=0.07\textwidth]{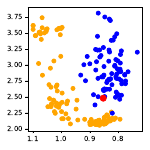} &
\\
{\tiny Old East Slavic} & {\tiny Old French} & {\tiny Persian} & {\tiny Polish} & {\tiny Portuguese} & {\tiny Romanian}
\\
\includegraphics[draft=false,width=0.07\textwidth]{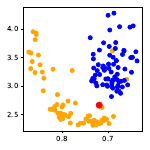} &
\includegraphics[draft=false,width=0.07\textwidth]{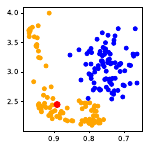} &
\includegraphics[draft=false,width=0.07\textwidth]{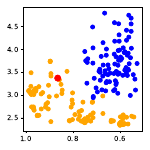} &
\includegraphics[draft=false,width=0.07\textwidth]{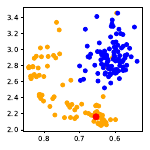} &
\includegraphics[draft=false,width=0.07\textwidth]{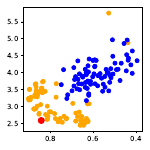} &
\includegraphics[draft=false,width=0.07\textwidth]{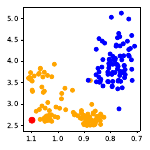} &
\\
{\tiny Russian} & {\tiny Sanskrit} & {\tiny Scottish Gaelic} & {\tiny Serbian} & {\tiny Slovak} & {\tiny Slovenian}
\\
\includegraphics[draft=false,width=0.07\textwidth]{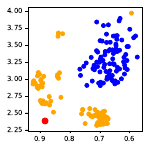} &
\includegraphics[draft=false,width=0.07\textwidth]{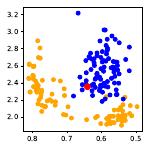} &
\includegraphics[draft=false,width=0.07\textwidth]{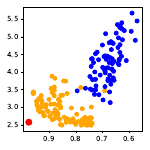} &
\includegraphics[draft=false,width=0.07\textwidth]{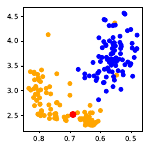} &
\includegraphics[draft=false,width=0.07\textwidth]{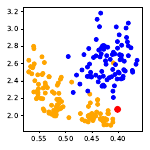} &
\includegraphics[draft=false,width=0.07\textwidth]{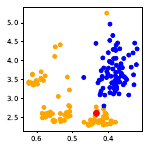} &
\\
{\tiny Spanish} & {\tiny Swedish} & {\tiny Tamil} & {\tiny Thai} & {\tiny Turkish} & {\tiny Ukrainian}
\\
\includegraphics[draft=false,width=0.07\textwidth]{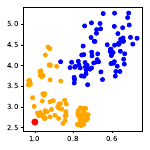} &
\includegraphics[draft=false,width=0.07\textwidth]{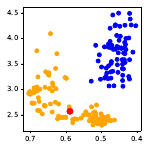} &
\includegraphics[draft=false,width=0.07\textwidth]{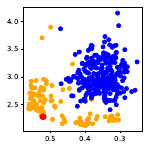} &
\includegraphics[draft=false,width=0.07\textwidth]{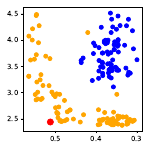} &
\includegraphics[draft=false,width=0.07\textwidth]{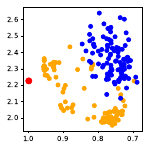} &
\includegraphics[draft=false,width=0.07\textwidth]{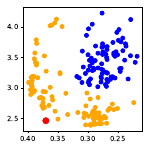} &
\\
{\tiny Upper Sorbian} & {\tiny Urdu} & {\tiny Uyghur} & {\tiny Vietnamese} & {\tiny Welsh} & {\tiny Western Armenian}
\\
\includegraphics[draft=false,width=0.07\textwidth]{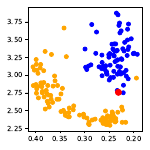} &
\includegraphics[draft=false,width=0.07\textwidth]{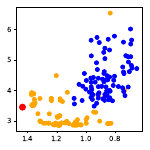} &
\includegraphics[draft=false,width=0.07\textwidth]{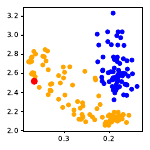} &
\includegraphics[draft=false,width=0.07\textwidth]{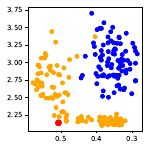} &
\includegraphics[draft=false,width=0.07\textwidth]{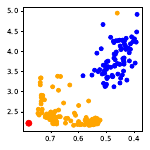} &
\includegraphics[draft=false,width=0.07\textwidth]{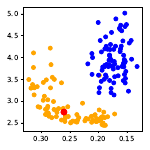} &
\\
{\tiny Wolof} & {\tiny Xibe}
\\
\includegraphics[draft=false,width=0.07\textwidth]{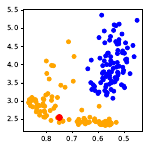} &
\includegraphics[draft=false,width=0.07\textwidth]{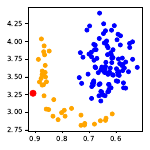} &
\\
\end{longtable}
	\caption{Position of optimized (orange) and baseline (blue) grammars for each of the 80 languages. Red dots indicate the attested ordering.}\label{fig:per-lang-rand}
\end{figure}

In Figure~\ref{fig:per-lang-rand}, we show the position of optimized (orange) and baseline (blue) grammars for each of the 80 languages.
Optimized grammars inhabit the area between the baseline grammars and the Pareto frontier.
Compare Section~\ref{sec:per-lang} for results by subject-object position congruence.

\newpage
\section{Within-Language Correlates of Basic Word Order}

Here, we show that basic word order reflects optimization for Dependency Length Minimization (DLM) not only on the level of languages, but also on the level of individual sentences.

In many SVO languages, certain intransitive subjects can appear after the verb (``along came a dog'').
This kind of ``intransitive inversion'' has been documented in many SVO languages, including English, Romance languages, and Chinese \citep[Chapter 17.2]{li1981mandarin}.
There are also languages whose basic word order is different in transitive and in intransitive clauses \citep{wals-82}; the World Atlas of Language Structures lists 13 languages with transitive SVO and intransitive VS basic word order \citep{wals-81,wals-82}, while it lists no languages with transitive VSO and intransitive SV order.
This observation has been formalized as the following language universal: \textit{If VS is dominant with transitives, it is also dominant with intransitives} (\citet[No 344]{plank2000the}, citing \citet{kozinsky1981Nekotorye}).
DLM provides an explanation for this universal.
We conjectured that, more generally, the rate of VS order is higher when no object is present than when an object is present.
For each language in our dataset, we collected statistics for all verbs with a subject and conducted the following logistic analysis:

\begin{equation}
\text{SV Order} \sim \text{Object is present}
\end{equation}

A positive effect indicates that presence of an object makes SV order more likely, compared to VS order.
Results are shown in the table below.
As predicted, in most languages where there is variation between SV and VS order, a significant positive effect was observed.

In some predominant VSO languages, SVO is an alternative word order in unembedded clauses, whereas embedded clauses tend to only allow VSO.
This is in accordance with the predictions of DLM, which favors high subject-object position congruence in embedded clauses (see Figure 1B in the main paper).
Examples include relative clauses in Afroasiatic and Celtic (Standard Arabic~\citep{alqurashi:2012}, Breton \citep[][p. 80]{timm1988relative}, Ancient Egyptian \citep{gardiner1957egyptian}, Tuareg \citep[Chapter 12.1.2]{heath2005a}).
Conversely, in some SVO languages, embedded clauses show VSO order (Bantu, \citet{demuth1999verb}); Miza (Chadic) has SVO/VOS in main clauses and VOS in embedded clauses~\citep{wals-81}.
However, it is not generally true that VS order is more common in embedded clauses across all languages that have variation in basic word order.
For instance, German and Dutch can have VS in main clauses, but are almost always SV in subordinate clauses; the same holds for Quileute (Chimakuan)~\citep{wals-81}.

\begin{longtable}{l|lllllll}
	\multicolumn{4}{p{0.8\textwidth}}{Coefficients in logistic analysis regressing SV/VS Order based on the presence of an object. `SV Frequency' indicates the overall rate of SV order (as opposed to VS) in the language. A positive coefficient ($\beta > 0$) indicates that SV is more common in the presence of an object than when there is no object.}
	\\
		\multicolumn{4}{p{0.8\textwidth}}{}\\
Language & SV Frequency & $\beta$ & $p$ \\ \hline
Afrikaans  &  0.989  &  -0.12  &  0.6568 \\ 
Akkadian  &  0.98  &  1.56  &  0.0449 \\ 
Amharic  &  0.665  &  -0.38  &  0.0011 \\ 
Ancient Greek  &  0.786  &  0.31  &  $<$ 0.00001 \\ 
Arabic  &  0.492  &  0.55  &  $<$ 0.00001 \\ 
Armenian  &  0.89  &  0.83  &  $<$ 0.00001 \\ 
Bambara  &  0.999  &  16.68  &  0.9948 \\ 
Basque  &  0.872  &  -0.11  &  0.0811 \\ 
Belarusian  &  0.773  &  1.24  &  $<$ 0.00001 \\ 
Breton  &  0.541  &  0.1  &  0.6465 \\ 
Bulgarian  &  0.813  &  1.29  &  $<$ 0.00001 \\ 
Buryat  &  0.996  &  -0.42  &  0.731 \\ 
Cantonese  &  0.994  &  1.12  &  0.3061 \\ 
Catalan  &  0.932  &  0.07  &  0.0502 \\ 
Chinese  &  0.999  &  17.7  &  0.985 \\ 
Classical Chinese  &  0.999  &  17.51  &  0.9852 \\ 
Coptic  &  0.922  &  18.18  &  0.9581 \\ 
Croatian  &  0.827  &  0.99  &  $<$ 0.00001 \\ 
Czech  &  0.733  &  0.47  &  $<$ 0.00001 \\ 
Danish  &  0.865  &  0.62  &  $<$ 0.00001 \\ 
Dutch  &  0.813  &  0.42  &  $<$ 0.00001 \\ 
English  &  0.962  &  3.49  &  $<$ 0.00001 \\ 
Erzya  &  0.677  &  0.94  &  $<$ 0.00001 \\ 
Estonian  &  0.737  &  0.49  &  $<$ 0.00001 \\ 
Faroese  &  0.854  &  -0.16  &  0.0559 \\ 
Finnish  &  0.867  &  1.23  &  $<$ 0.00001 \\ 
French  &  0.957  &  1.17  &  $<$ 0.00001 \\ 
Galician  &  0.877  &  1.09  &  $<$ 0.00001 \\ 
German  &  0.843  &  0.63  &  $<$ 0.00001 \\ 
Gothic  &  0.733  &  0.6  &  $<$ 0.00001 \\ 
Greek  &  0.839  &  0.65  &  $<$ 0.00001 \\ 
Hebrew  &  0.692  &  0.93  &  $<$ 0.00001 \\ 
Hindi  &  0.996  &  2.03  &  $<$ 0.00001 \\ 
Hungarian  &  0.81  &  0.77  &  $<$ 0.00001 \\ 
Icelandic  &  0.75  &  0.2  &  $<$ 0.00001 \\ 
Indonesian  &  0.943  &  4.54  &  $<$ 0.00001 \\ 
Irish  &  0.162  &  0.45  &  $<$ 0.00001 \\ 
Italian  &  0.821  &  1.76  &  $<$ 0.00001 \\ 
Japanese  &  1  &  15.4  &  0.9952 \\ 
Kazakh  &  0.992  &  -0.36  &  0.6167 \\ 
Kiche  &  0.474  &  0.97  &  $<$ 0.00001 \\ 
Komi Zyrian  &  0.762  &  1  &  3e-04 \\ 
Korean  &  1  &  15.18  &  0.9953 \\ 
Kurmanji  &  0.997  &  16.57  &  0.9948 \\ 
Latin  &  0.833  &  0.56  &  $<$ 0.00001 \\ 
Latvian  &  0.79  &  0.76  &  $<$ 0.00001 \\ 
Lithuanian  &  0.785  &  0.33  &  9e-04 \\ 
Maltese  &  0.731  &  2.34  &  $<$ 0.00001 \\ 
Manx  &  0.001  &  -16.59  &  0.9969 \\ 
Mbya Guarani  &  0.866  &  1.73  &  0.0929 \\ 
Naija  &  0.982  &  17.81  &  0.959 \\ 
North Sami  &  0.799  &  1.91  &  $<$ 0.00001 \\ 
Norwegian  &  0.837  &  0.85  &  $<$ 0.00001 \\ 
Old Church Slavonic  &  0.686  &  0.76  &  $<$ 0.00001 \\ 
Old East Slavic  &  0.661  &  0.38  &  $<$ 0.00001 \\ 
Old French  &  0.861  &  0.78  &  $<$ 0.00001 \\ 
Persian  &  0.999  &  0.22  &  0.462 \\ 
Polish  &  0.756  &  0.83  &  $<$ 0.00001 \\ 
Portuguese  &  0.909  &  2.14  &  $<$ 0.00001 \\ 
Romanian  &  0.74  &  0.58  &  $<$ 0.00001 \\ 
Russian  &  0.772  &  1.09  &  $<$ 0.00001 \\ 
Sanskrit  &  0.893  &  0.35  &  0.0091 \\ 
Scottish Gaelic  &  0.013  &  -0.55  &  0.2019 \\ 
Serbian  &  0.801  &  1.36  &  $<$ 0.00001 \\ 
Slovak  &  0.724  &  0.71  &  $<$ 0.00001 \\ 
Slovenian  &  0.778  &  0.37  &  $<$ 0.00001 \\ 
Spanish  &  0.849  &  1.28  &  $<$ 0.00001 \\ 
Swedish  &  0.865  &  0.72  &  $<$ 0.00001 \\ 
Tamil  &  0.987  &  1.71  &  0.1 \\ 
Thai  &  0.999  &  17.45  &  0.9967 \\ 
Turkish  &  0.972  &  -0.25  &  $<$ 0.0001 \\ 
Ukrainian  &  0.806  &  1.05  &  $<$ 0.00001 \\ 
Upper Sorbian  &  0.799  &  0.84  &  5e-04 \\ 
Urdu  &  0.996  &  1.24  &  0.0038 \\ 
Uyghur  &  0.961  &  2.69  &  $<$ 0.00001 \\ 
Vietnamese  &  0.989  &  0.76  &  0.0209 \\ 
Welsh  &  0.053  &  1.21  &  0.0011 \\ 
Western Armenian  &  0.915  &  0.81  &  $<$ 0.0001 \\ 
Wolof  &  0.999  &  16.7  &  0.9914 \\ 
\end{longtable}

\section{Coexpression of Subjects and Objects}
\begin{figure}
    \centering
	\begin{tabular}{cc}
		All Languages & Excluding Indo-European \\
		\includegraphics[draft=false, width=0.4\textwidth]{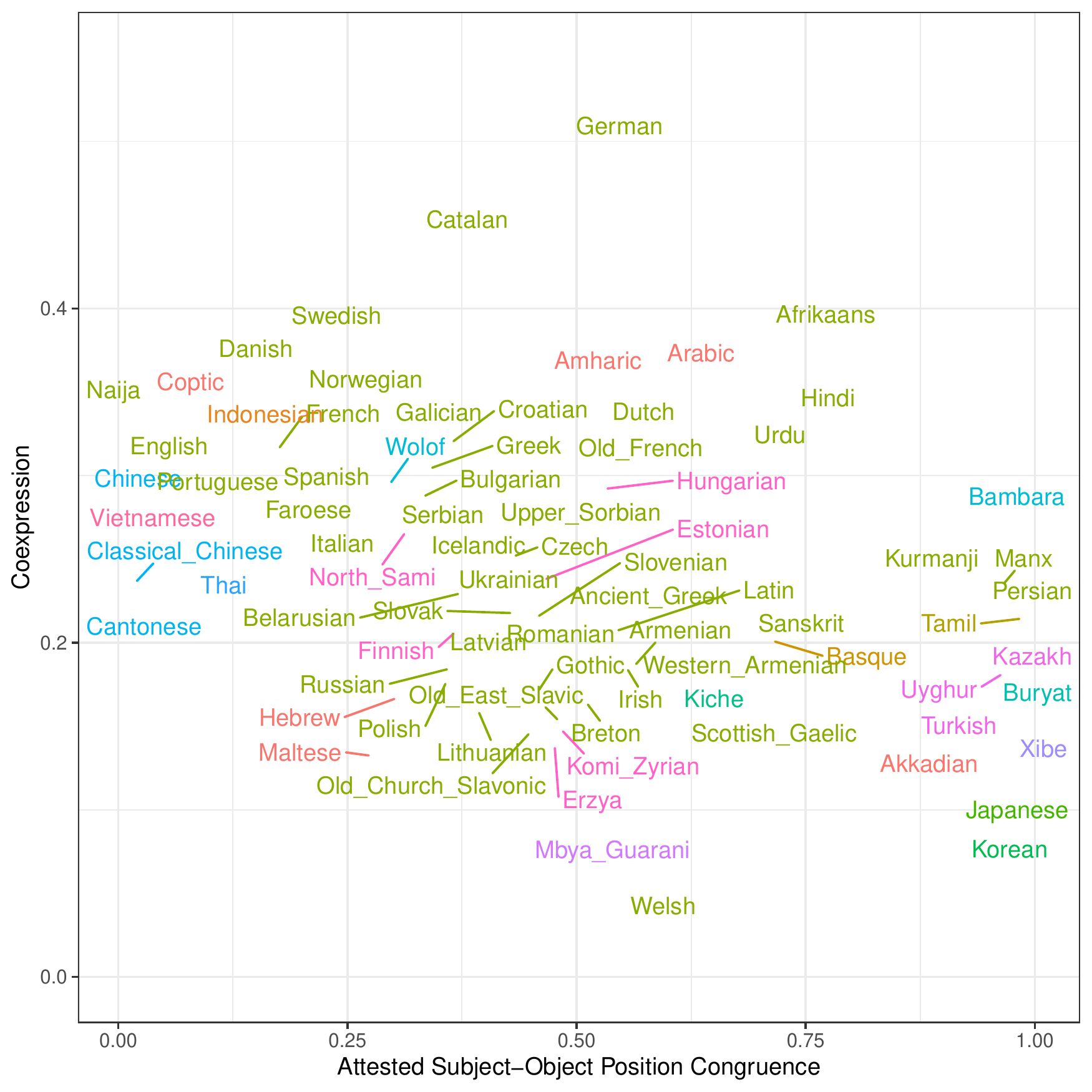} &
   \includegraphics[draft=false, width=0.4\textwidth]{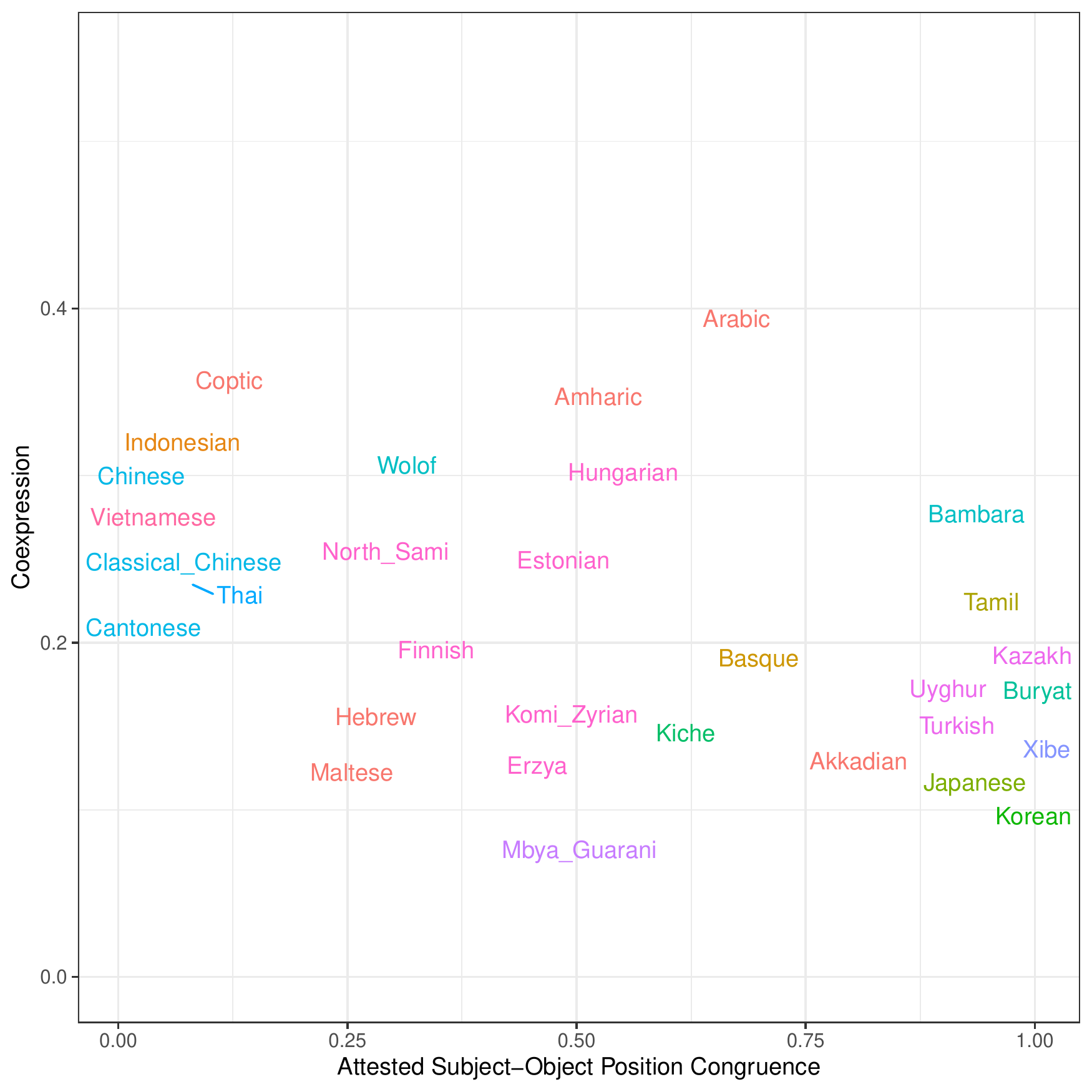} \\
		$\beta=-0.11$, 95\% CrI $[-0.20, -0.03]$ & $\beta=-0.1$, 95\% CrI $[-0.2, 0]$
	\end{tabular}
	\caption{Comparison of attested subject-object position congruence (x-axis) and the fraction of verbs that simultaneously express a subject and an object, among those verbs expressing at least one (``coexpression'', y-axis). Attested subject-object position congruence predicts coexpression in a linear mixed-effects regression with per-family intercept and slope ($\beta=-0.11$, $SE=0.04$, $95\%$ CrI $[-0.20, -0.03]$, $\operatorname{P}(\beta>0) = 0.006$).}
    \label{fig:study2}
\end{figure}

\begin{figure}
    \centering
	\begin{tabular}{cc}
		All Languages & Excluding Indo-European \\
		\includegraphics[draft=false, width=0.4\textwidth]{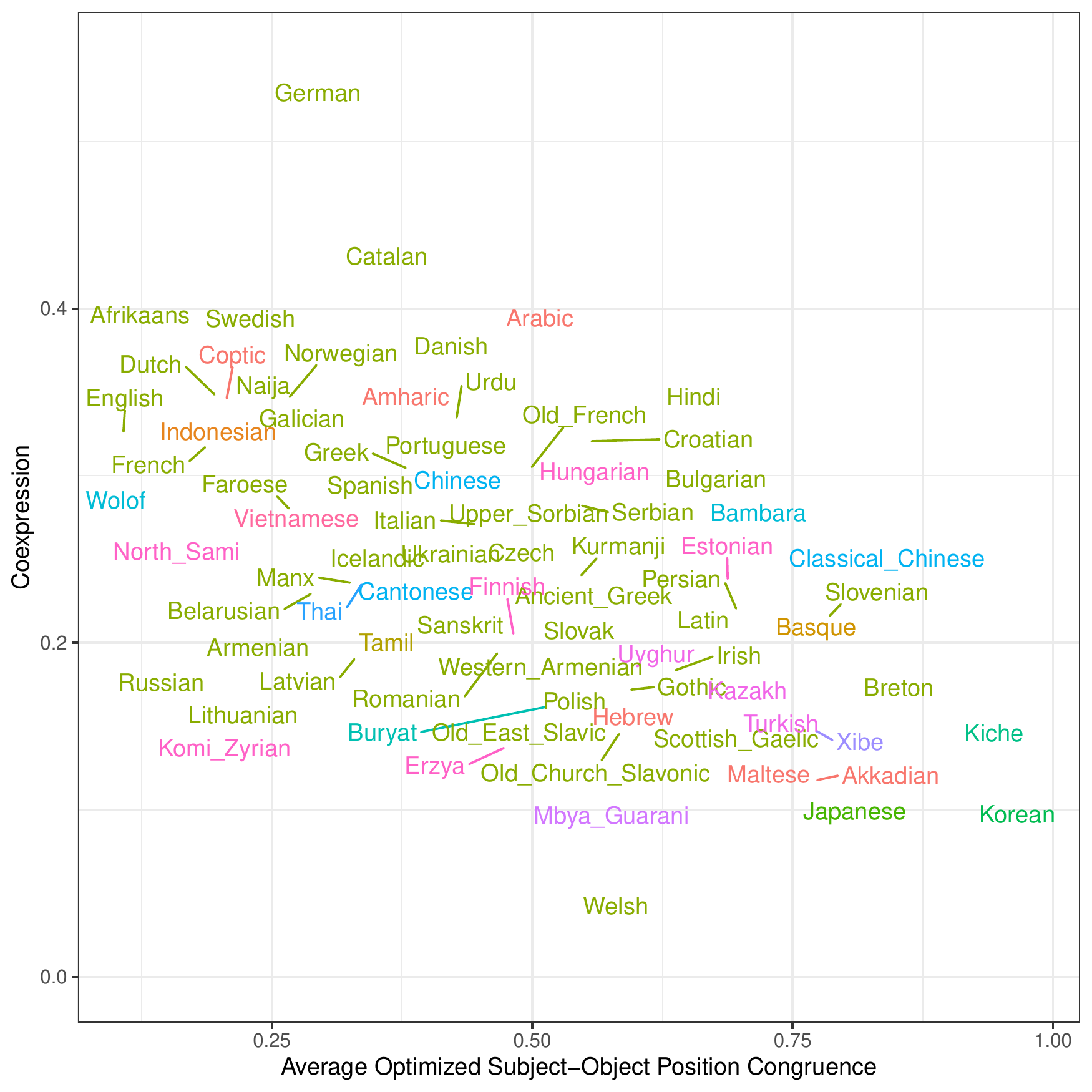} &
   \includegraphics[draft=false, width=0.4\textwidth]{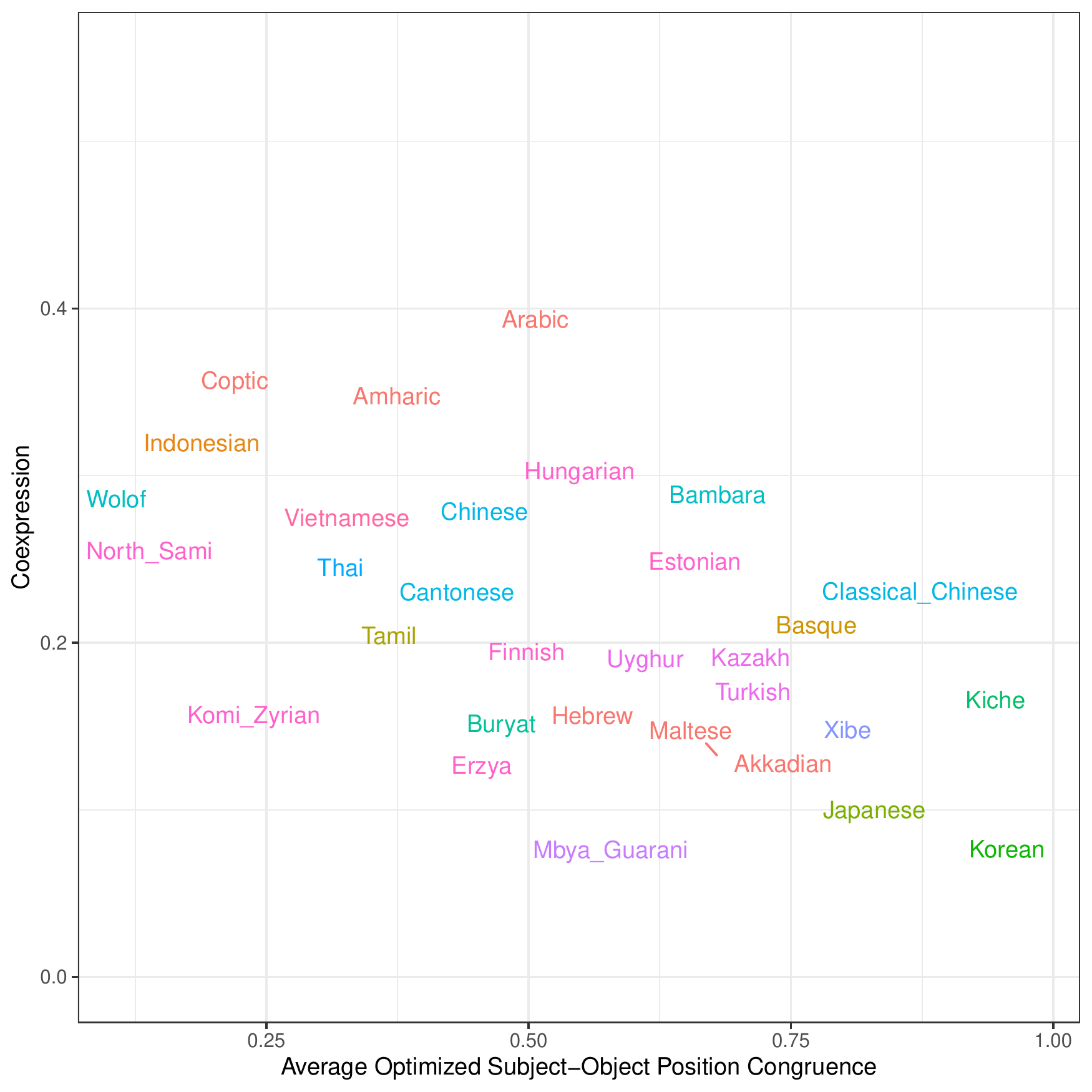} \\
		$\beta=-0.23$, $95\%$ CrI $[-0.33,  -0.13]$ & $\beta=-0.2$, $95\%$ CrI $[-0.33, -0.07]$ \\
	\end{tabular}
	\caption{Comparison of average subject-object position congruence along the Pareto frontier (x-axis) and the fraction of verbs that simultaneously express a subject and an object,  among those verbs expressing at least one (``coexpression'', y-axis). We show coefficients and Bayesian coefficients of determination in a linear mixed-effects regression with per-family intercept and slope.}
    \label{fig:study2b}
\end{figure}

In Figure~\ref{fig:study2}, we show attested subject-object congruence together with the fraction of verbs that simultaneously express a subject and an object among those verbs expressing at least one, for each language.
In Figure~\ref{fig:study2b}, we compare this fraction with the average subject-object position congruence along the Pareto frontier.

\section{Details for Mixed-Effects Analyses}

\paragraph{Priors}

We conducted standard Bayesian linear mixed-effects regressions \citep{gelman1995bayesian} where the response $y_i$ belonging to language $i$ is given by
\begin{equation}
	y_i = (\alpha + \alpha_{f_i}) + (\beta + \beta_{f_i}) x_i + \epsilon_i
\end{equation}
where $x_i$ is the predictor (e.g., attested subject-object position congruence), $f_i$ is the family of language $i$, $\epsilon_i \sim \mathcal{N}(0,\sigma^2)$, and $\alpha_{f_i}$ and $\beta_{f_i}$ are per-family adjustments to the intercept $\alpha$ and the slope $\beta$ respectively.

As described in the main paper, we assumed the prior $N(0,1)$ for the fixed effects slopes, $N(0.5,1)$ for the intercepts, weakly informative Student's $t$ priors ($\nu=3$ degrees of freedom, location $0$, and scale $\sigma=2.5$) for the standard deviations of the residuals and the random effects, and an LKJ(1) prior \citep{lewandowski2009generating} for the correlation matrix of random effects.

\subsection{Insensitivity to Priors}

A potential concern is that, because our dataset includes many families represented by only one or a few languages, the mixed-effects model might suffer from inflated estimates of the variance components, as the slopes cannot be individually estimated for those families.

We repeated the analysis predicting attested subject-object position congruence from optimized subject-object position congruence with several more strongly regularizing priors on the variance components.

\begin{figure}
	\begin{tabular}{c||ccccc}
		Prior & $\beta$ & SD $\tau$ of Random Slope & Response SD $\sigma$ \\  \hline\hline
		$\sigma, \tau \sim t(3,0,2.5)$ &
{	\includegraphics[width=0.2\textwidth]{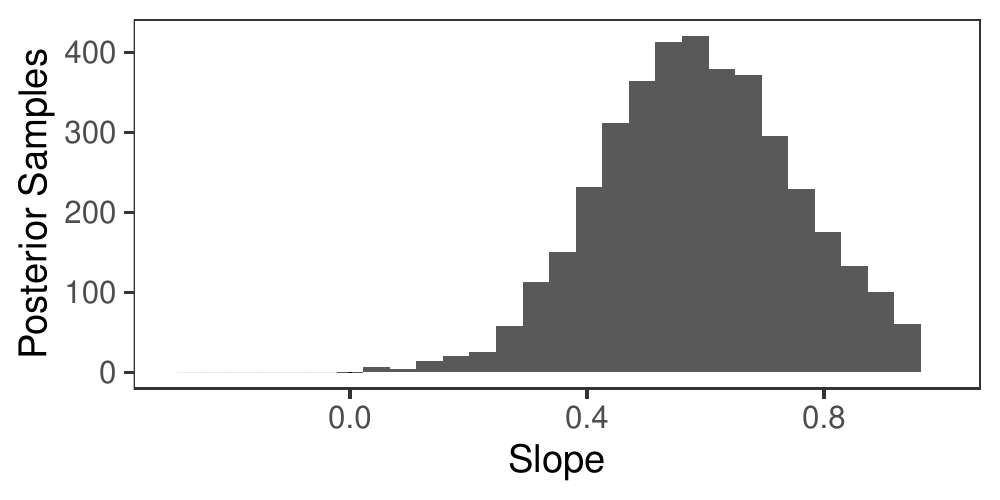}} &
{\includegraphics[width=0.2\textwidth]{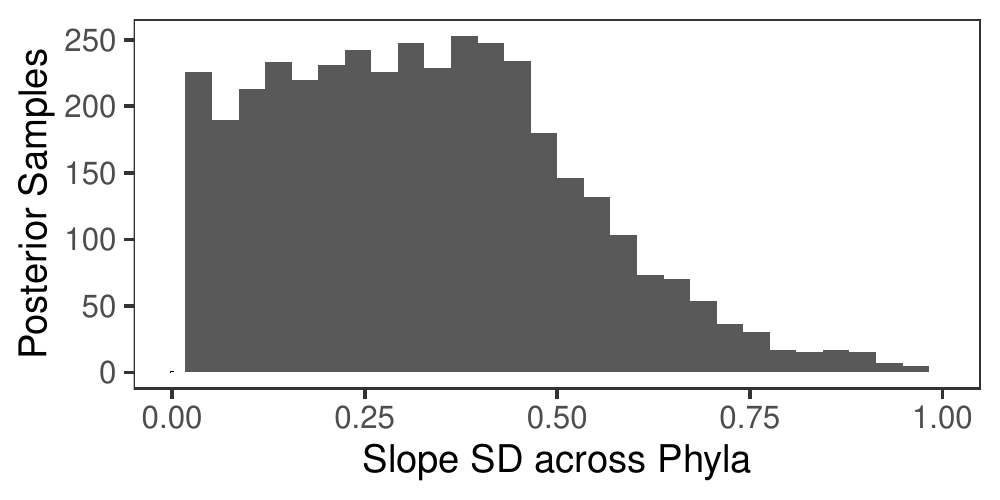}} &
{\includegraphics[width=0.2\textwidth]{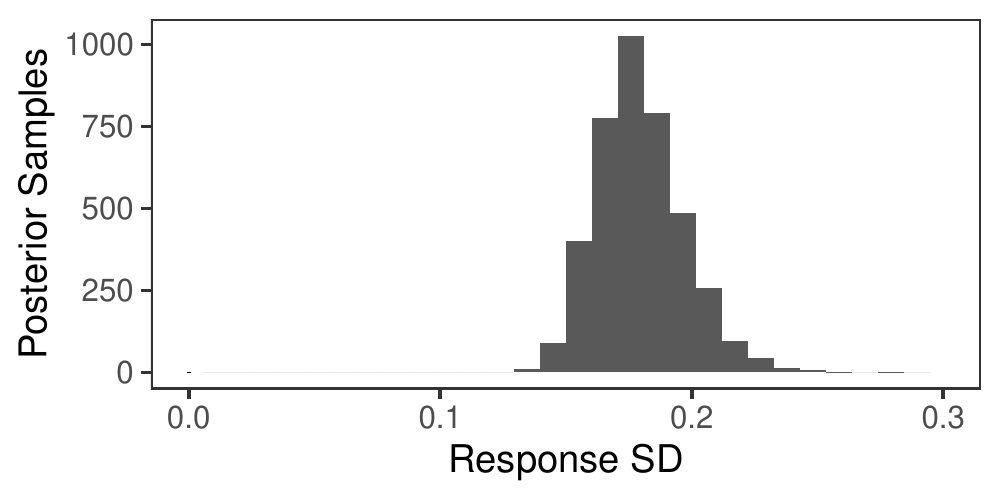} }
\\ \hline
			$\sigma, \tau \sim t(3,0,10)$ &
		\includegraphics[width=0.2\textwidth]{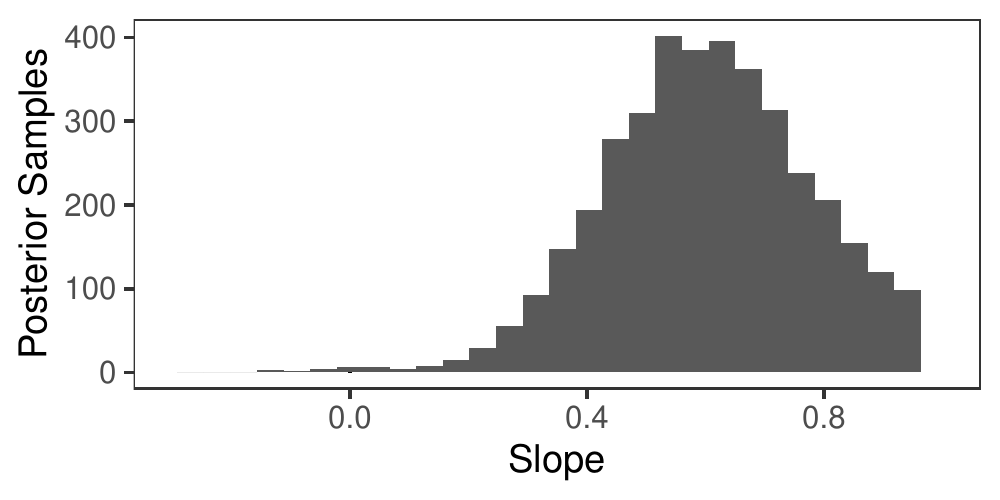} &
\includegraphics[width=0.2\textwidth]{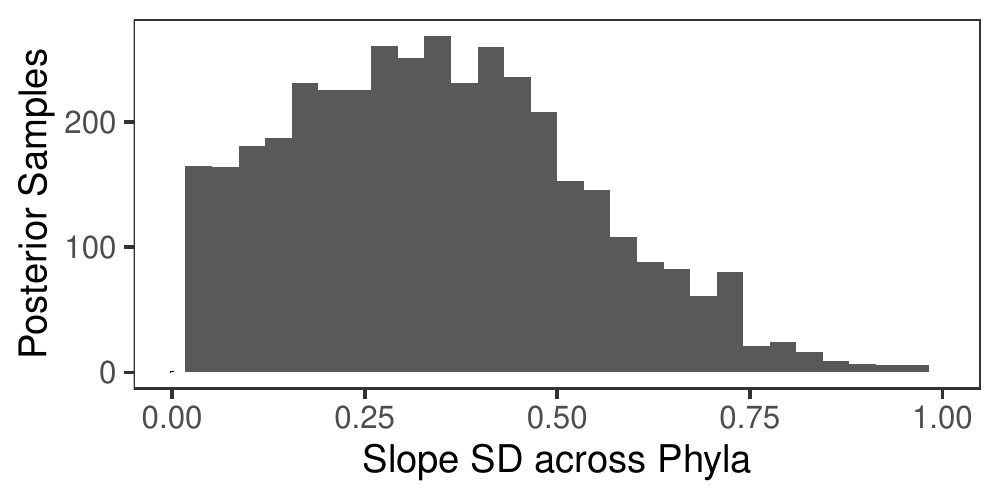} &
\includegraphics[width=0.2\textwidth]{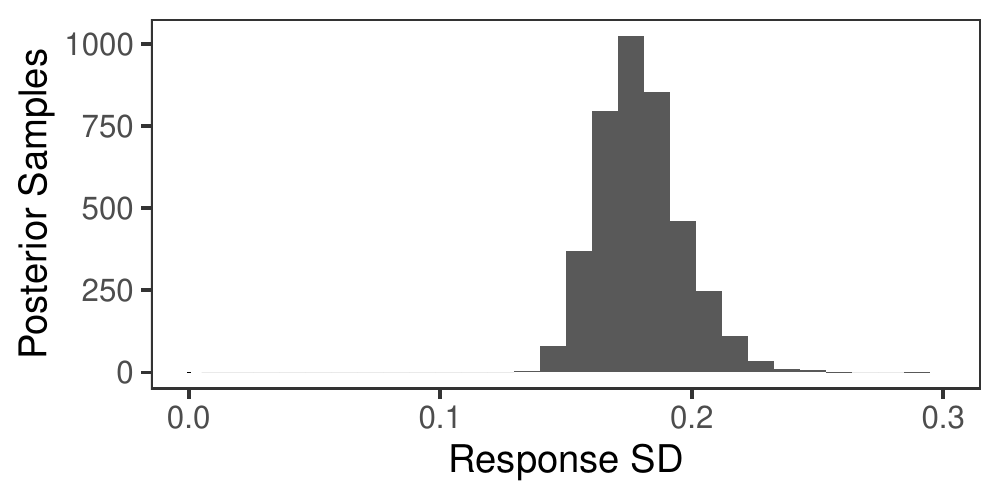} 
\\ \hline
		$\sigma, \tau \sim \mathcal{N}(0,1)$ &
		\includegraphics[width=0.2\textwidth]{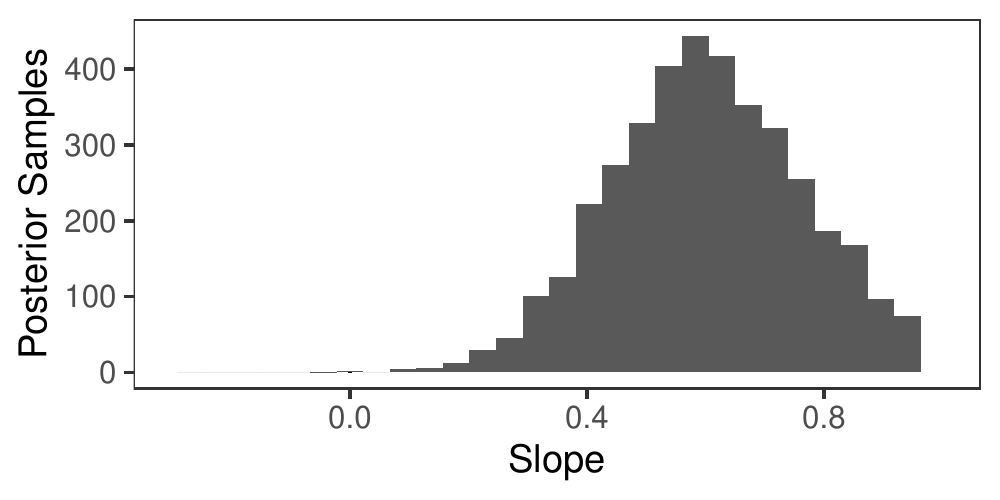} &
\includegraphics[width=0.2\textwidth]{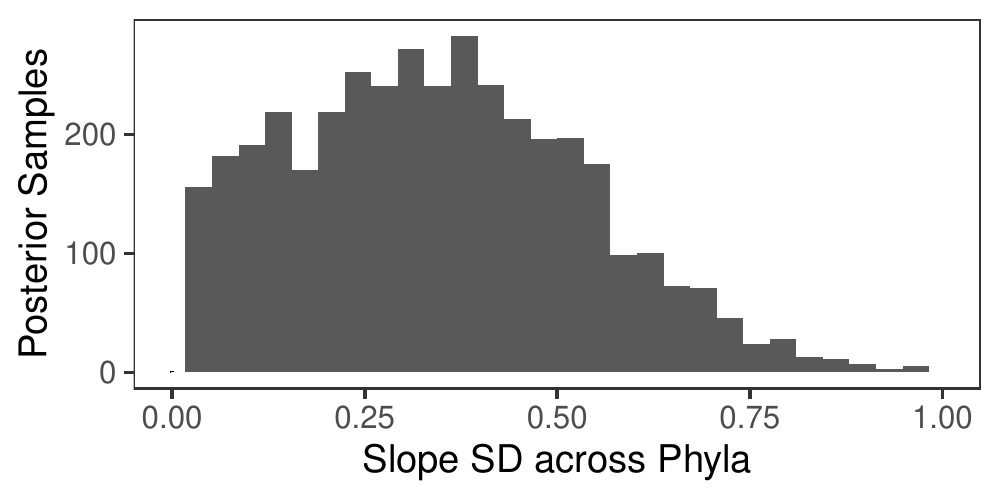} &
\includegraphics[width=0.2\textwidth]{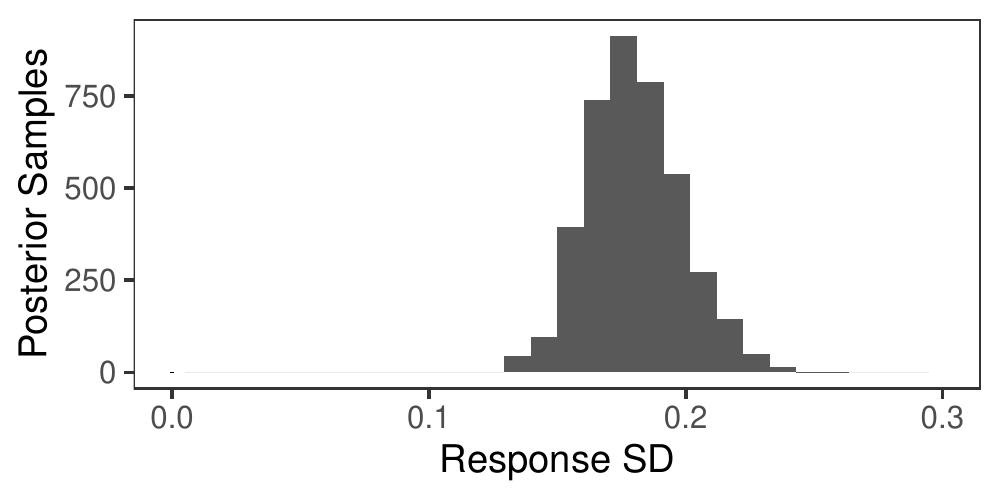} 
\\ \hline
$\sigma, \tau \sim t(3,0,0.5)$ &
		\includegraphics[width=0.2\textwidth]{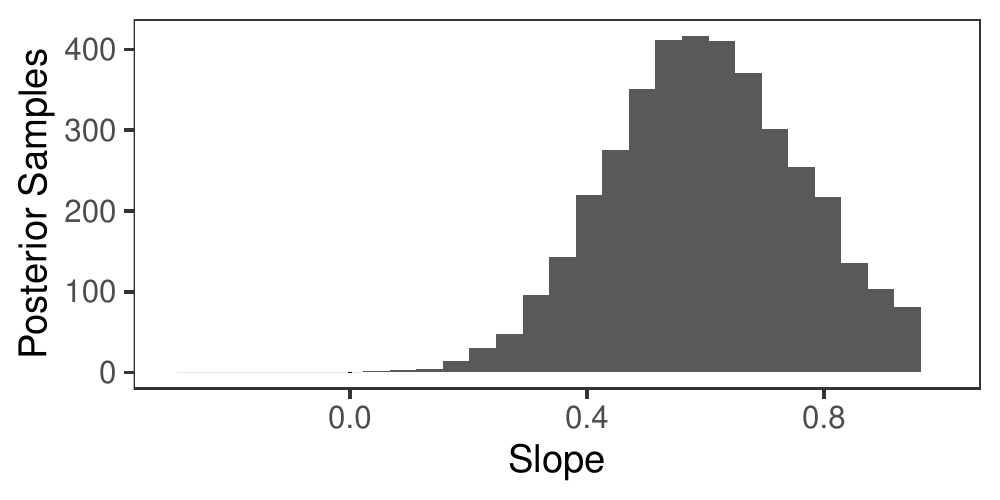} &
\includegraphics[width=0.2\textwidth]{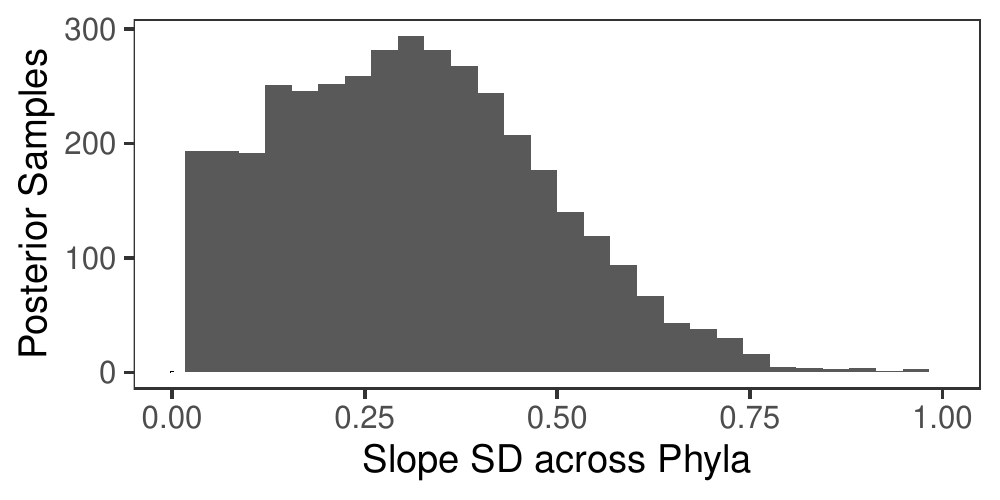} &
\includegraphics[width=0.2\textwidth]{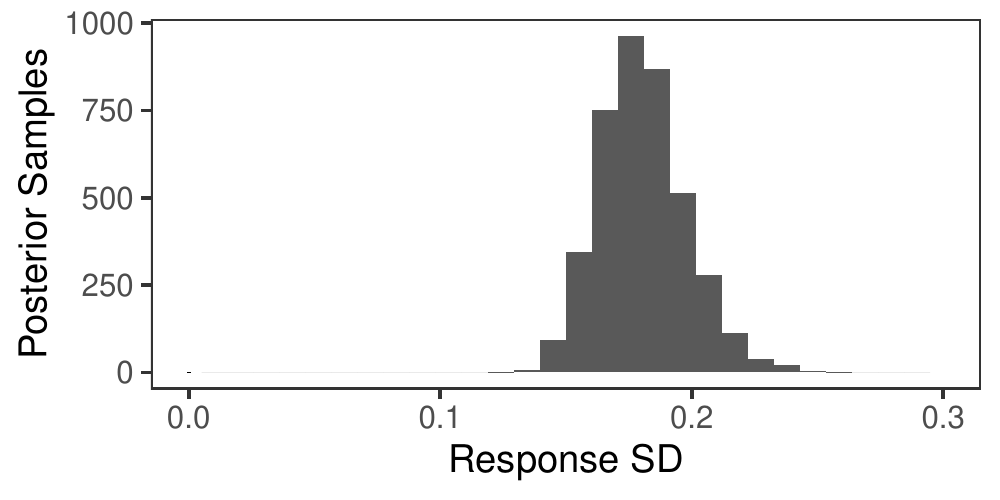} 
\\ \hline
		$\sigma, \tau \sim t(3,0,0.1)$ &
		\includegraphics[width=0.2\textwidth]{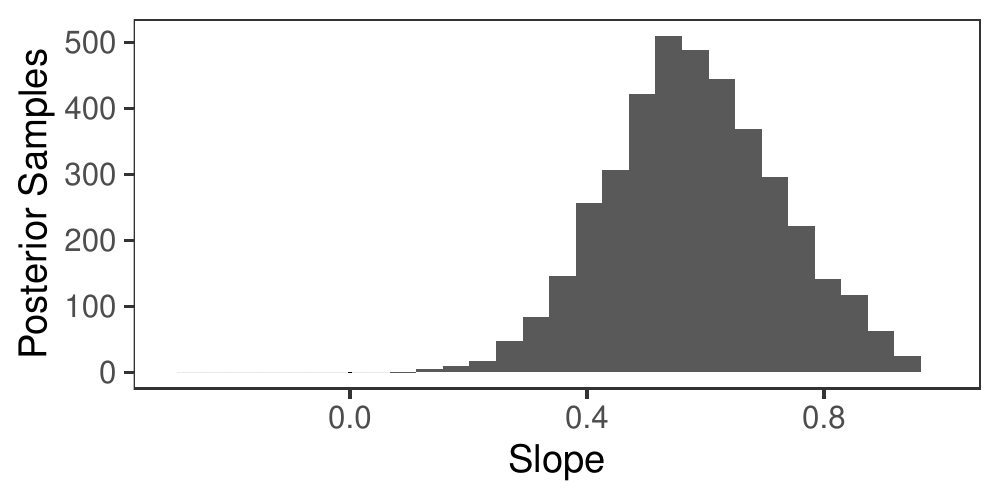} &
\includegraphics[width=0.2\textwidth]{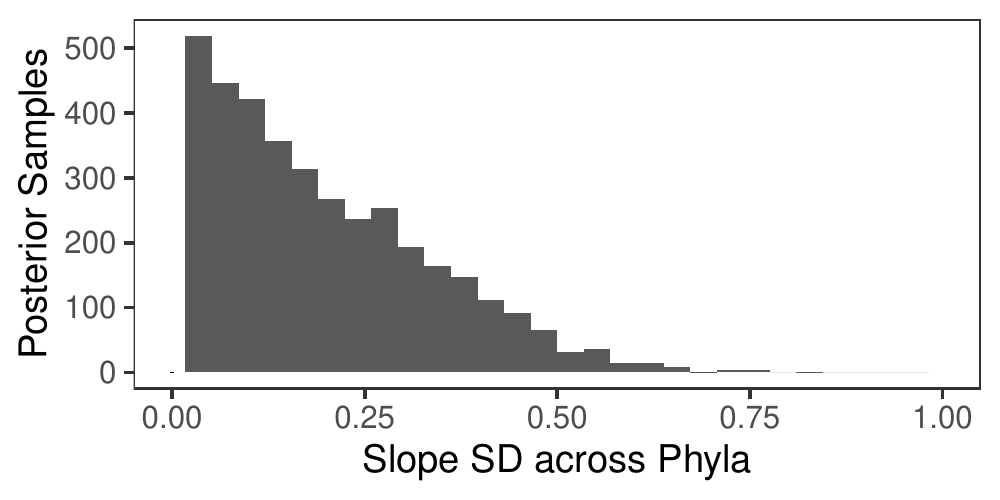} &
\includegraphics[width=0.2\textwidth]{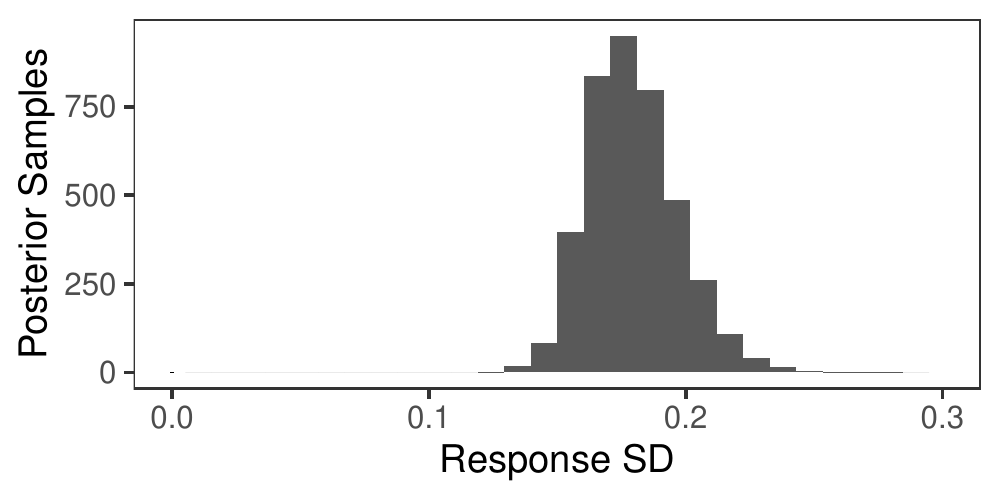} 
\\ \hline
	\end{tabular}
	\caption{Impact of the prior for the variance terms on the posterior in the Bayesian mixed-effects analysis predicting attested subject-object position congruence from average congruence along the Pareto frontier.
	The first line corresponds to the prior used in our analysis; the other priors differ in the degree to which they regularize towards $0$, from mild regularization (top) to very strong regularization (bottom).
	We write $t(\nu, \mu, \sigma)$ for the Student's $t$ distribution with $\nu$ degrees of freedom, location $\mu$, and scale $\sigma$.
	For each prior, we show the posterior of the coefficient $\beta$ (the quantity of interest), the standard deviation of the slope across families, and the standard deviation of the Gaussian response. While changing the prior affects the estimated posterior of the slope variance across families, it has little effect on the estimate of $\beta$. This shows that the estimate of $\beta$ is not impacted by a possible inflation of the variance components linked to the large number of isolated languages.}\label{fig:variance-priors}
\end{figure}

In Figure~\ref{fig:variance-priors}, we plot how the posteriors for $\beta$, the standard deviation $\tau$ of the per-family adjustments $\beta_f$, and the response standard deviation $\sigma$ vary as a function of the prior.
The priors for the fixed effect coefficient $\beta$ ($N(0,1)$) and the intercept ($N(0.5,1)$) are as in the main analysis.
Results show that, while more strongly regularizing priors shrink the estimated range of $\tau$, they have limited impact on the posterior of the key quantity, $\beta$.
Even an unrealistic extremely regularizing prior $t(3,0,0.1)$ does not change the posterior of $\beta$ much.

\section{Further Visualizations for Coadaptation}

\begin{figure*}
    \centering
	    \begin{tabular}{cccc}
    A. Entire Pareto Frontier & B. Only IL & C. Only DL \\
		    \includegraphics[width=0.3\textwidth]{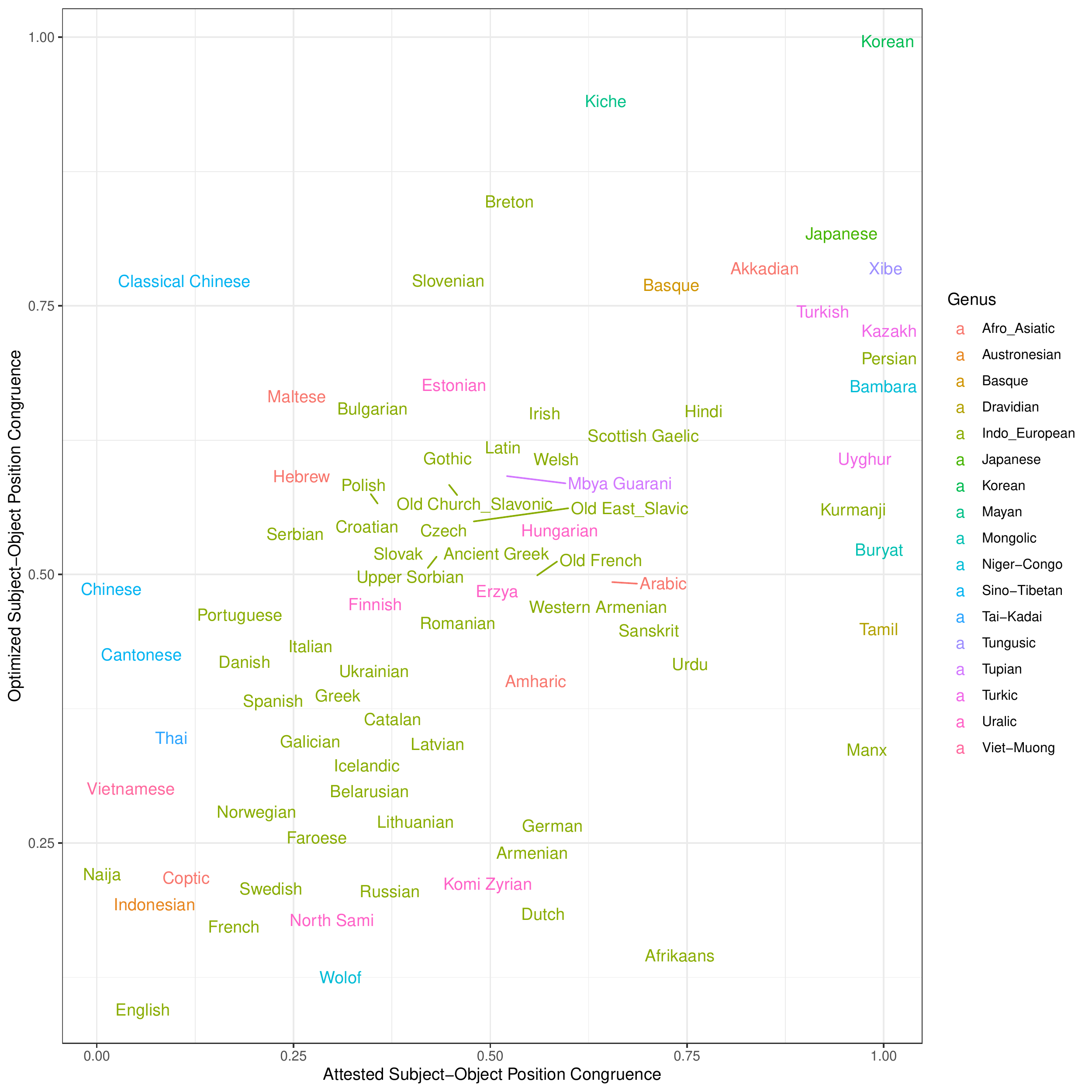} &
		    \includegraphics[width=0.3\textwidth]{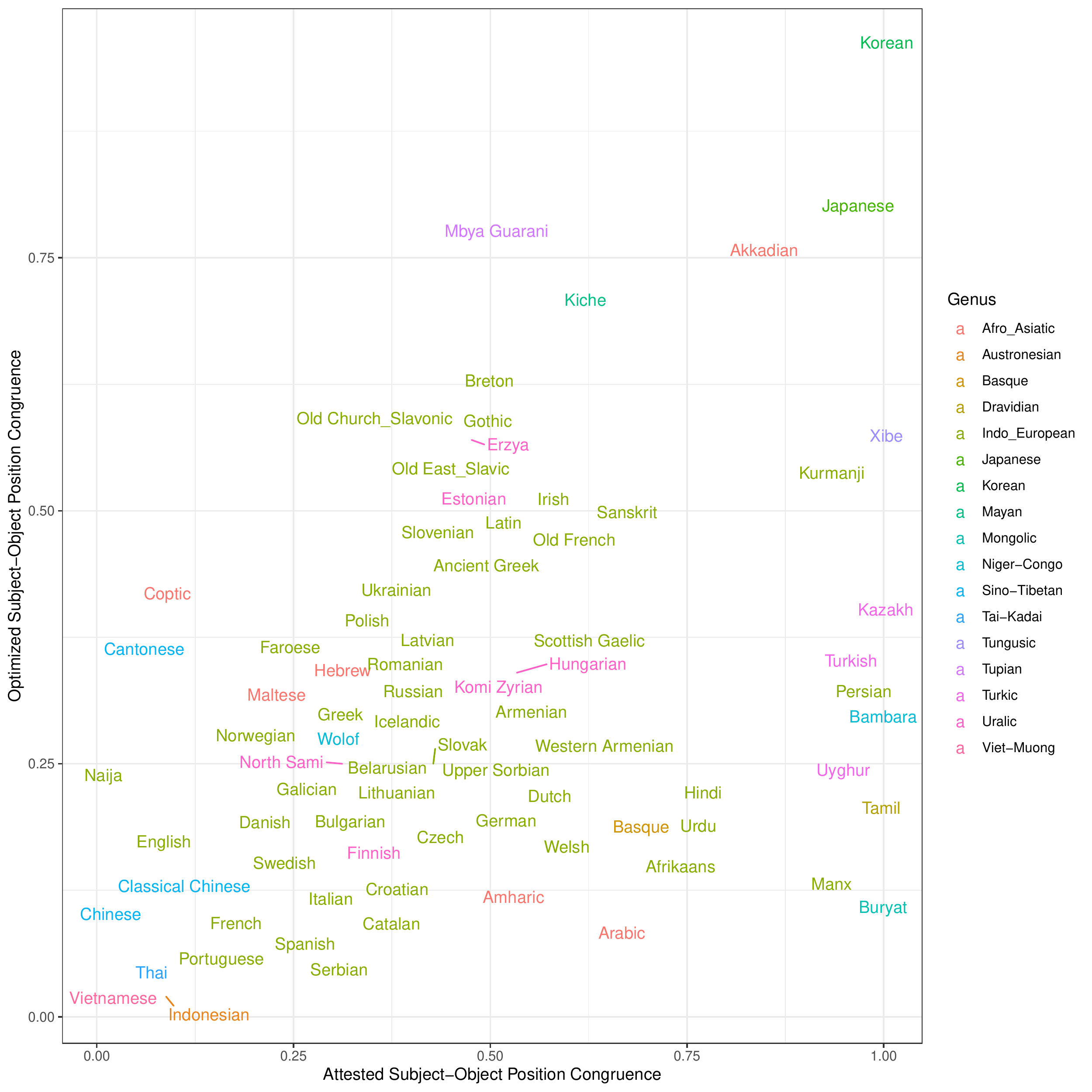} &
	    \includegraphics[width=0.3\textwidth]{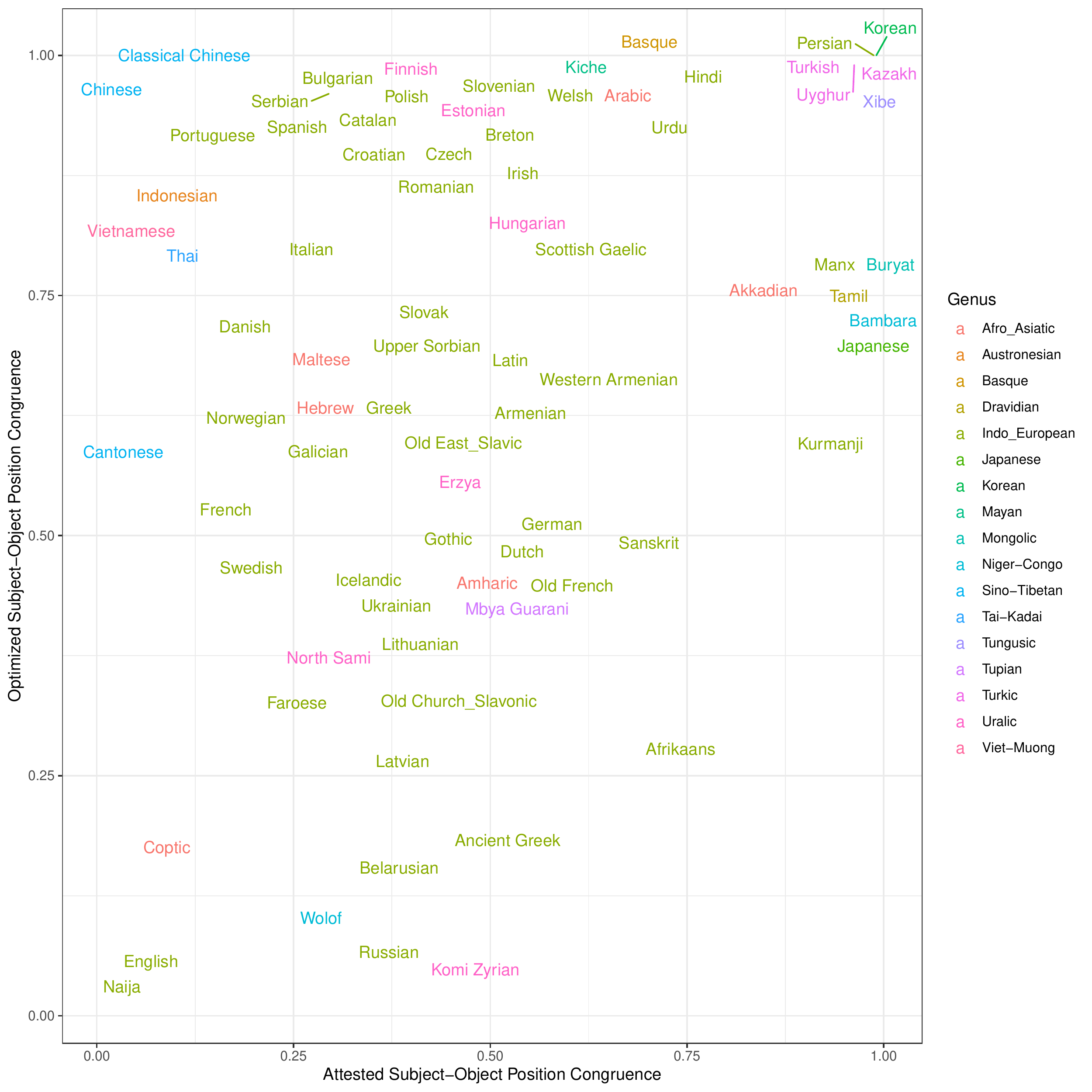}
\end{tabular}

    \caption{
	    Attested and optimized subject-object position congruence (compare Figure~3 in the main paper), with language names, colored by the 17 families represented in the dataset.
	    Compare Figure~\ref{fig:coadaptation-names-joint} for results from joint analysis in higher resolution.
	}
    \label{fig:coadaptation-names}
\end{figure*}

\begin{figure*}
    \centering
	    \includegraphics[width=0.7\textwidth]{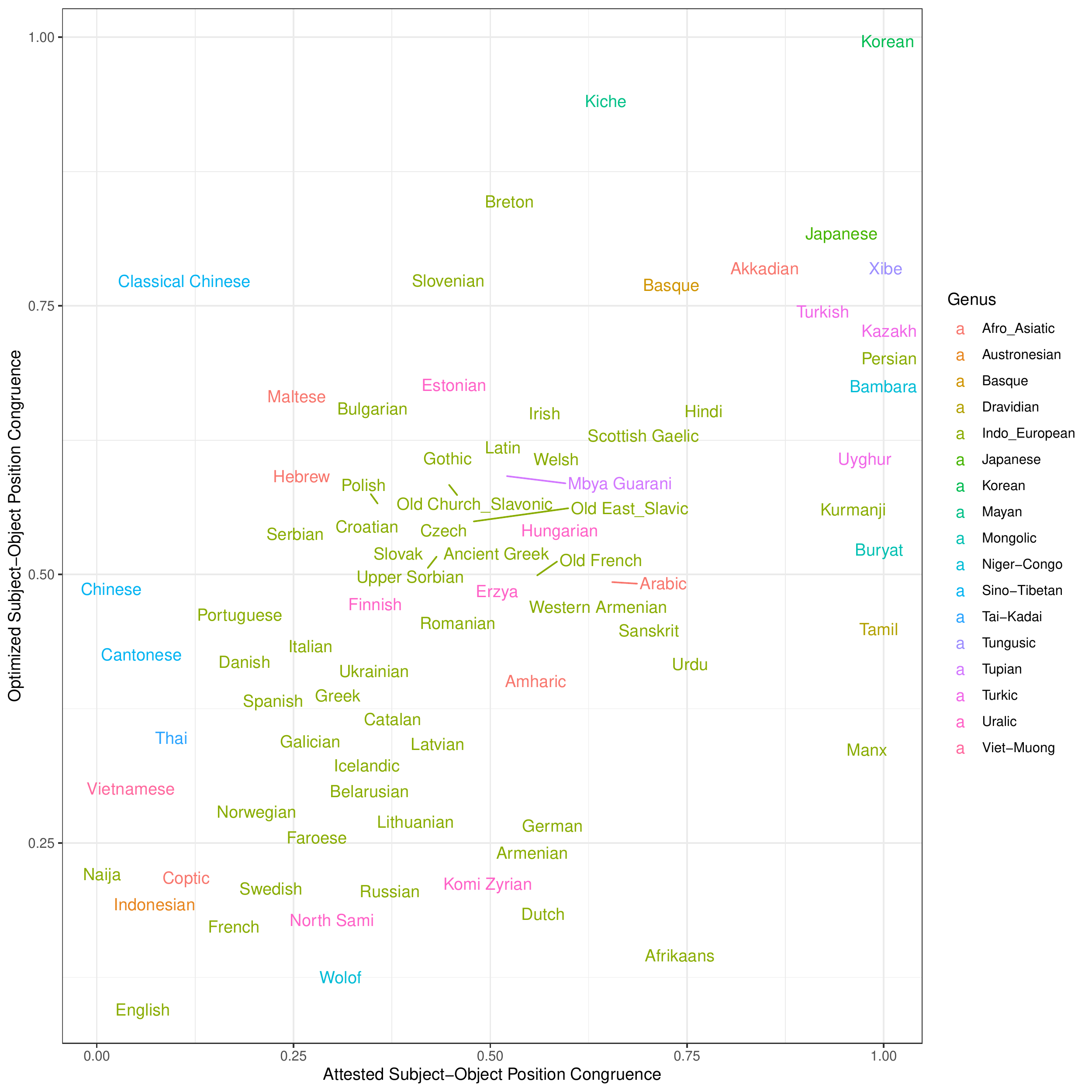}

    \caption{
	    Attested and optimized subject-object position congruence (compare Figure~3 in the main paper), with language names, colored by the 17 families represented in the dataset.
	    Compare Figure~\ref{fig:coadaptation-names} for results optimizing only for DL or IL.
	}
    \label{fig:coadaptation-names-joint}
\end{figure*}

\begin{figure*}
    \centering

	    \includegraphics[width=0.7\textwidth]{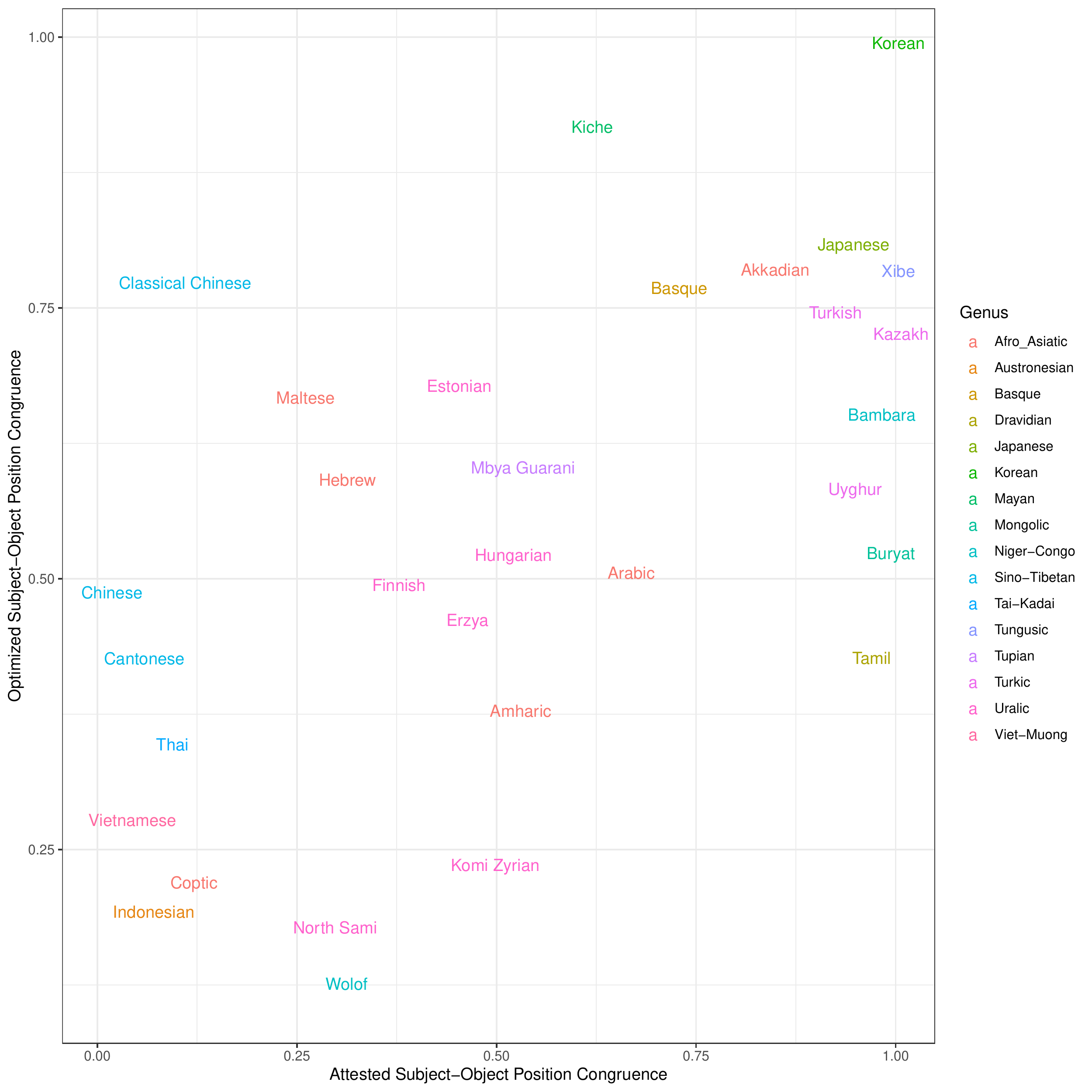}

    \caption{
	    Attested and optimized subject-object position congruence (compare Figure~3 in the main paper), excluding the Indo-European languages.
	    Compare Figure~\ref{fig:coadaptation-names-joint} for results on all 80 langiages.
	}
    \label{fig:coadaptation-names-noIE}
\end{figure*}

See Figures~\ref{fig:coadaptation-names}--\ref{fig:coadaptation-names-noIE} for versions of Figure 4 in the main paper with language names.

We further investigated the robustness of the correlation between attested and average optimized subject-object position congruence to possible outliers.
Correlations, in particular Pearson correlations, are vulnerable to outliers and points of high leverage.
In order the evaluate whether this impacted the results, we considered all subsets of $\leq 3$ languages, and recomputed the correlation when excluding this subset.
The correlation was in the range $[0.42, 0.61]$ for all such subsets.
This suggests that the correlation is not inflated due to individual points of high leverage.

\newpage\section{Per-Family Results and Fitted Slopes}

\begin{figure*}
    \centering
	\begin{tabular}{ccccc}
		A. Fit by Family & B. Residual by Family \\
		\includegraphics[width=0.4\textwidth]{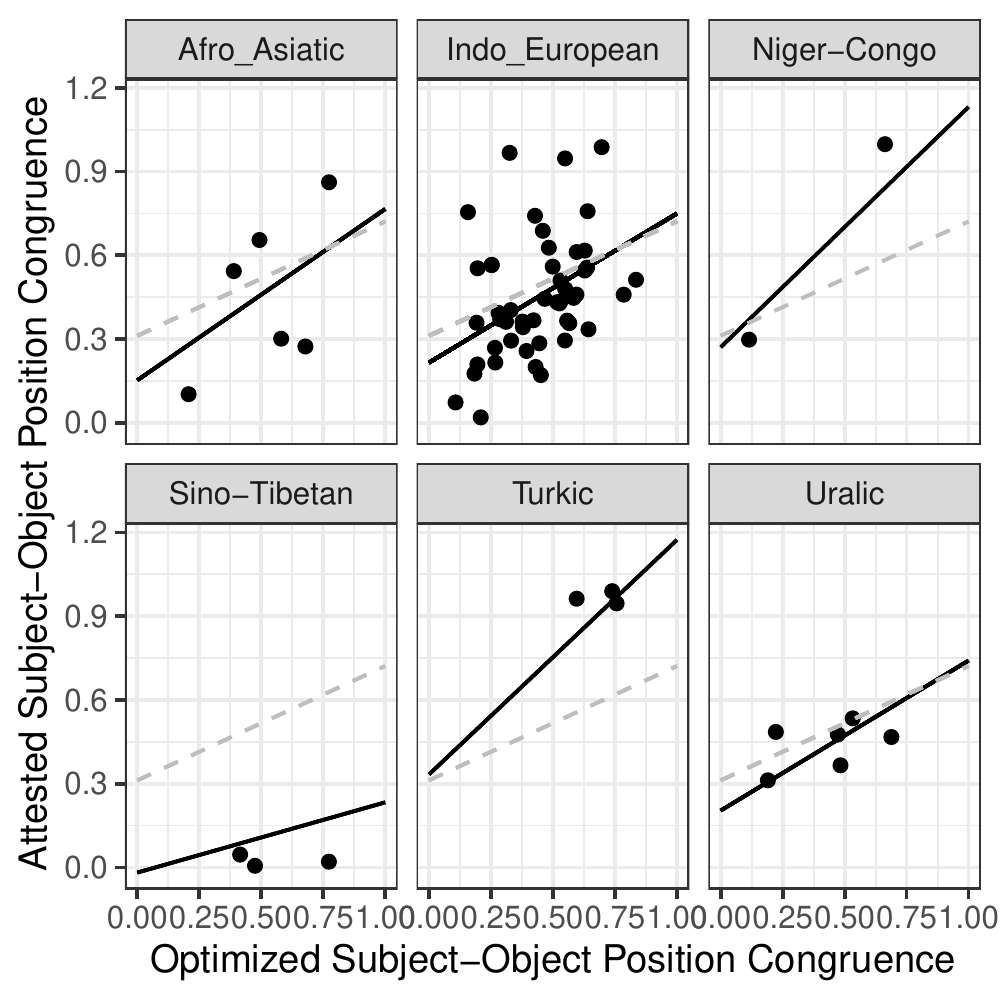} &
            \includegraphics[width=0.4\textwidth]{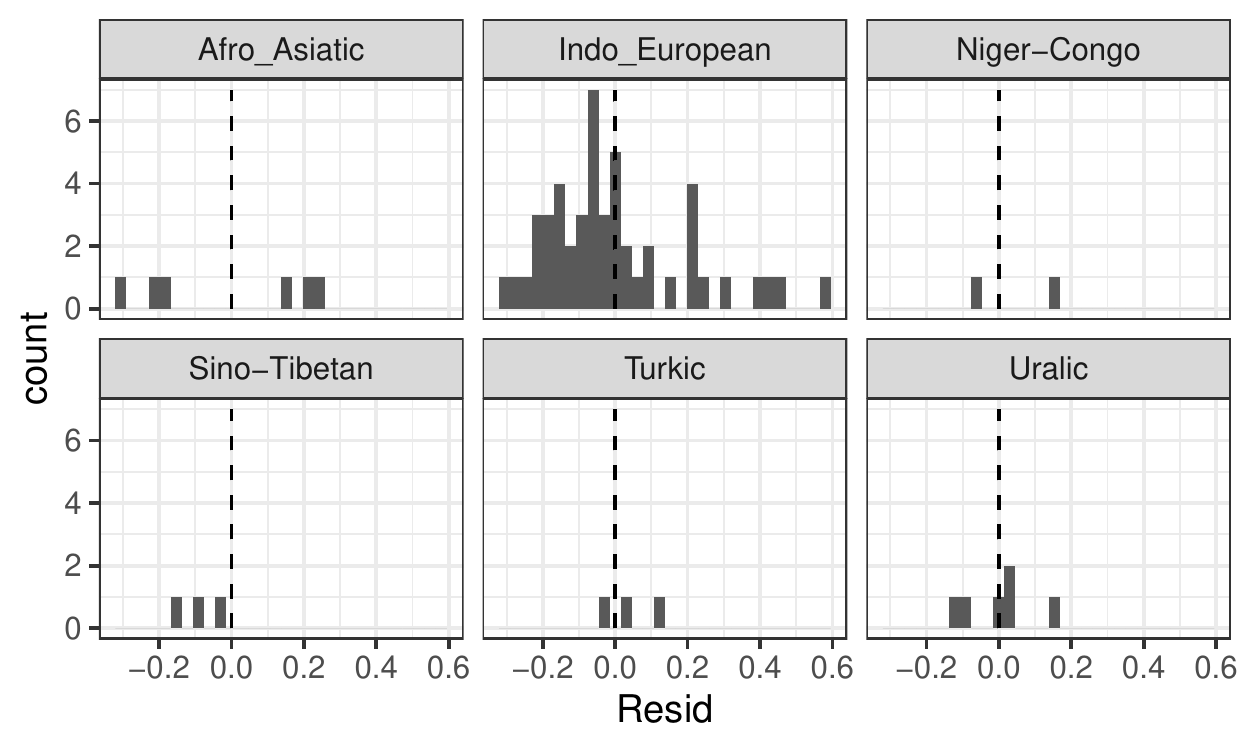} \\
C. Means per Family \\
	                   \includegraphics[width=0.4\textwidth]{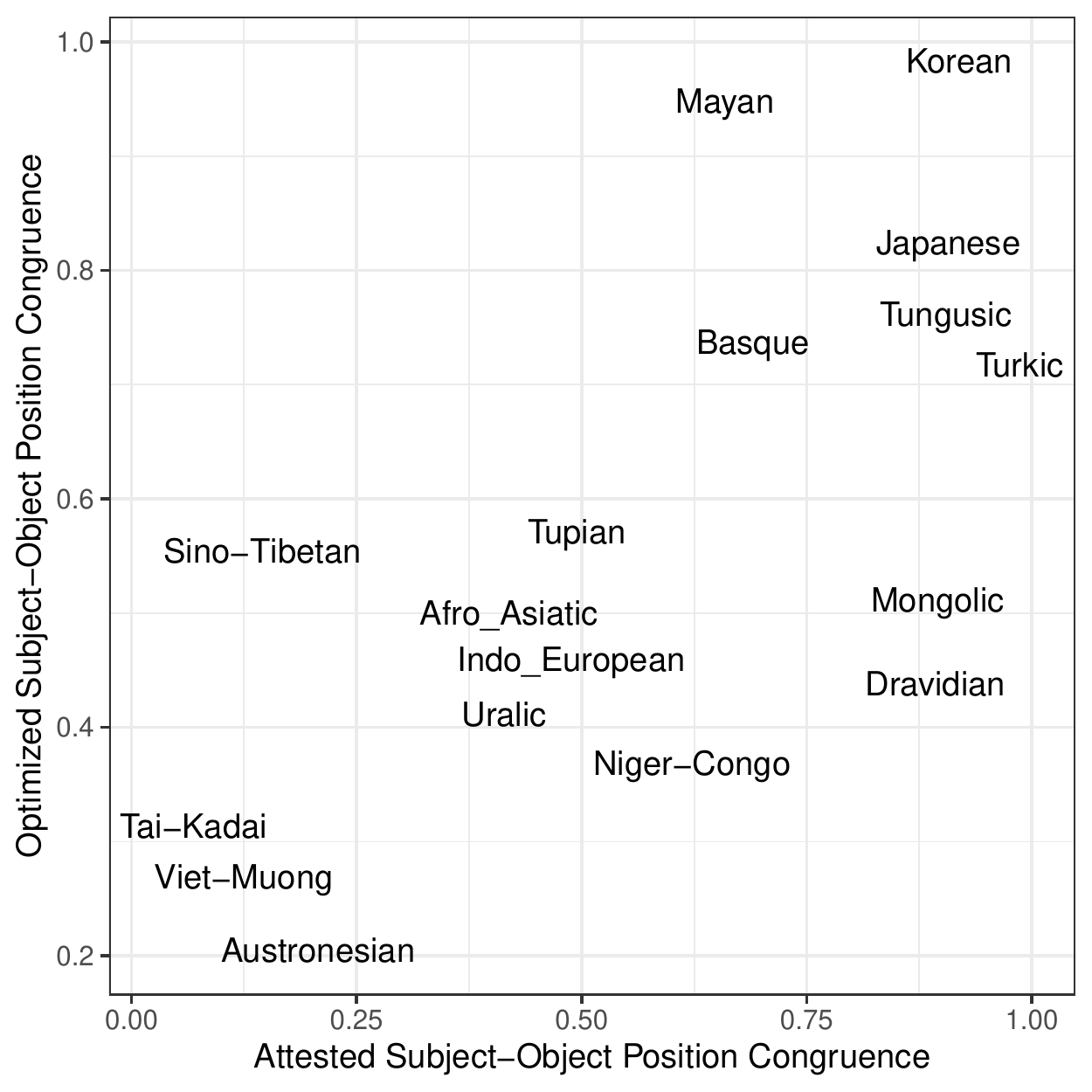}
	\end{tabular}

    \caption{
	    A: Fit of the mixed-effects model across the 6 families represented by at least two languages in the Universal Dependencies dataset. We show the overall slope fitted by the mixed-effects analysis across the 80 languages as a dashed line, and the per-family adjusted slope as a solid line.
	In both cases, we use the posterior mean of intercepts and slopes. Note that, for less well represented families, the model has substantial uncertainty about the slope, not well represented by the point estimates, and even in families with seemingly divergent slope, the data are statistically consistent with the slope being in fact the same across families, see Figure~\ref{fig:per-phylum-posteriors-1}.
	B: Residuals by family tend to be centered around zero.
	C: Means across all languages within each family. This illustrates that the per-family means also exhibit a positive correlation: That is, a positive correlation is supported both across families, and within the larger families individually.
	}
    \label{fig:per-phylum-plot}
\end{figure*}

\begin{figure*}
    \centering
            \includegraphics[width=0.8\textwidth]{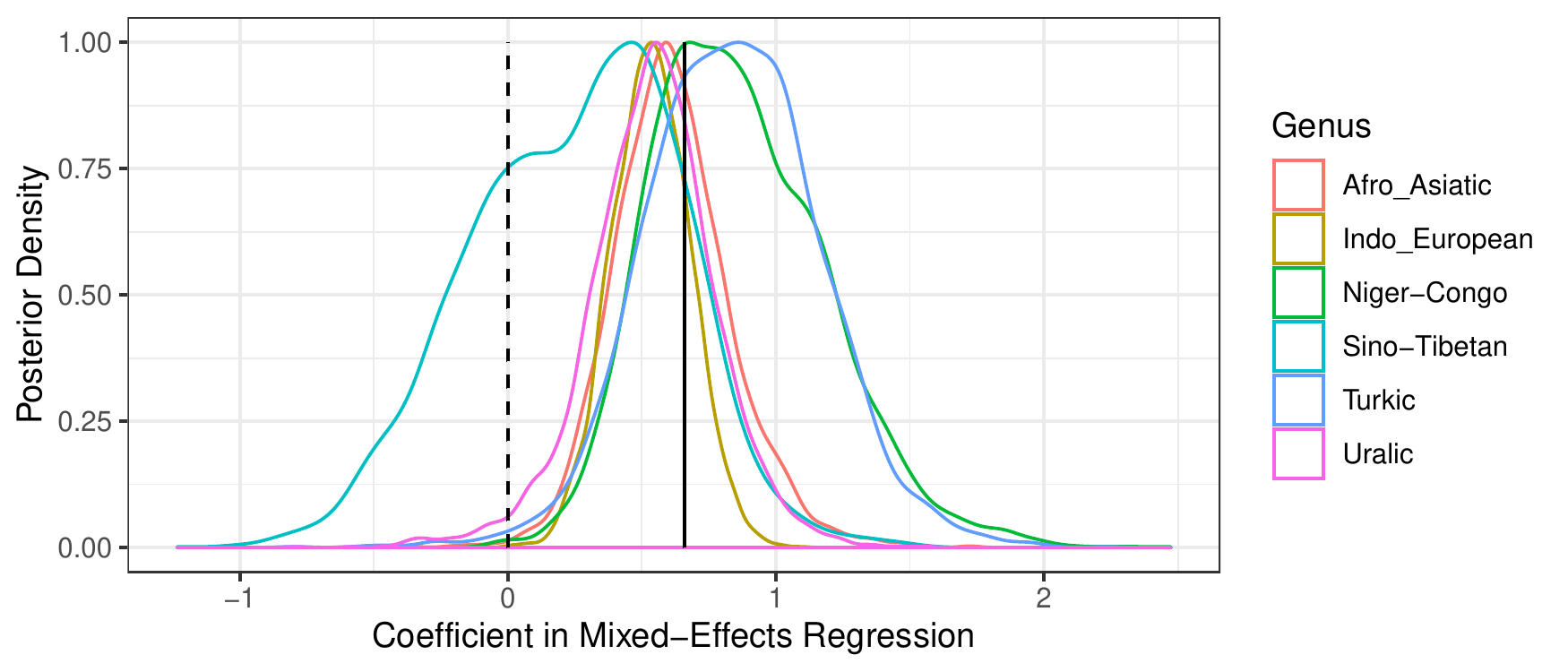}
	\caption{Posterior Densities (scaled so all are bounded by 1) for the slope in the linear mixed-effects regression in the six families with at least two languages. For poorly represented families, the posterior is wider. Nonetheless, across families, the model assigns almost all of the posterior probability mass to a positive sign, except in Sino-Tibetan, where the dependent variable has almost no variance. While the posterior mode differs between the families, the posterior is always well compatible with the overall estimated $\beta$, indicated by a solid vertical line.}
    \label{fig:per-phylum-posteriors-1}
\end{figure*}

Figure~\ref{fig:per-phylum-plot} shows results across the 17 families, including the six ones represented by at least two languages, for the analysis of optimized and attested subject-object position congruence.
In Figure~\ref{fig:per-phylum-posteriors-1}, we show the fitted slope $\beta+\beta_f$ (fixed effects slope $\beta$ plus per-family adjustment $\beta_f$) for each family that has at least two languages.

We note that, while smaller families do not provide sufficient evidence for a positive relationship on their own, estimating the overall slope in a mixed-effects analysis does not require independent estimates of the slopes in each family.
Instead, the mixed-effects regression obtains its slope estimate by combining (i) the data across isolates and smaller families, and (ii) the slope within the well-represented Indo-European family.
Thus, for the purposes of the mixed-effects analyses, the presence of many families, even isolates and sparsely represented ones, can provide an advantage, because it increases the amount of statistical independence in the dataset.

\newpage\section{Detailed Results for Figure 3 in Main Paper}

\begin{figure}
	\centering
		\includegraphics[draft=false,width=0.9\textwidth]{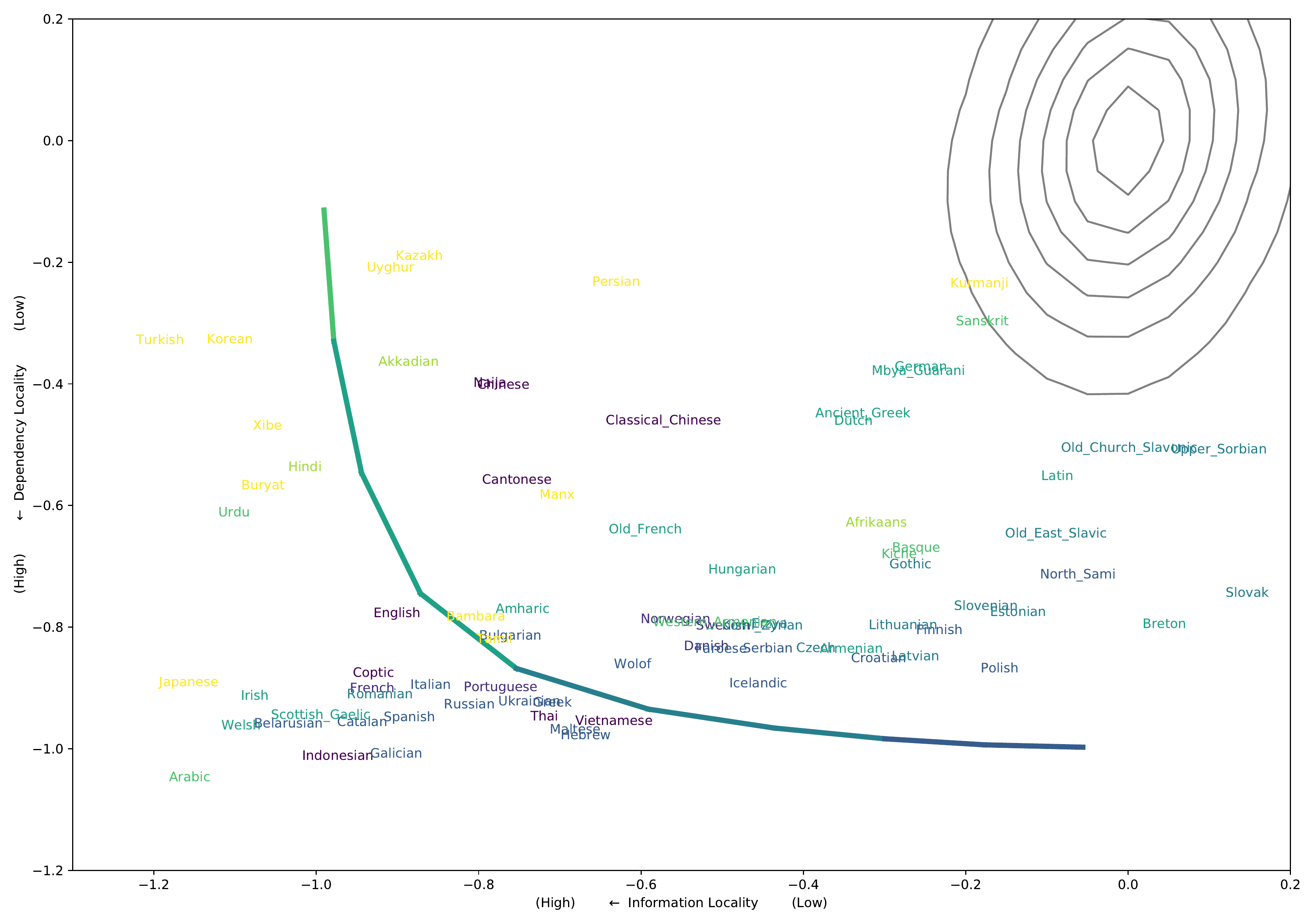}

   Colors denoting Subject-Object Position Congruence:

   0 \ \ \ \ \ \ \ \ \ \ \ \ \ \ \ \ \ \ \ \ 0.5 \ \ \ \ \ \ \ \ \ \ \ \ \ \ \ \ \ \ \ 1

     \includegraphics[width=0.3\textwidth]{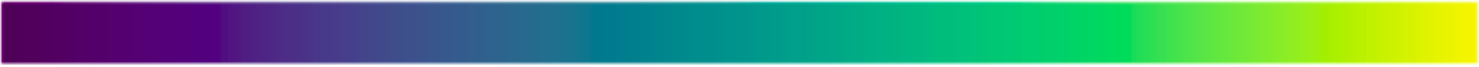}

	\caption{Position of the 80 languages in the efficiency plane with all language names. Compare Figure 3 in the main paper.}\label{fig:plane-with-names}
\end{figure}

\begin{figure}
	\centering
	\begin{tabular}{ccccc}
	\ \ \ 	\ \ \ SVO & \ \ \ \ \ \ SOV & \ \ \ \ \ \ VSO & \ \ \ \ \ \ No dominant order \\
		\includegraphics[draft=false,width=0.2\textwidth]{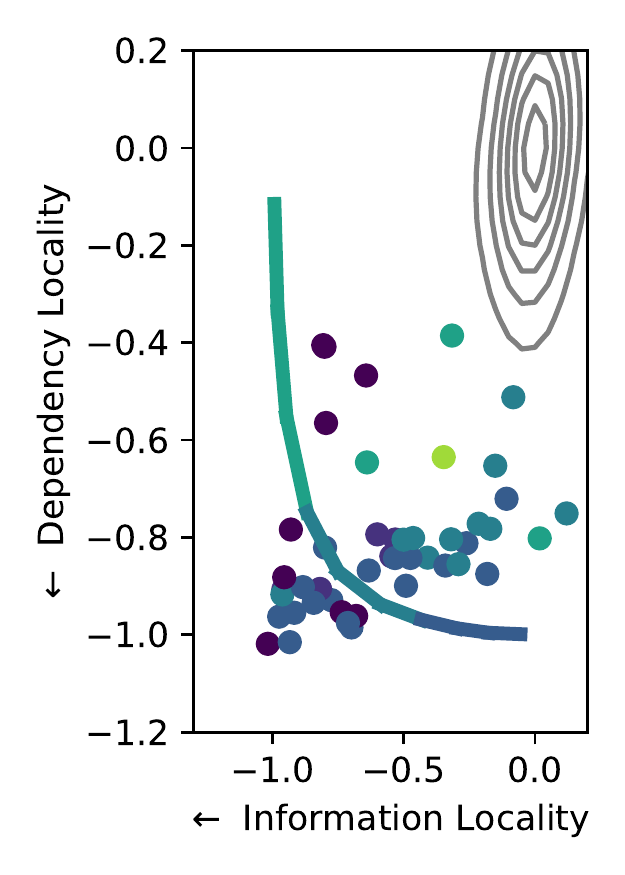} &
		\includegraphics[draft=false,width=0.2\textwidth]{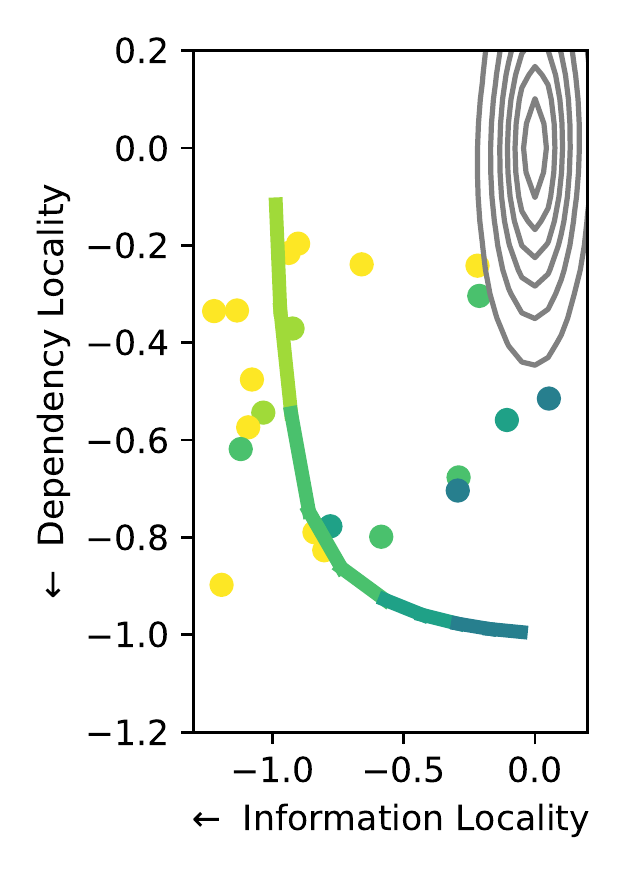} &
		\includegraphics[draft=false,width=0.2\textwidth]{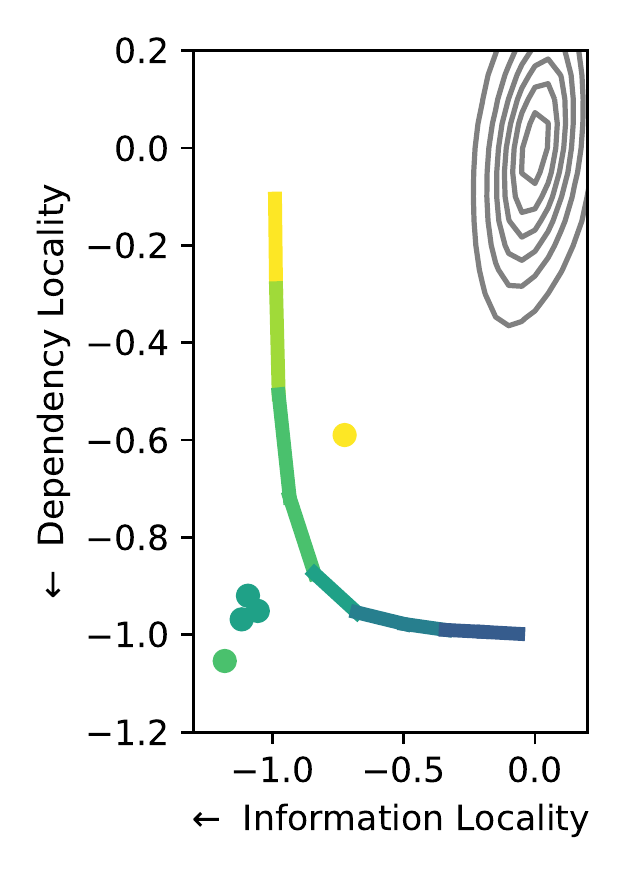} &
\includegraphics[draft=false,width=0.2\textwidth]{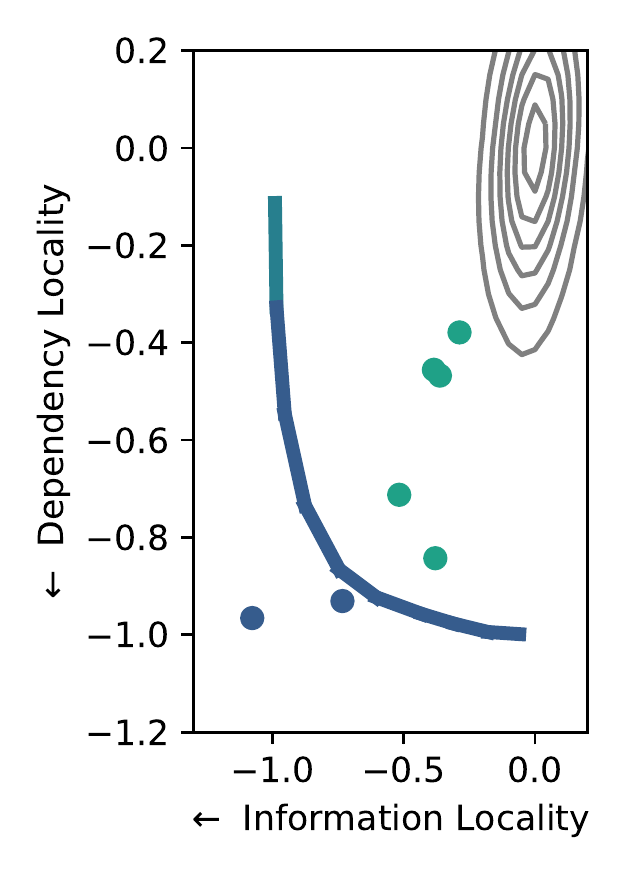} 
	\end{tabular}

   Colors denoting Subject-Object Position Congruence:

   0 \ \ \ \ \ \ \ \ \ \ \ \ \ \ \ \ \ \ \ \ 0.5 \ \ \ \ \ \ \ \ \ \ \ \ \ \ \ \ \ \ \ 1

     \includegraphics[width=0.3\textwidth]{Screenshot_from_2021-11-14_17-04-11.png}

	\caption{Position of languages in the efficiency plane spanned by IL and DL, per word order category. Compare Figure 3 in the main paper.}
	\label{fig:plane-by-categorical}
\end{figure}

See Figure~\ref{fig:plane-by-categorical} for results per word order category, including less frequent categories ``VSO'' and ``No dominant order''.

\newpage\section{Results using Raw Counts}\label{sec:raw-counts}

\begin{figure}
	\centering

    \begin{tabular}{cccc}
\ \ \    A. Average & \ \ \ B. Only IL & \ \ \ C. Only DL \\
	    \includegraphics[draft=false,width=0.25\textwidth]{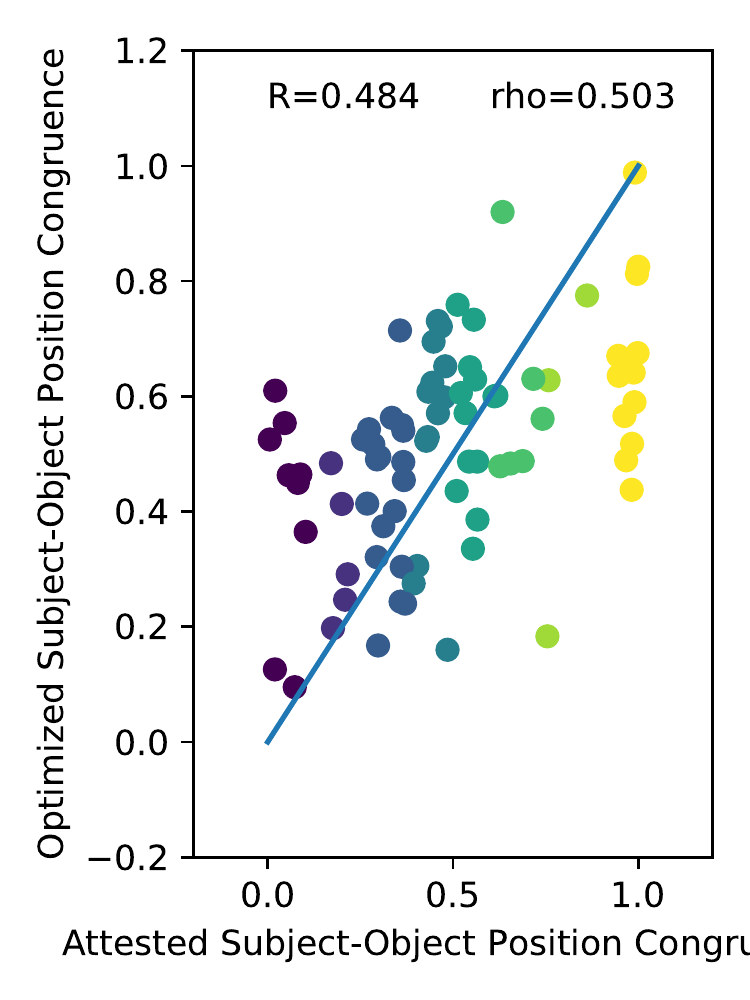}  &
	    \includegraphics[draft=false,width=0.25\textwidth]{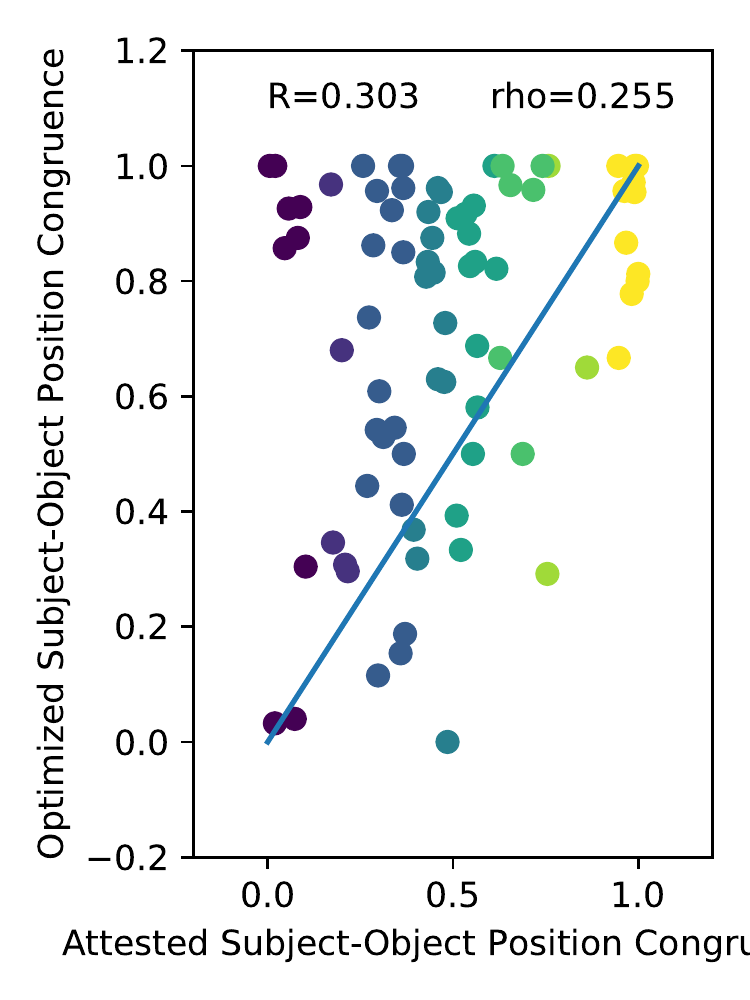}  &
\includegraphics[draft=false,width=0.25\textwidth]{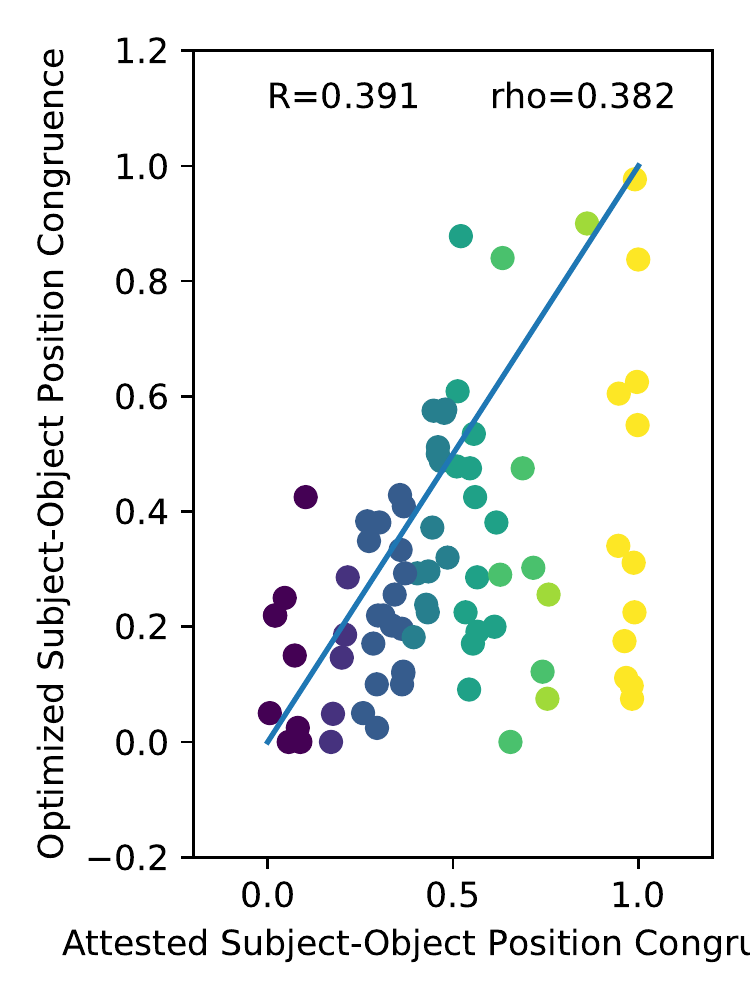} 
    \end{tabular}

   Colors denoting Subject-Object Position Congruence:

   0 \ \ \ \ \ \ \ \ \ \ \ \ \ \ \ \ \ \ \ \ 0.5 \ \ \ \ \ \ \ \ \ \ \ \ \ \ \ \ \ \ \ 1

     \includegraphics[width=0.3\textwidth]{Screenshot_from_2021-11-14_17-04-11.png}

	\caption{Results using raw counts for grammars optimized only for IL (center), only DL (right), and the average of the two counts (left). Results are very similar to those obtained using smoothed counts along the interpolated Pareto frontier, but do not depend on the method used to interpolate along the frontier.}
	\label{fig:raw-coadaptation}
\end{figure}

Here, we show that results concerning co-adaptation do not depend on the choice of a specific method for interpolating the Pareto frontier, or for interpolating position congruence along it.
Figure~\ref{fig:raw-coadaptation} shows results corresponding to Figure~4 in the main paper, but representing the average optimized subject-object position congruence directly in terms of the average over optimized grammars, instead of smoothed values along the interpolated frontier.
Results closely mirror those reported in the main paper.

\newpage\section{Comparison to Greenberg's Correlations}\label{sec:greenberg}

\begin{figure}
	\begin{center}
\begin{tabular}{|c|ll|c|ccc|cc}
	\hline
	&	\multicolumn{2}{c|}{Correlates with...}   &         \multirow{2}{*}{Real}   &  \multirow{2}{*}{Optimized} & \multirow{2}{*}{Optimized} & \\ 
	&	verb & object  &    & \multirow{2}{*}{for IL} & \multirow{2}{*}{for IL+DL} &\\ 
	&	\emph{wrote} & \emph{letters} & & & & \\ \hline \hline 
	\multirow{2}{*}{\raisebox{.5pt}{\textcircled{\raisebox{-.9pt} {1}}}}	&	adposition    &    noun phrase       
	&   \multirow{2}{*}{  \includegraphics[draft=false,width=0.06\textwidth]{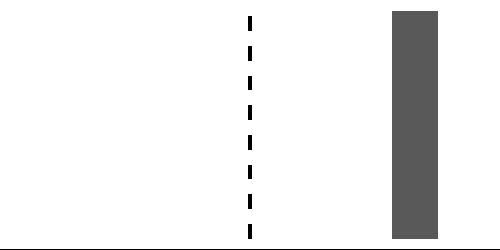}     } 
	&   \multirow{2}{*}{  \includegraphics[draft=false,width=0.06\textwidth]{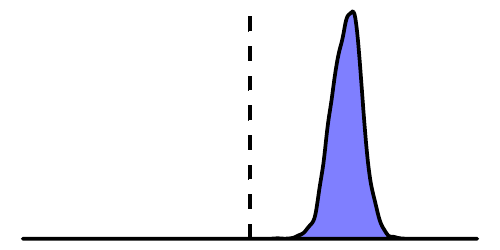}     } 
	&   \multirow{2}{*}{  \includegraphics[draft=false,width=0.06\textwidth]{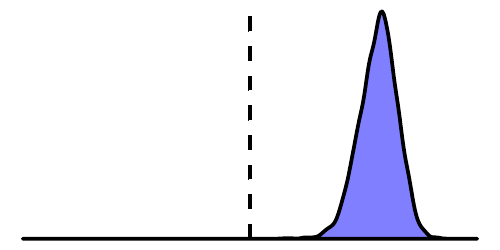}     } &
	\\
	&		\emph{to}            & \emph{a friend} &&&&\\ \hline
	\multirow{2}{*}{\raisebox{.5pt}{\textcircled{\raisebox{-.9pt} {2}}}}	&copula    &    noun phrase         
	&   \multirow{2}{*}{  \includegraphics[draft=false,width=0.06\textwidth]{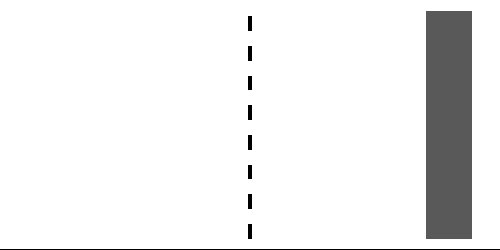}     } 
	&   \multirow{2}{*}{  \includegraphics[draft=false,width=0.06\textwidth]{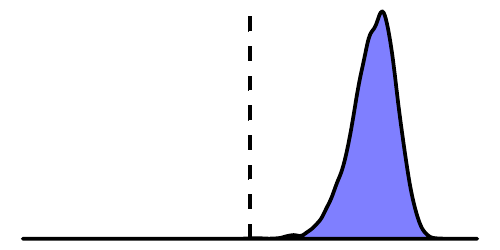}     } 
	&   \multirow{2}{*}{  \includegraphics[draft=false,width=0.06\textwidth]{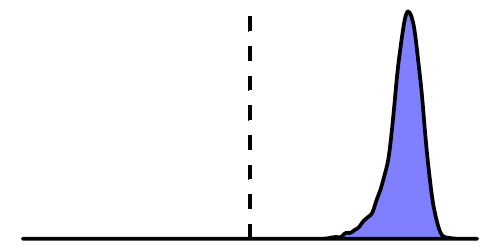}     }& 
	\\
	&	\emph{is}        & \emph{a friend}  &&&&\\ \hline
	\multirow{2}{*}{\raisebox{.5pt}{\textcircled{\raisebox{-.9pt} {3}}}}	&auxiliary    &    verb phrase       
	&   \multirow{2}{*}{  \includegraphics[draft=false,width=0.06\textwidth]{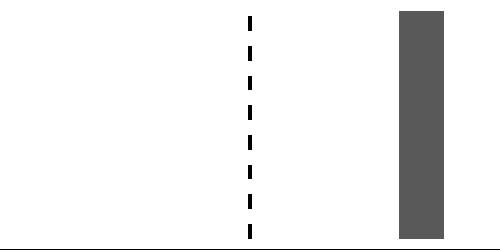}     } 
	&   \multirow{2}{*}{  \includegraphics[draft=false,width=0.06\textwidth]{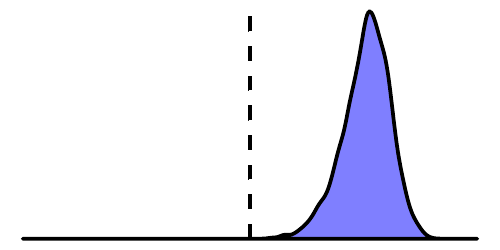}     } 
	&   \multirow{2}{*}{  \includegraphics[draft=false,width=0.06\textwidth]{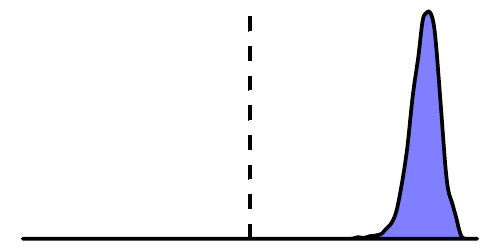}     } &
	\\
	&	\emph{has}          & \emph{written}  &&&&\\ \hline
	\multirow{2}{*}{\raisebox{.5pt}{\textcircled{\raisebox{-.9pt} {4}}}}	&noun    &    genitive      
	&   \multirow{2}{*}{  \includegraphics[draft=false,width=0.06\textwidth]{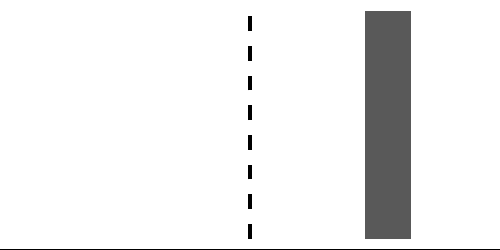}     } 
	&   \multirow{2}{*}{  \includegraphics[draft=false,width=0.06\textwidth]{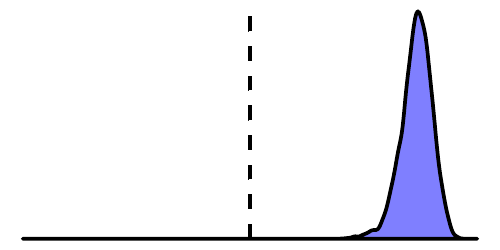}     } 
	&   \multirow{2}{*}{  \includegraphics[draft=false,width=0.06\textwidth]{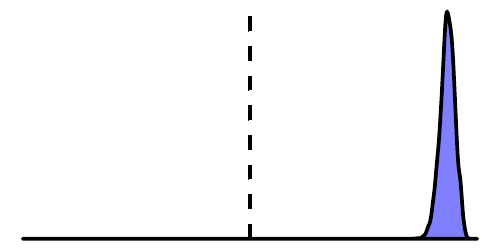}     } &
	\\
	&	\emph{friend} &  \emph{of John}  &&&&\\ \hline
	\multirow{2}{*}{\raisebox{.5pt}{\textcircled{\raisebox{-.9pt} {5}}}}	&noun    &    relative clause      
	&   \multirow{2}{*}{  \includegraphics[draft=false,width=0.06\textwidth]{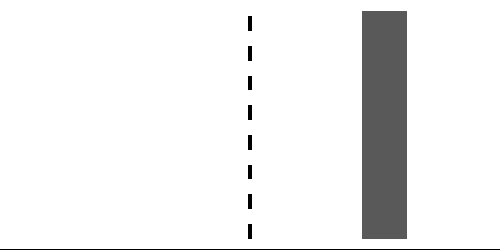}     } 
	&   \multirow{2}{*}{  \includegraphics[draft=false,width=0.06\textwidth]{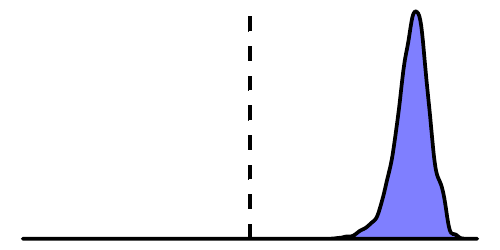}     } 
	&   \multirow{2}{*}{  \includegraphics[draft=false,width=0.06\textwidth]{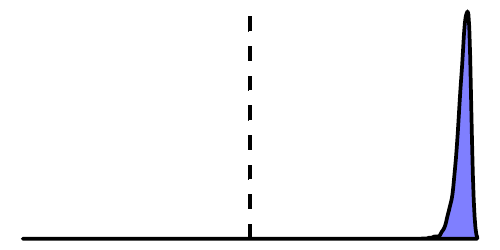}     } &
	\\
	&	\emph{books} & \emph{that you read}  &&&&\\ \hline
	\multirow{2}{*}{\raisebox{.5pt}{\textcircled{\raisebox{-.9pt} {6}}}}	&complementizer    &    sentence        
	&   \multirow{2}{*}{  \includegraphics[draft=false,width=0.06\textwidth]{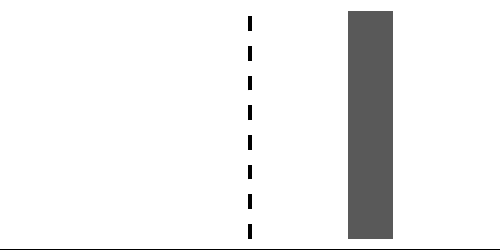}     } 
	&   \multirow{2}{*}{  \includegraphics[draft=false,width=0.06\textwidth]{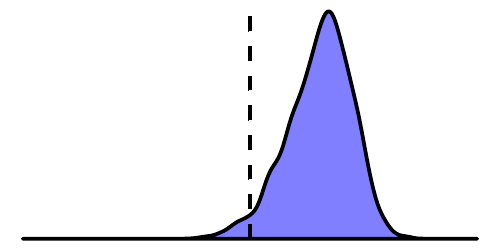}     } 
	&   \multirow{2}{*}{  \includegraphics[draft=false,width=0.06\textwidth]{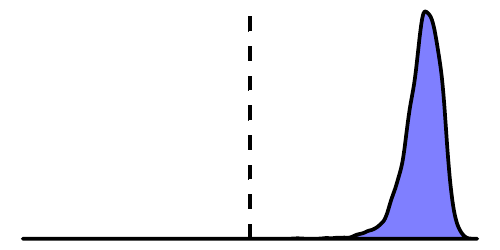}     } &
	\\
	&	\emph{that} & \emph{she has arrived}  &&&&\\ \hline
	\multirow{2}{*}{	\raisebox{.5pt}{\textcircled{\raisebox{-.9pt} {7}}}}	&verb    &    adp. phrase         
	&   \multirow{2}{*}{  \includegraphics[draft=false,width=0.06\textwidth]{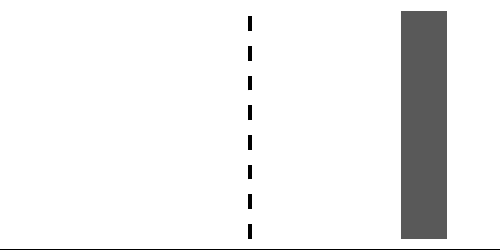}     } 
	&   \multirow{2}{*}{  \includegraphics[draft=false,width=0.06\textwidth]{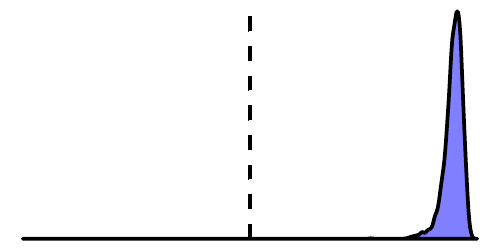}     } 
	&   \multirow{2}{*}{  \includegraphics[draft=false,width=0.06\textwidth]{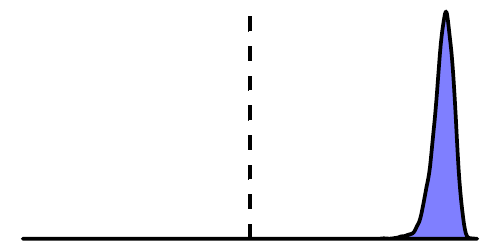}     } &
	\\
	&	\emph{went} & \emph{to school}  &&&&\\ \hline
	\multirow{2}{*}{\raisebox{.5pt}{\textcircled{\raisebox{-.9pt} {8}}}}	&want    &    verb phrase        
	&   \multirow{2}{*}{  \includegraphics[draft=false,width=0.06\textwidth]{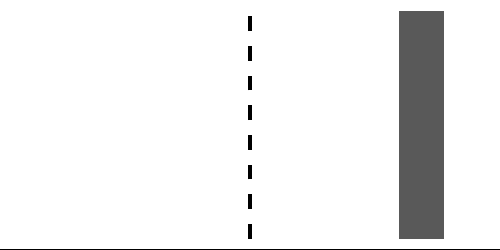}     } 
	&   \multirow{2}{*}{  \includegraphics[draft=false,width=0.06\textwidth]{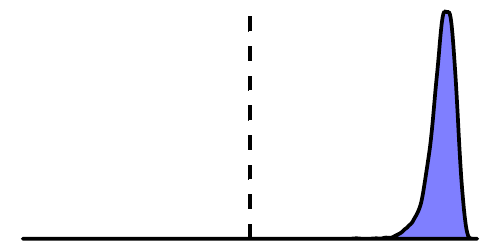}     } 
		&   \multirow{2}{*}{  \includegraphics[draft=false,width=0.06\textwidth]{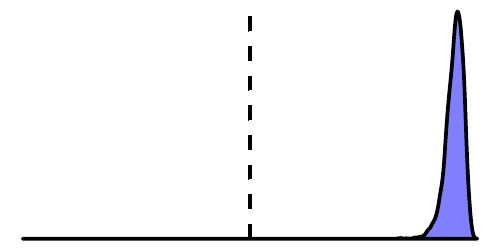}     } &
	\\
	& \emph{wants}   &  \emph{to leave}  &&&&\\ \hline
\end{tabular}
	\end{center}

\caption{Optimizing for Information Locality predicts the Greenberg correlations.
Following \citet{dryer1992greenbergian}, each correlation defines a pair of syntactic elements whose ordering is correlated with the relative order of object and verb; for instance, languages that place the object after the verb (``wrote letters'') tend to place adpositions before the noun phrase (``to a friend''); languages that place the object after the verb (letters -- wrote, Japanese) tend to place adpositions after the noun phrase (friend -- to).
For each correlation, we provide its prevalence (between 0\% and 100\%) among actual grammars of languages represented in Universal Dependencies (left, from \citet{hahn2020universals}), and the posterior distribution of the prevalence among grammars optimized for IL and DL, obtained from a mixed-effects analysis with by-language and by-family random effects (as in the analysis of \citet{hahn2020universals}, but using the 80 languages from our sample used here).
Optimization predicts all eight correlations to hold in the majority of grammars, matching the distribution observed in real languages. 
}\label{table:corr-resu}
\end{figure}

Here, we show that Greenberg's correlation universals \citep{greenberg-universals-1963, dryer1992greenbergian} arise from both IL and DL individually.
Prior work has argued, using theoretical arguments, that these universals arise from optimizing DL \citep{frazier1985syntactic, rijkhoff-word-1986,  hawkins-performance-1994}.
This was confirmed by \citet[][SI Appendix, Table S15]{hahn2020universals} using word order grammars optimized for DL on 51 UD languages.\footnote{We note that the predictions of DL for three of the correlations (1, 2, 6) are affected by specific properties of the Universal Dependencies format that deviate from the psycholinguistic theories underlying DL \citep{gibson-linguistic-1998,lewis-activation-based-2005} and from some other syntactic theories \citep{Gerdes2018SUDOS,Osborne2019TheSO}.
\citet{hahn2020universals} followed \citet{futrell-large-scale-2015} in measuring dependency length in terms of a converted representation closer to those other theories; such a representation format is necessary to derive correlations 1, 2, 6 from DL \citep{frazier1985syntactic, rijkhoff-word-1986, hawkins-performance-1994}.
In contrast, IL predicts Greenberg's correlations irrespective of these modeling assumptions, as it does not directly refer to syntactic structures.
}
Here, we show that IL (and IL+DL) also predict these universals.
Figure~\ref{table:corr-resu} shows the eight correlations as formalized in the Universal Dependencies format by \citet{hahn2020universals}.
Results show that optimization for IL and IL+DL each predicts all of the correlations to hold in the majority of optimized grammars.
This shows that, unlike in basic word order, the predictions of IL and DL converge on the Greenberg correlation universals, and explains why these tend to hold across languages, whereas basic word order is much more variable.

\citet{hahn2020universals} further argued that the Greenberg correlation universals can be derived from a principle of communicative efficiency closely related to efficiency principles that have found success in other domains of language \citep[e.g.][]{ferreri2003least,xu2016historical,zaslavsky2018efficient,Zaslavsky2021LetsT,Mollicae2025993118}, balancing predictability with parseability, noting that optimizing communicative effiency also leads to efficiency in DL.
We believe that communicative efficiency might be seen best as a possible justification of DL rather than being an orthogonal pressure.
Evaluating the grammars optimized by \citet{hahn2020universals} for communicative efficiency on 51 languages, we found that they exhibit evidence for coadaptation, but overpredict SVO in a way very similar to grammars optimized solely for DL. 

\newpage\section{Raw and Interpolated Efficiency Plane  per Language}\label{sec:per-lang}
Here, we report per-language results for the efficiency planes.
For each language, we first plot both the set of grammar samples, including both randomly constructed baseline grammars, and approximately optimized grammars inhabiting the area close to the Pareto frontier.
These are colored depending on their subject-object position congruence, which is either 0 (green) or 1 (yellow).
The red dot denotes the position of the real observed orderings.
Second, we plot the interpolated average subject-object position congruence throughout the entire efficiency plane, the interpolated approximate Pareto frontier, and the distribution of randomly generated baseline grammars.

Note that some languages are beyond the approximate Pareto frontier; this can happen both because the optimization algorithm is approximate, and because real orderings are not subject to the same representational constraints as the grammars, enabling them to potentially be more efficient than is possible in the grammar formalism (see Section~\ref{sec:fitted}).

\begin{longtable}{llccccccccccc}
	Language &ISO Code & Samples & Interpolated  \\ \hline
Afrikaans &
afr &
\includegraphics[draft=false,width=0.1\textwidth]{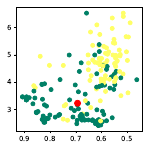} &
\includegraphics[draft=false,width=0.1\textwidth]{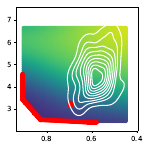} \\
Akkadian &
akk &
\includegraphics[draft=false,width=0.1\textwidth]{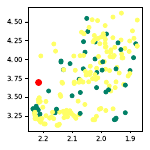} &
\includegraphics[draft=false,width=0.1\textwidth]{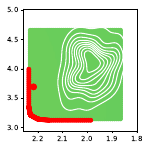} \\
Amharic &
amh &
\includegraphics[draft=false,width=0.1\textwidth]{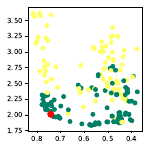} &
\includegraphics[draft=false,width=0.1\textwidth]{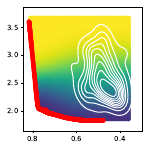} \\
Ancient Greek &
grc &
\includegraphics[draft=false,width=0.1\textwidth]{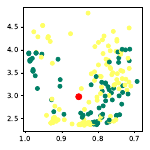} &
\includegraphics[draft=false,width=0.1\textwidth]{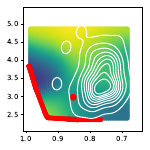} \\
Arabic &
arb &
\includegraphics[draft=false,width=0.1\textwidth]{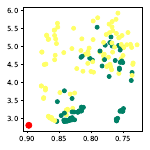} &
\includegraphics[draft=false,width=0.1\textwidth]{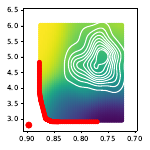} \\
Armenian &
hye &
\includegraphics[draft=false,width=0.1\textwidth]{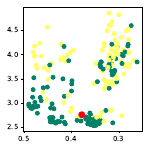} &
\includegraphics[draft=false,width=0.1\textwidth]{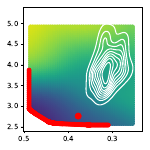} \\
Bambara &
bam &
\includegraphics[draft=false,width=0.1\textwidth]{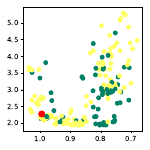} &
\includegraphics[draft=false,width=0.1\textwidth]{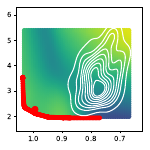} \\
Basque &
eus &
\includegraphics[draft=false,width=0.1\textwidth]{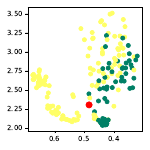} &
\includegraphics[draft=false,width=0.1\textwidth]{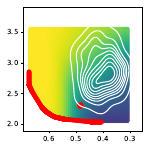} \\
Belarusian &
bel &
\includegraphics[draft=false,width=0.1\textwidth]{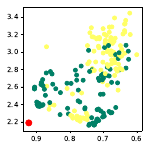} &
\includegraphics[draft=false,width=0.1\textwidth]{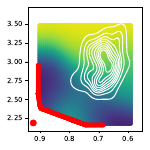} \\
Breton &
bre &
\includegraphics[draft=false,width=0.1\textwidth]{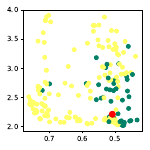} &
\includegraphics[draft=false,width=0.1\textwidth]{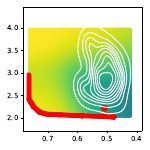} \\
Bulgarian &
bul &
\includegraphics[draft=false,width=0.1\textwidth]{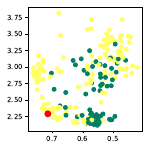} &
\includegraphics[draft=false,width=0.1\textwidth]{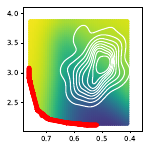} \\
Buryat &
bua &
\includegraphics[draft=false,width=0.1\textwidth]{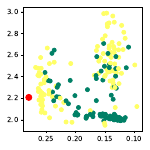} &
\includegraphics[draft=false,width=0.1\textwidth]{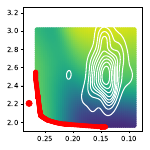} \\
Cantonese &
yue &
\includegraphics[draft=false,width=0.1\textwidth]{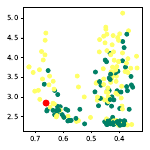} &
\includegraphics[draft=false,width=0.1\textwidth]{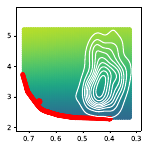} \\
Catalan &
cat &
\includegraphics[draft=false,width=0.1\textwidth]{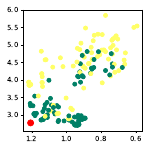} &
\includegraphics[draft=false,width=0.1\textwidth]{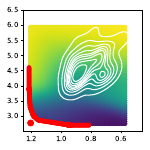} \\
Chinese &
cmn &
\includegraphics[draft=false,width=0.1\textwidth]{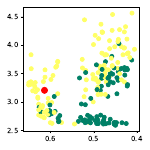} &
\includegraphics[draft=false,width=0.1\textwidth]{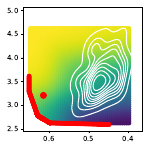} \\
Classical Chinese &
lzh &
\includegraphics[draft=false,width=0.1\textwidth]{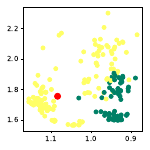} &
\includegraphics[draft=false,width=0.1\textwidth]{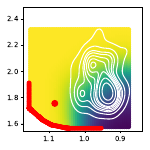} \\
Coptic &
cop &
\includegraphics[draft=false,width=0.1\textwidth]{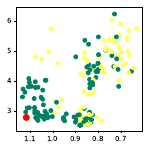} &
\includegraphics[draft=false,width=0.1\textwidth]{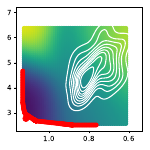} \\
Croatian &
hrv &
\includegraphics[draft=false,width=0.1\textwidth]{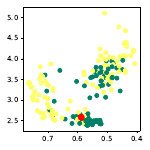} &
\includegraphics[draft=false,width=0.1\textwidth]{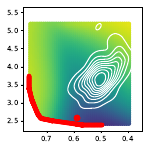} \\
Czech &
ces &
\includegraphics[draft=false,width=0.1\textwidth]{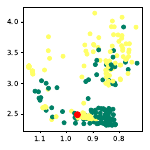} &
\includegraphics[draft=false,width=0.1\textwidth]{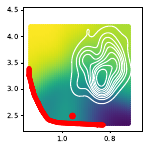} \\
Danish &
dan &
\includegraphics[draft=false,width=0.1\textwidth]{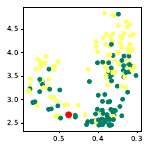} &
\includegraphics[draft=false,width=0.1\textwidth]{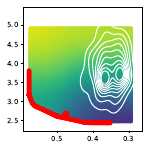} \\
Dutch &
nld &
\includegraphics[draft=false,width=0.1\textwidth]{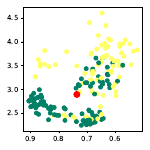} &
\includegraphics[draft=false,width=0.1\textwidth]{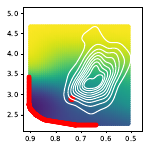} \\
English &
eng &
\includegraphics[draft=false,width=0.1\textwidth]{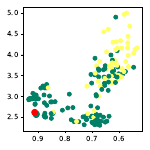} &
\includegraphics[draft=false,width=0.1\textwidth]{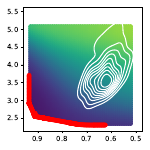} \\
Erzya &
myv &
\includegraphics[draft=false,width=0.1\textwidth]{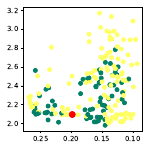} &
\includegraphics[draft=false,width=0.1\textwidth]{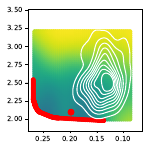} \\
Estonian &
est &
\includegraphics[draft=false,width=0.1\textwidth]{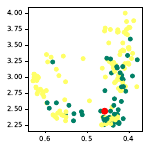} &
\includegraphics[draft=false,width=0.1\textwidth]{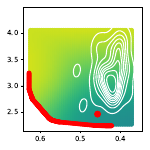} \\
Faroese &
fao &
\includegraphics[draft=false,width=0.1\textwidth]{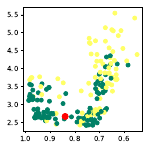} &
\includegraphics[draft=false,width=0.1\textwidth]{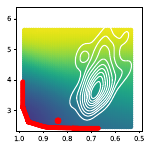} \\
Finnish &
fin &
\includegraphics[draft=false,width=0.1\textwidth]{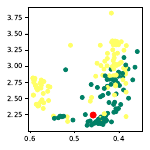} &
\includegraphics[draft=false,width=0.1\textwidth]{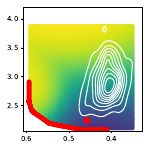} \\
French &
fra &
\includegraphics[draft=false,width=0.1\textwidth]{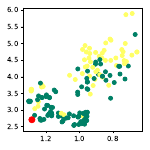} &
\includegraphics[draft=false,width=0.1\textwidth]{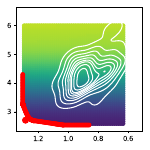} \\
Galician &
glg &
\includegraphics[draft=false,width=0.1\textwidth]{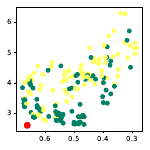} &
\includegraphics[draft=false,width=0.1\textwidth]{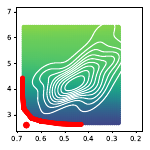} \\
German &
deu &
\includegraphics[draft=false,width=0.1\textwidth]{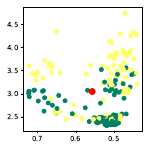} &
\includegraphics[draft=false,width=0.1\textwidth]{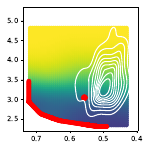} \\
Gothic &
got &
\includegraphics[draft=false,width=0.1\textwidth]{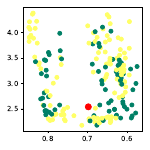} &
\includegraphics[draft=false,width=0.1\textwidth]{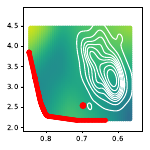} \\
Greek &
ell &
\includegraphics[draft=false,width=0.1\textwidth]{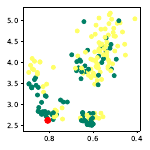} &
\includegraphics[draft=false,width=0.1\textwidth]{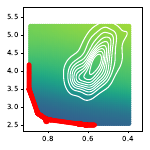} \\
Hebrew &
heb &
\includegraphics[draft=false,width=0.1\textwidth]{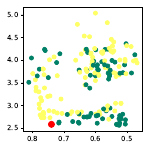} &
\includegraphics[draft=false,width=0.1\textwidth]{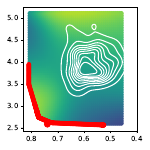} \\
Hindi &
hin &
\includegraphics[draft=false,width=0.1\textwidth]{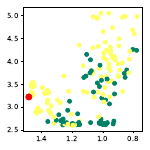} &
\includegraphics[draft=false,width=0.1\textwidth]{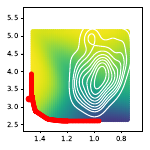} \\
Hungarian &
hun &
\includegraphics[draft=false,width=0.1\textwidth]{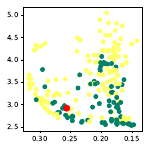} &
\includegraphics[draft=false,width=0.1\textwidth]{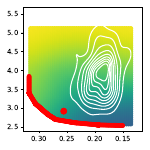} \\
Icelandic &
isl &
\includegraphics[draft=false,width=0.1\textwidth]{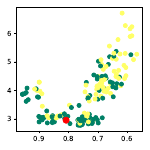} &
\includegraphics[draft=false,width=0.1\textwidth]{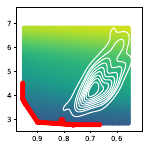} \\
Indonesian &
ind &
\includegraphics[draft=false,width=0.1\textwidth]{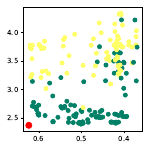} &
\includegraphics[draft=false,width=0.1\textwidth]{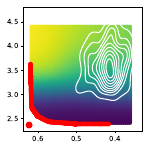} \\
Irish &
gle &
\includegraphics[draft=false,width=0.1\textwidth]{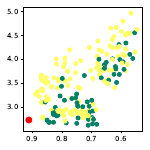} &
\includegraphics[draft=false,width=0.1\textwidth]{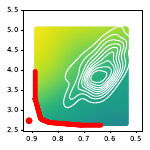} \\
Italian &
ita &
\includegraphics[draft=false,width=0.1\textwidth]{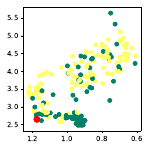} &
\includegraphics[draft=false,width=0.1\textwidth]{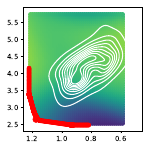} \\
Japanese &
jpn &
\includegraphics[draft=false,width=0.1\textwidth]{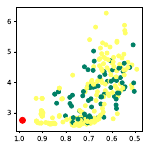} &
\includegraphics[draft=false,width=0.1\textwidth]{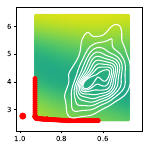} \\
Kazakh &
kaz &
\includegraphics[draft=false,width=0.1\textwidth]{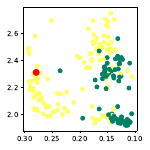} &
\includegraphics[draft=false,width=0.1\textwidth]{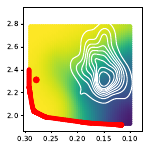} \\
Kiche &
quc &
\includegraphics[draft=false,width=0.1\textwidth]{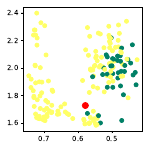} &
\includegraphics[draft=false,width=0.1\textwidth]{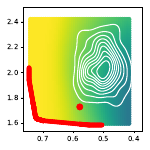} \\
Komi Zyrian &
kpv &
\includegraphics[draft=false,width=0.1\textwidth]{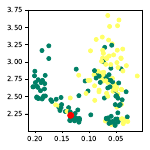} &
\includegraphics[draft=false,width=0.1\textwidth]{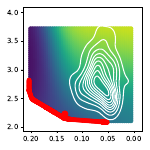} \\
Korean &
kor &
\includegraphics[draft=false,width=0.1\textwidth]{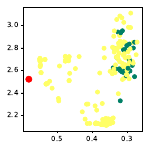} &
\includegraphics[draft=false,width=0.1\textwidth]{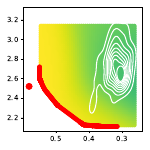} \\
Kurmanji &
kmr &
\includegraphics[draft=false,width=0.1\textwidth]{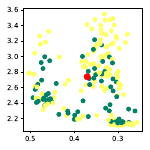} &
\includegraphics[draft=false,width=0.1\textwidth]{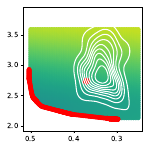} \\
Latin &
lat &
\includegraphics[draft=false,width=0.1\textwidth]{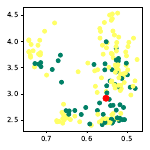} &
\includegraphics[draft=false,width=0.1\textwidth]{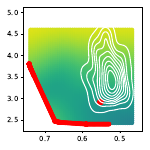} \\
Latvian &
lav &
\includegraphics[draft=false,width=0.1\textwidth]{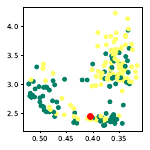} &
\includegraphics[draft=false,width=0.1\textwidth]{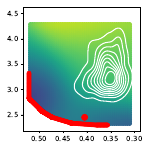} \\
Lithuanian &
lit &
\includegraphics[draft=false,width=0.1\textwidth]{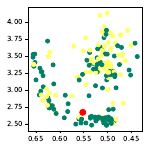} &
\includegraphics[draft=false,width=0.1\textwidth]{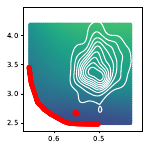} \\
Maltese &
mlt &
\includegraphics[draft=false,width=0.1\textwidth]{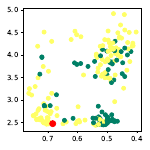} &
\includegraphics[draft=false,width=0.1\textwidth]{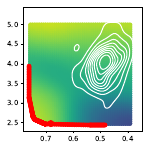} \\
Manx &
glv &
\includegraphics[draft=false,width=0.1\textwidth]{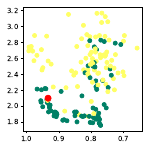} &
\includegraphics[draft=false,width=0.1\textwidth]{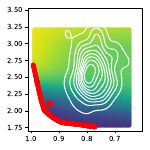} \\
Mbya Guarani &
gun &
\includegraphics[draft=false,width=0.1\textwidth]{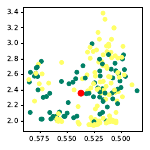} &
\includegraphics[draft=false,width=0.1\textwidth]{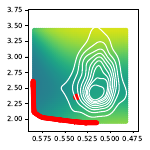} \\
Naija &
pcm &
\includegraphics[draft=false,width=0.1\textwidth]{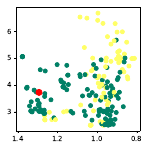} &
\includegraphics[draft=false,width=0.1\textwidth]{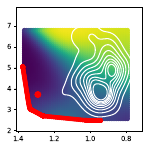} \\
North Sami &
sme &
\includegraphics[draft=false,width=0.1\textwidth]{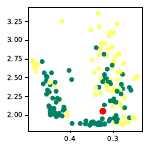} &
\includegraphics[draft=false,width=0.1\textwidth]{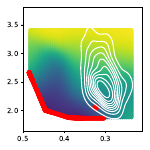} \\
Norwegian &
nob &
\includegraphics[draft=false,width=0.1\textwidth]{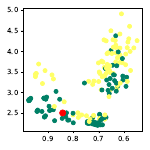} &
\includegraphics[draft=false,width=0.1\textwidth]{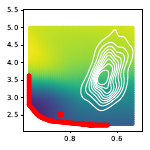} \\
Old Church Slavonic &
chu &
\includegraphics[draft=false,width=0.1\textwidth]{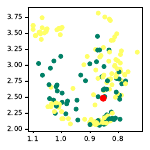} &
\includegraphics[draft=false,width=0.1\textwidth]{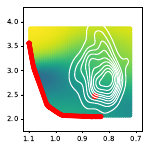} \\
Old East Slavic &
orv &
\includegraphics[draft=false,width=0.1\textwidth]{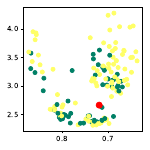} &
\includegraphics[draft=false,width=0.1\textwidth]{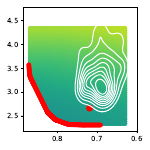} \\
Old French &
fro &
\includegraphics[draft=false,width=0.1\textwidth]{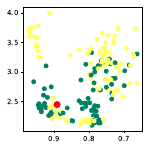} &
\includegraphics[draft=false,width=0.1\textwidth]{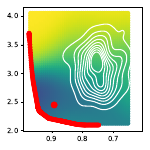} \\
Persian &
pes &
\includegraphics[draft=false,width=0.1\textwidth]{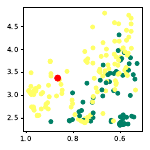} &
\includegraphics[draft=false,width=0.1\textwidth]{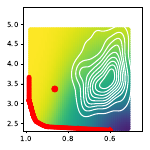} \\
Polish &
pol &
\includegraphics[draft=false,width=0.1\textwidth]{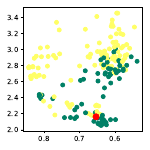} &
\includegraphics[draft=false,width=0.1\textwidth]{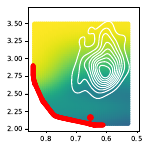} \\
Portuguese &
por &
\includegraphics[draft=false,width=0.1\textwidth]{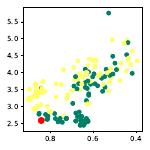} &
\includegraphics[draft=false,width=0.1\textwidth]{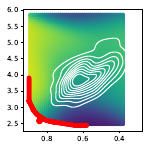} \\
Romanian &
ron &
\includegraphics[draft=false,width=0.1\textwidth]{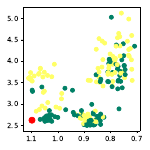} &
\includegraphics[draft=false,width=0.1\textwidth]{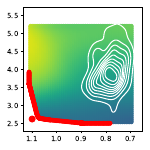} \\
Russian &
rus &
\includegraphics[draft=false,width=0.1\textwidth]{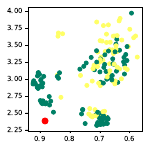} &
\includegraphics[draft=false,width=0.1\textwidth]{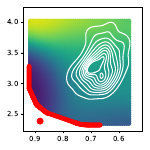} \\
Sanskrit &
san &
\includegraphics[draft=false,width=0.1\textwidth]{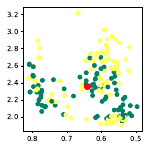} &
\includegraphics[draft=false,width=0.1\textwidth]{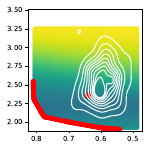} \\
Scottish Gaelic &
gla &
\includegraphics[draft=false,width=0.1\textwidth]{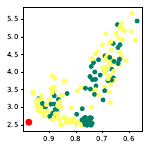} &
\includegraphics[draft=false,width=0.1\textwidth]{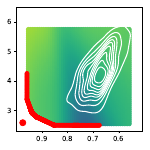} \\
Serbian &
srp &
\includegraphics[draft=false,width=0.1\textwidth]{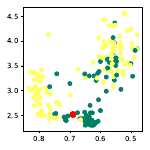} &
\includegraphics[draft=false,width=0.1\textwidth]{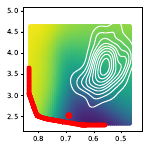} \\
Slovak &
slk &
\includegraphics[draft=false,width=0.1\textwidth]{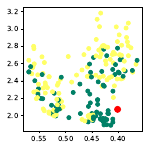} &
\includegraphics[draft=false,width=0.1\textwidth]{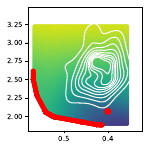} \\
Slovenian &
slv &
\includegraphics[draft=false,width=0.1\textwidth]{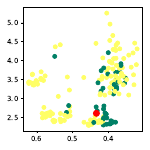} &
\includegraphics[draft=false,width=0.1\textwidth]{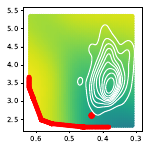} \\
Spanish &
spa &
\includegraphics[draft=false,width=0.1\textwidth]{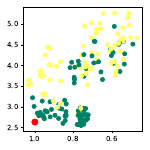} &
\includegraphics[draft=false,width=0.1\textwidth]{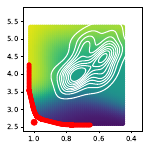} \\
Swedish &
swe &
\includegraphics[draft=false,width=0.1\textwidth]{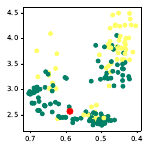} &
\includegraphics[draft=false,width=0.1\textwidth]{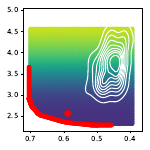} \\
Tamil &
tam &
\includegraphics[draft=false,width=0.1\textwidth]{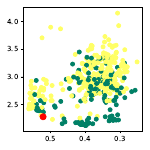} &
\includegraphics[draft=false,width=0.1\textwidth]{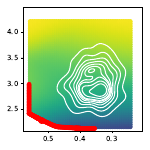} \\
Thai &
tha &
\includegraphics[draft=false,width=0.1\textwidth]{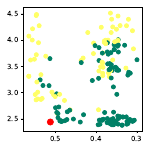} &
\includegraphics[draft=false,width=0.1\textwidth]{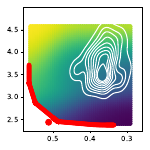} \\
Turkish &
tur &
\includegraphics[draft=false,width=0.1\textwidth]{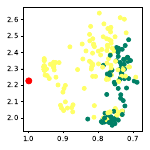} &
\includegraphics[draft=false,width=0.1\textwidth]{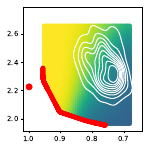} \\
Ukrainian &
ukr &
\includegraphics[draft=false,width=0.1\textwidth]{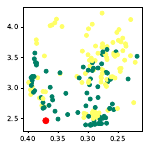} &
\includegraphics[draft=false,width=0.1\textwidth]{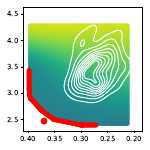} \\
Upper Sorbian &
hsb &
\includegraphics[draft=false,width=0.1\textwidth]{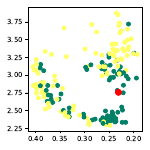} &
\includegraphics[draft=false,width=0.1\textwidth]{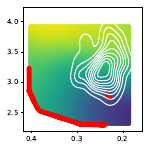} \\
Urdu &
urd &
\includegraphics[draft=false,width=0.1\textwidth]{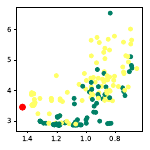} &
\includegraphics[draft=false,width=0.1\textwidth]{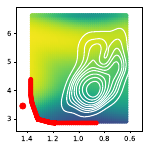} \\
Uyghur &
uig &
\includegraphics[draft=false,width=0.1\textwidth]{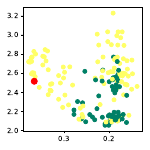} &
\includegraphics[draft=false,width=0.1\textwidth]{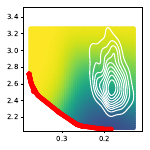} \\
Vietnamese &
vie &
\includegraphics[draft=false,width=0.1\textwidth]{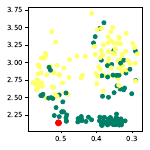} &
\includegraphics[draft=false,width=0.1\textwidth]{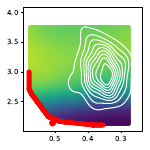} \\
Welsh &
cym &
\includegraphics[draft=false,width=0.1\textwidth]{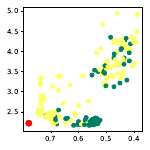} &
\includegraphics[draft=false,width=0.1\textwidth]{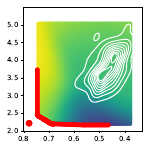} \\
Western Armenian &
hye2 &
\includegraphics[draft=false,width=0.1\textwidth]{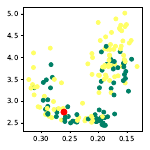} &
\includegraphics[draft=false,width=0.1\textwidth]{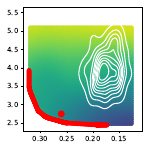} \\
Wolof &
wol &
\includegraphics[draft=false,width=0.1\textwidth]{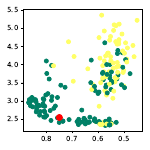} &
\includegraphics[draft=false,width=0.1\textwidth]{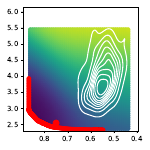} \\
Xibe &
sjo &
\includegraphics[draft=false,width=0.1\textwidth]{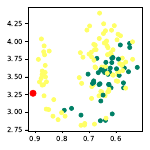} &
\includegraphics[draft=false,width=0.1\textwidth]{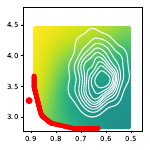} \\
\end{longtable}

\newpage\section{Neural Network Estimates of Information Locality}\label{sec:neural}

Here, we compare our formalization and estimation method for information locality to the method used by \citet{Hahn2020modeling}, which is based on neural language models, estimating the next-word predictive distribution using LSTM recurrent neural networks \citep{hochreiter-long-1997}.
Compared to the method used here, the use of recurrent neural networks can potentially result in better modeling of longer-range statistical relationships, and we asked whether this impacts our estimates of IL.
Running the estimation method used by \citet{Hahn2020modeling} on all 80 languages was not feasible due to the high computational cost of neural network estimators.\footnote{\citet{Hahn2020modeling} estimated information locality for 10--20 grammars in 54 languages. In contrast, we have $\approx$ 150 approximately optimized grammars for each of 80 languages.}
We thus selected twelve languages representing typological, genetic, and geographic diversity within the bounds afforded by the UD data, and particularly where the Pareto frontier shows variability in subject-object position congruence:

\begin{enumerate}
\item Arabic (VSO, Afro-Asiatic, Asia/Africa) 
\item Basque (SOV, isolate, European)
\item Chinese (SVO, Sino-Tibetan, Asia) 
\item English (SVO, Indo-European, European)
\item Finnish (SVO, Uralic, European)
\item Hindi (SOV, Indo-European, Asia)
\item Indonesian (SVO, Austronesian, Asia)
\item Persian (SOV, Indo-European, Asia)
\item Polish (SVO, Indo-European, Europe) 
\item Thai (SVO, Tai-Kadai, Asia)
\item Turkish (SOV, Turkic, Asia) 
\item Wolof (SVO, Niger-Congo, Africa)
\end{enumerate}

We used the neural network-based method of \citet{Hahn2020modeling} to compute both $I_1$ and the AUC measure they used to quantify IL (see Section~\ref{sec:info-locality}), for the approximately optimized grammars, the real orderings, and for $\geq 30$ randomly constructed baseline grammars per language.\footnote{The baselines are different from those used in the other studies, as we had not recorded the baseline grammars, only their IL/DL values.}
For the LSTM network, we used the hyperparameters that they had determined for each of the 12 languages to minimize surprisal on random baseline grammars.

Here, we show for each of the twelve languages the efficiency plane, with IL represented by (i) $I_1$ as computed by our method (Section~\ref{sec:estimating-mi}), (ii) $I_1$ as computed using neural language models, and (iii) the AUC measure also computed using neural language models.
For comparability across the three methods, we normalize DL and IL as in the main paper.
Results are very similar, in the shape of the Pareto frontier, in the position of the real ordering, and in the distribution of subject-object position congruence throughout the efficiency plane.

\phantom{.}

\phantom{.}

\begin{longtable}{llcccccc}
	Language & $I_1$ (Our) & $I_1$ (Neural) & AUC (Neural) \\ \hline
Arabic &
\includegraphics[draft=false,width=0.1\textwidth]{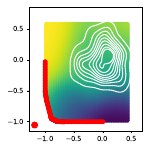} &
	\includegraphics[draft=false,width=0.1\textwidth]{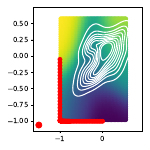} &
\includegraphics[draft=false,width=0.1\textwidth]{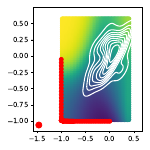} \\
Basque &
\includegraphics[draft=false,width=0.1\textwidth]{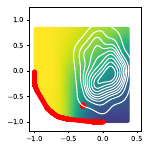} &
	\includegraphics[draft=false,width=0.1\textwidth]{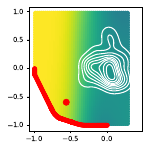} &
\includegraphics[draft=false,width=0.1\textwidth]{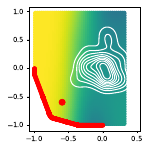} 
	\\
Chinese &
\includegraphics[draft=false,width=0.1\textwidth]{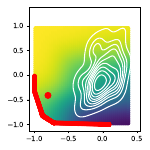} &
	\includegraphics[draft=false,width=0.1\textwidth]{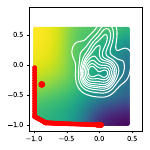} &
	\includegraphics[draft=false,width=0.1\textwidth]{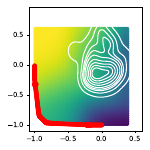} 
	\\
English &
\includegraphics[draft=false,width=0.1\textwidth]{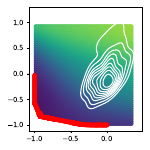} &
	\includegraphics[draft=false,width=0.1\textwidth]{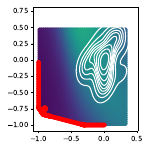} &
	\includegraphics[draft=false,width=0.1\textwidth]{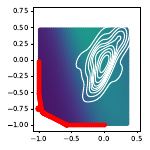} 
	\\
Finnish &
\includegraphics[draft=false,width=0.1\textwidth]{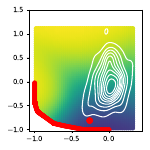} &
	\includegraphics[draft=false,width=0.1\textwidth]{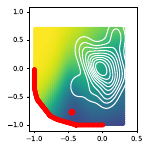} &
	\includegraphics[draft=false,width=0.1\textwidth]{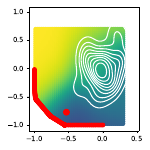} 
	\\
Hindi &
\includegraphics[draft=false,width=0.1\textwidth]{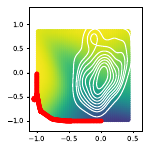} &
	\includegraphics[draft=false,width=0.1\textwidth]{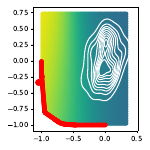} &
	\includegraphics[draft=false,width=0.1\textwidth]{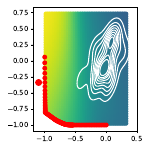} 
	\\
Indonesian &
\includegraphics[draft=false,width=0.1\textwidth]{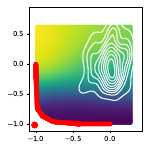} &
	\includegraphics[draft=false,width=0.1\textwidth]{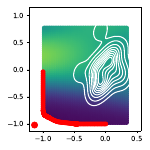} &
	\includegraphics[draft=false,width=0.1\textwidth]{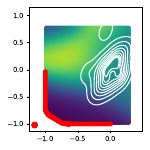} 
	\\
Persian &
\includegraphics[draft=false,width=0.1\textwidth]{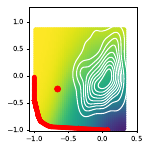} &
	\includegraphics[draft=false,width=0.1\textwidth]{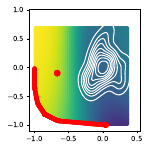} &
	\includegraphics[draft=false,width=0.1\textwidth]{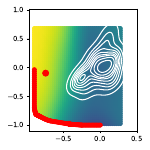} 
	\\
Polish &
\includegraphics[draft=false,width=0.1\textwidth]{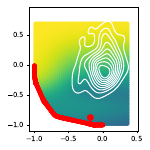} &
	\includegraphics[draft=false,width=0.1\textwidth]{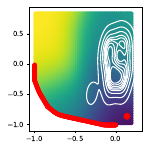} &
	\includegraphics[draft=false,width=0.1\textwidth]{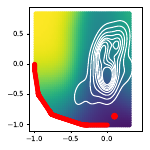} 
	\\
Thai &
\includegraphics[draft=false,width=0.1\textwidth]{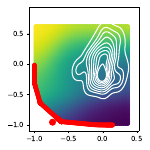} &
	\includegraphics[draft=false,width=0.1\textwidth]{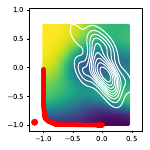} &
	\includegraphics[draft=false,width=0.1\textwidth]{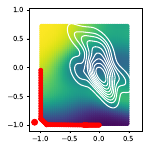} 
	\\
Turkish &
\includegraphics[draft=false,width=0.1\textwidth]{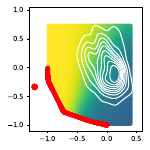} &
	\includegraphics[draft=false,width=0.1\textwidth]{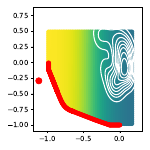} &
	\includegraphics[draft=false,width=0.1\textwidth]{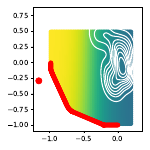} 
	\\
Wolof &
\includegraphics[draft=false,width=0.1\textwidth]{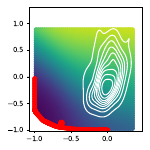} &
	\includegraphics[draft=false,width=0.1\textwidth]{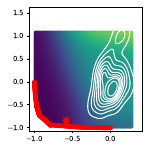} &
	\includegraphics[draft=false,width=0.1\textwidth]{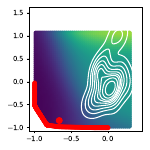} 
	\\
\end{longtable}

\newpage\section{Historical Changes}\label{sec:historical-all-chains}

\begin{figure}
	\centering
	\begin{tabular}{ccccccc}
		East Slavic & 	East South Slavic & English \\
		\includegraphics[draft=false,width=0.35\textwidth]{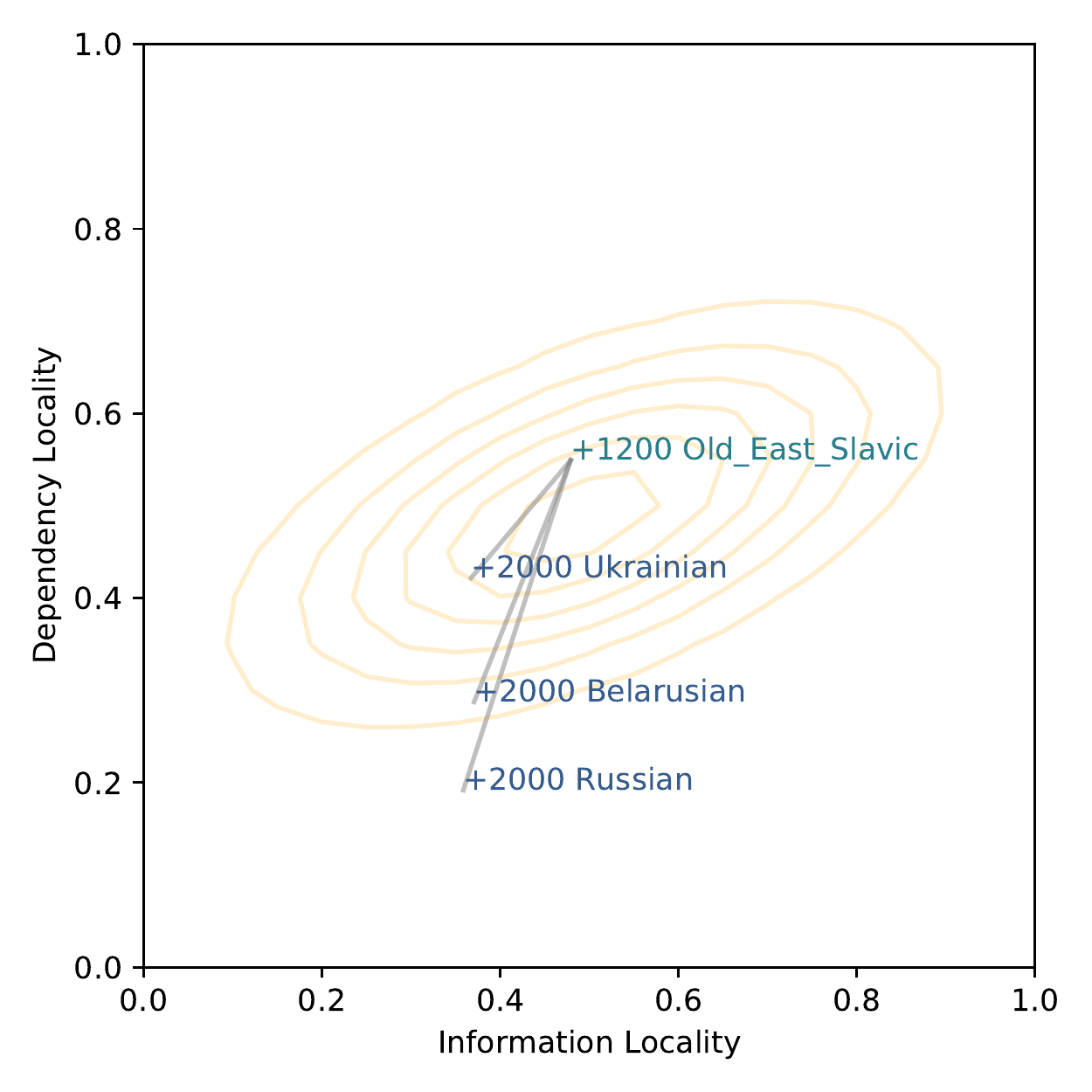}  &
\includegraphics[draft=false,width=0.35\textwidth]{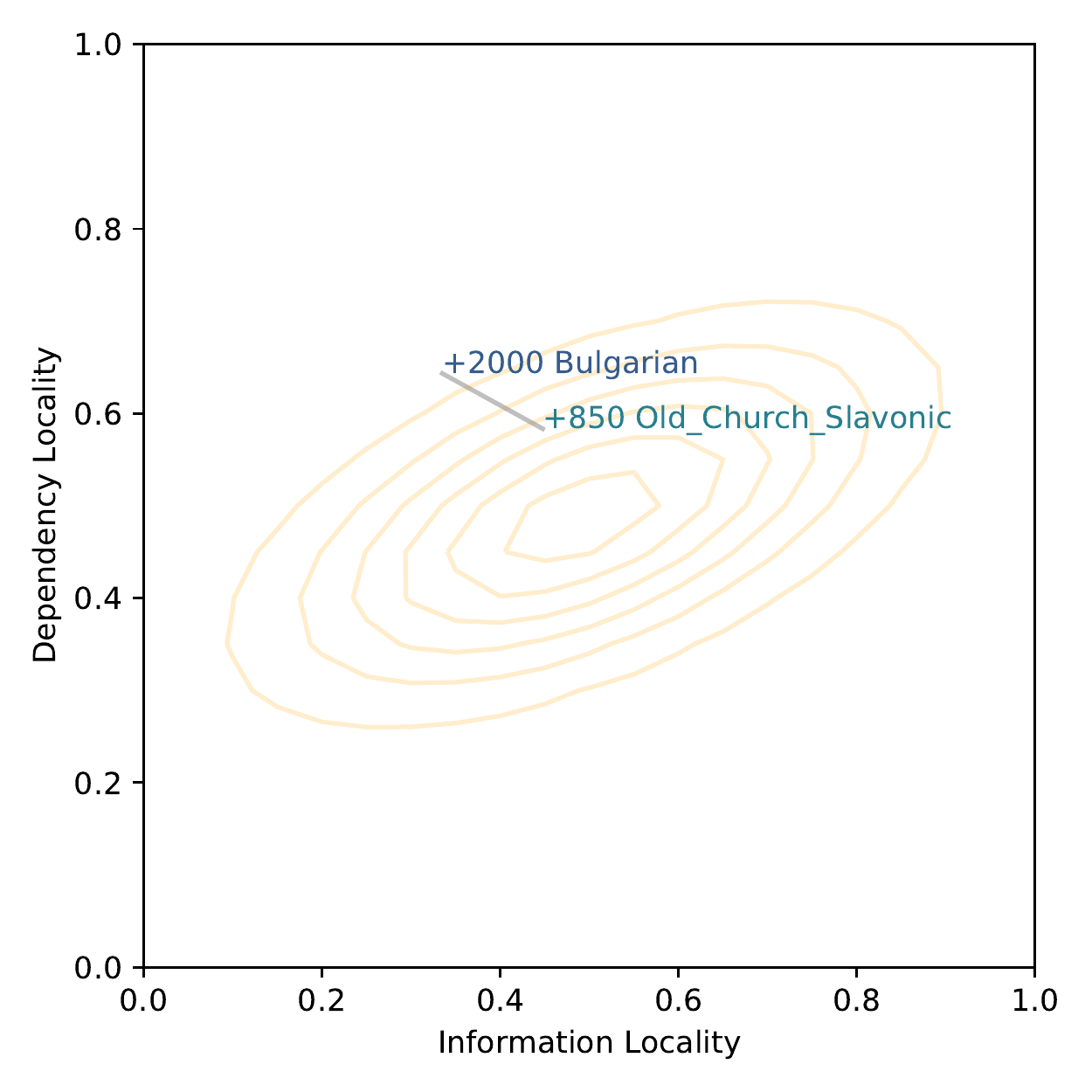} &
		\includegraphics[draft=false,width=0.35\textwidth]{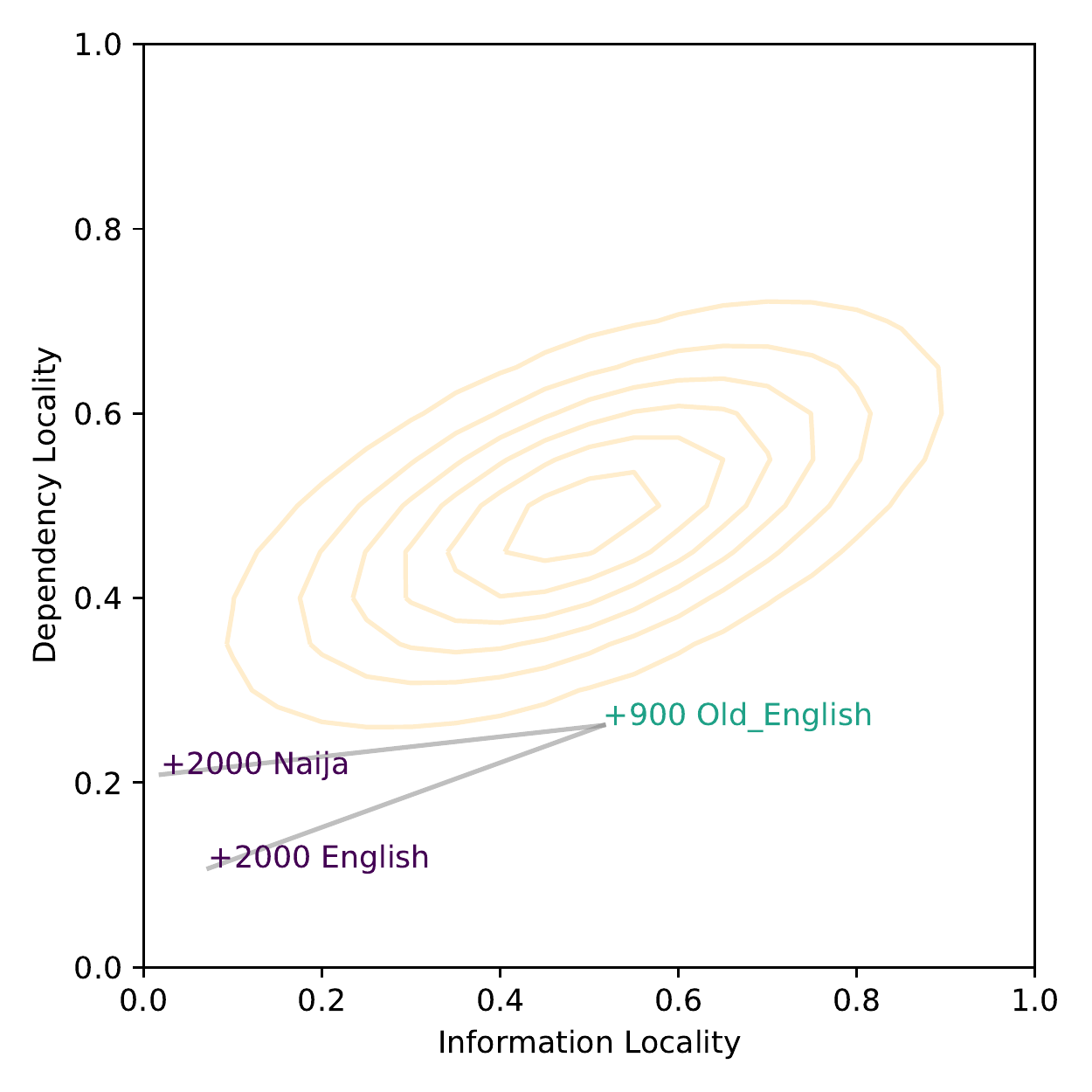}  \\
		Greek & Hindi/Urdu & Icelandic \\
		\includegraphics[draft=false,width=0.35\textwidth]{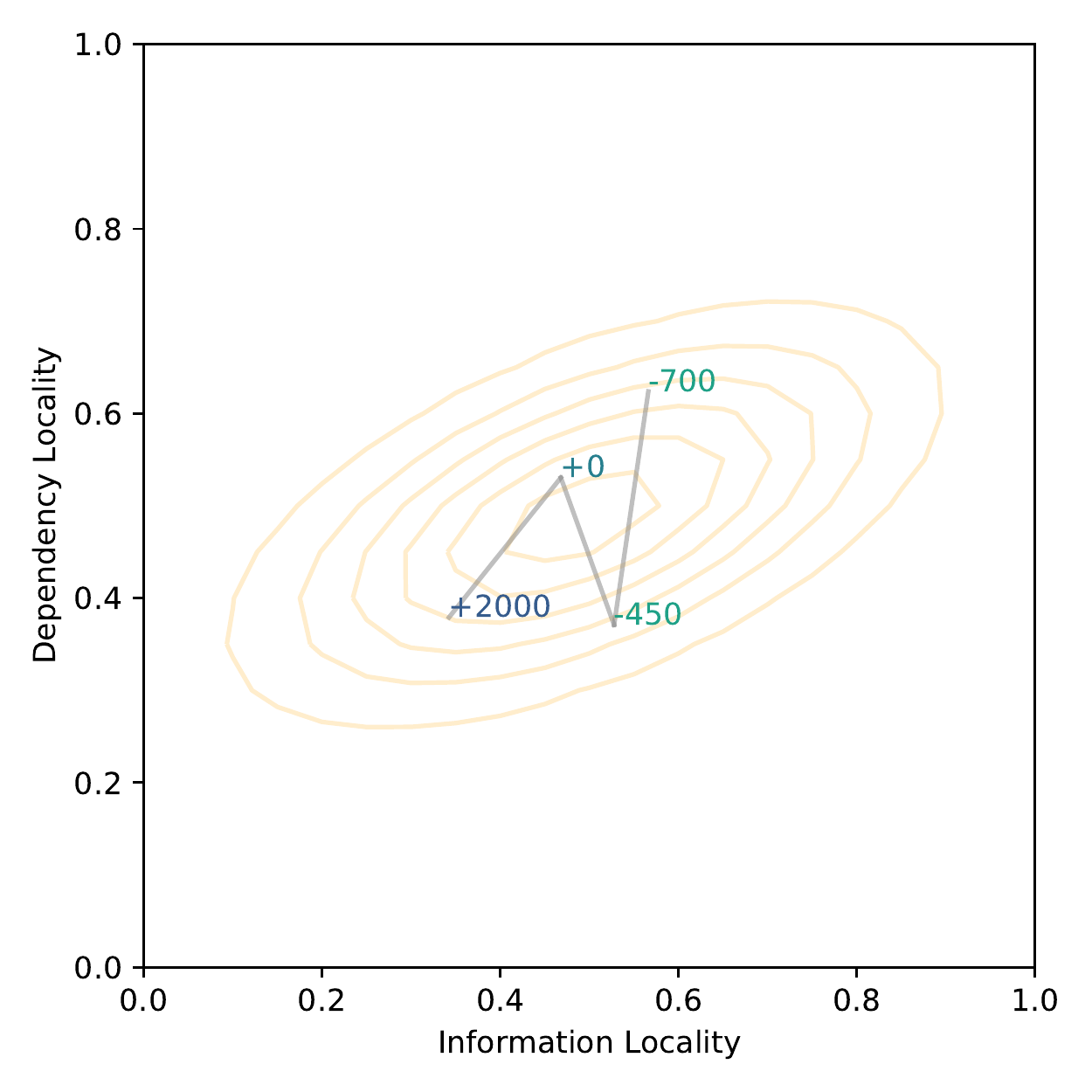}  &
		\includegraphics[draft=false,width=0.35\textwidth]{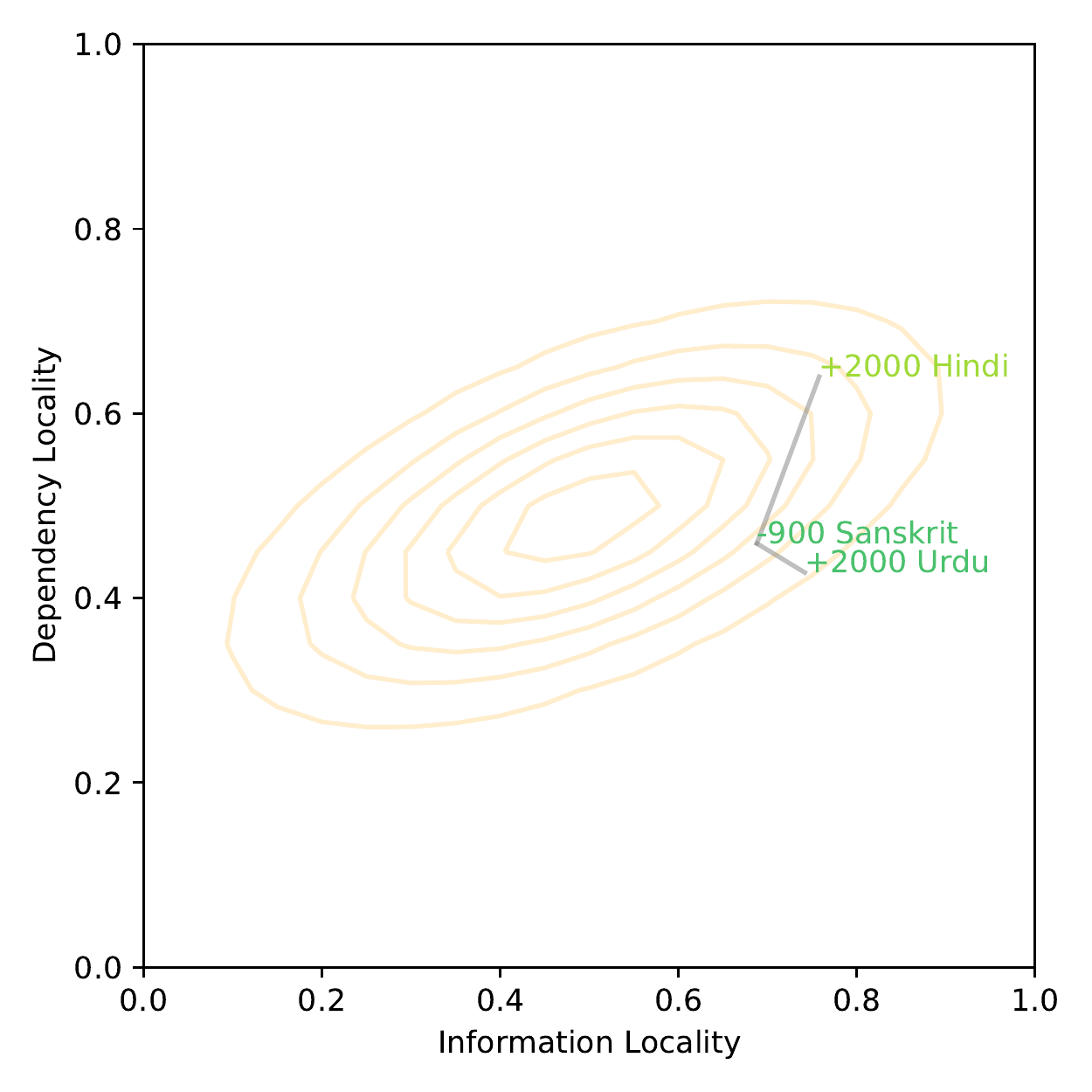} &
		\includegraphics[draft=false,width=0.35\textwidth]{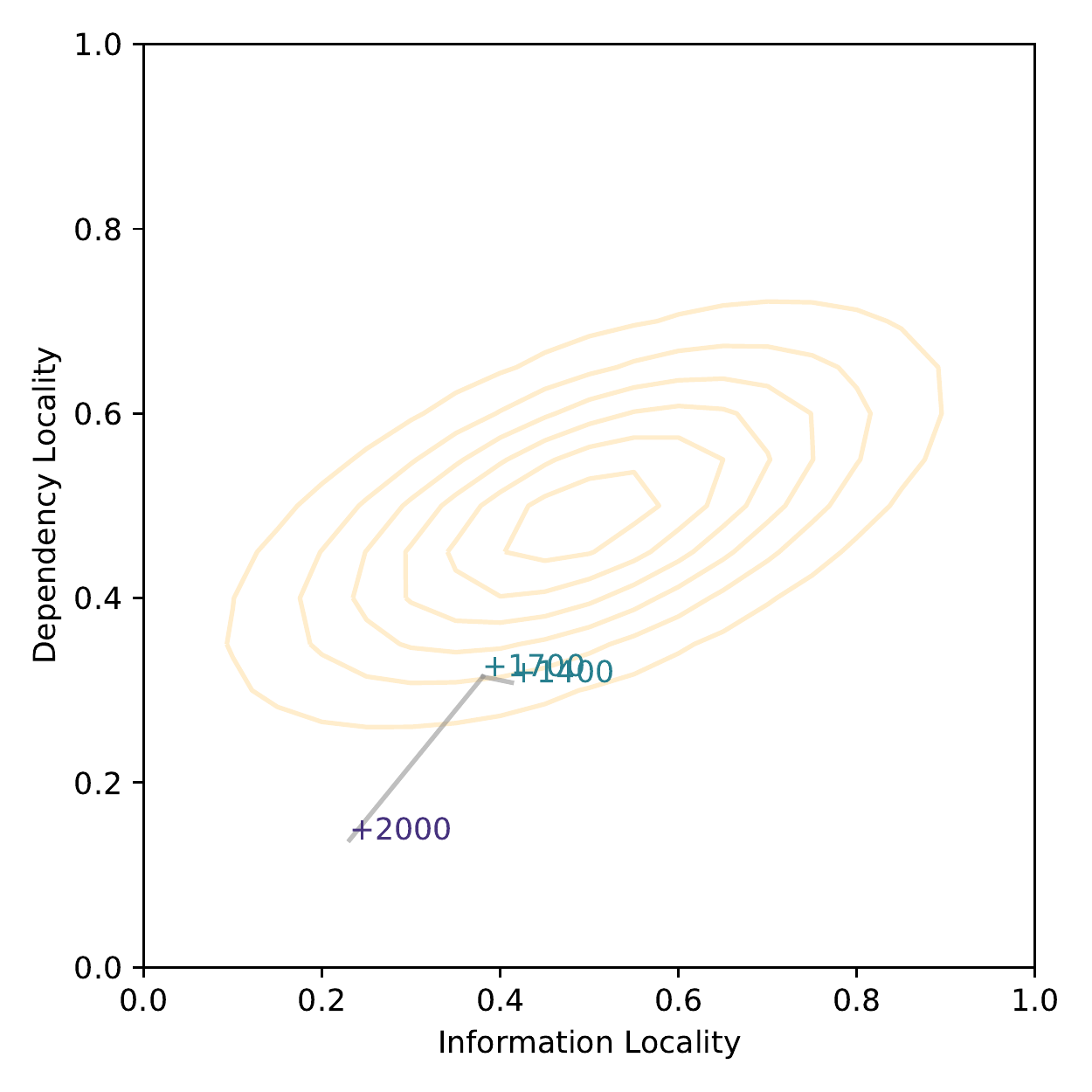}  \\
		Romance & Sinitic \\
		\includegraphics[draft=false,width=0.35\textwidth]{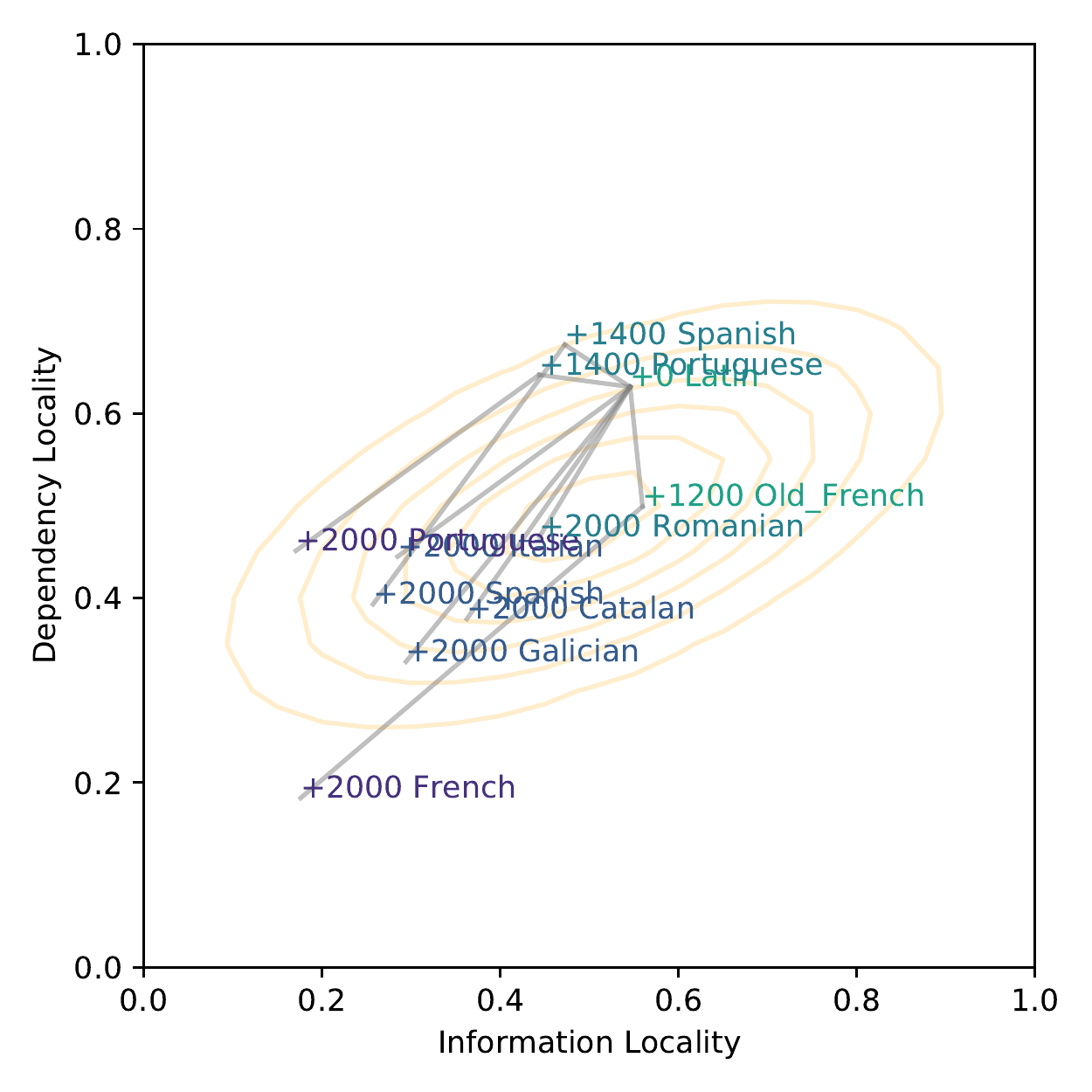}  &
		\includegraphics[draft=false,width=0.35\textwidth]{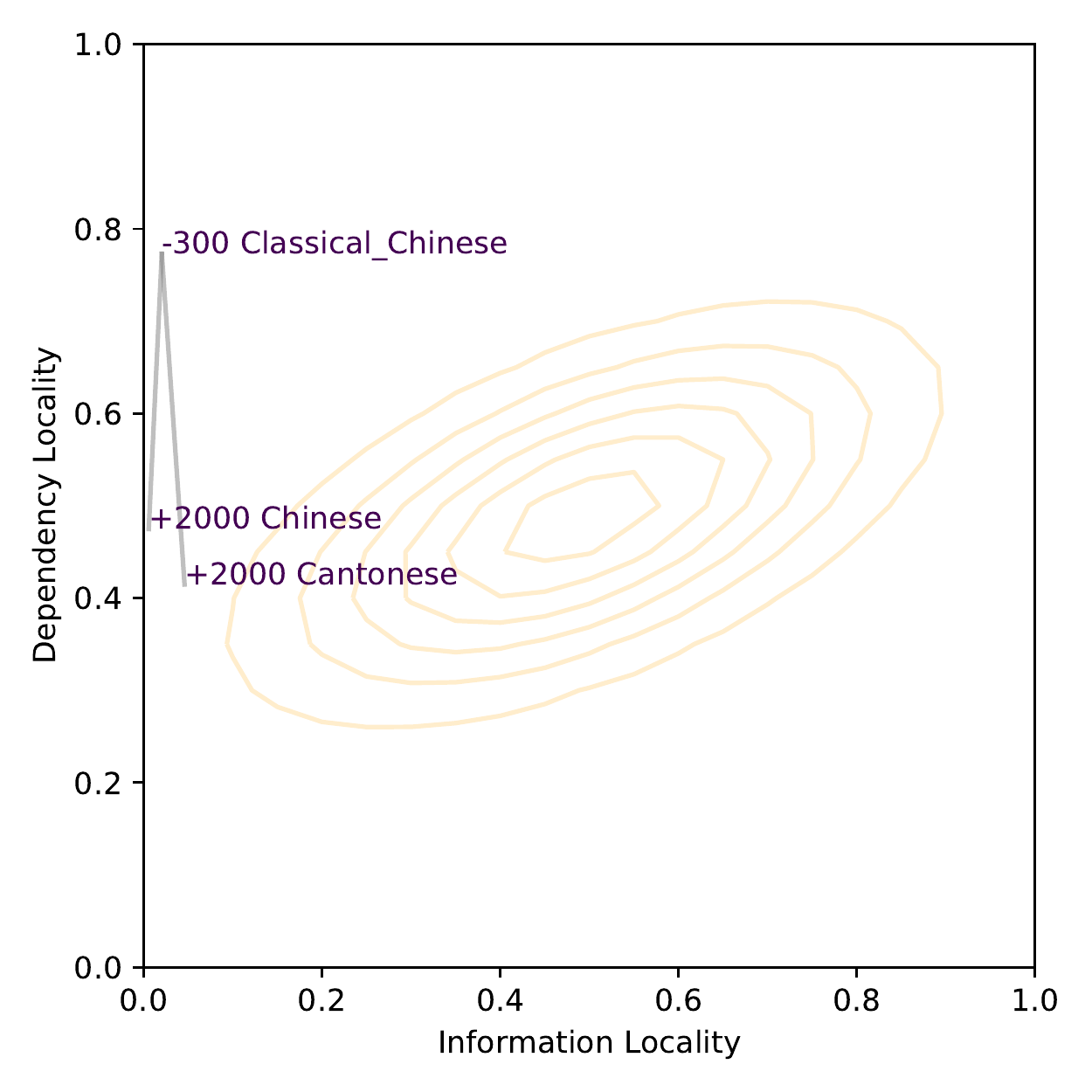} 
	\end{tabular}

	\caption{Historical trajectories of attested and average optimized subject-object position congruence. Faint contours describe the stationary distribution identified by the phylogenetic model. See Section~\ref{sec:historical-all-chains}. }\label{fig:congruence-trajectories}
\end{figure}

Below, we show efficiency planes for all languages that are attested in our dataset at multiple points in time.\footnote{For space and clarity, we only those descendants of Latin with an intermediate attested stage (French, Spanish, Portuguese). Trajectories for languages without data for intermediate stages (Italian, Catalan, Galician, Romanian) are similar; compare Figure~\ref{fig:congruence-trajectories}.} For comparability, we normalize DL and IL as in the main paper.

For each such language, we show the trajectory of subject-object position congruence (left), and the efficiency planes over time (right).
In three cases of closely related languages with essentially identical subject-object position congrunece values, we plot those together (Sinitic, Hindi/Urdu, East Slavic).

In one case (Icelandic), data is available continuously across a word order change (compare Section~\ref{sec:other-historical}); for the others, it is available at two or more points in time.

In terms of word order changes, there are multiple cases of changes towards SVO (lower subject-object position congruence; English, Romance, Icelandic), and also cases with only limited change (Sinitic, Greek, Hindi/Urdu, East and East South Slavic).
In terms of efficiency, languages have moved closely along the frontier (e.g. English, Icelandic) or towards the frontier (e.g., Romance).
Languages with change towards SVO (English, Romance, Icelandic) correspondingly exhibit an increase of SVO-like orderings (dark blue) along the Pareto frontier (compare Figure~\ref{fig:congruence-trajectories}).

\phantom{.}

\phantom{.}

\begin{longtable}{|lcccc|ccccc}\hline
	English	& +900 & +2000 & +2000 & \\
			\multirow{2}{*}{		\includegraphics[draft=false,width=0.2\textwidth]{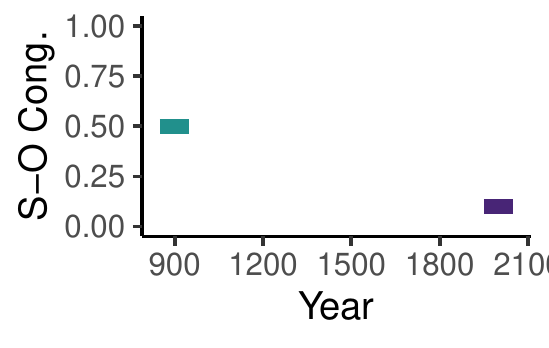}}	& \textit{Old English} & \textit{English} & \textit{Naija (Nigerian Pidgin)} &\\
		& \includegraphics[draft=false,width=0.1\textwidth]{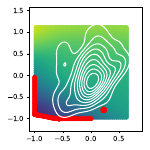}
	& \includegraphics[draft=false,width=0.1\textwidth]{process9_frontier18_Viz.py_English_2.8_smoothed_scaled_noBounds.png-1.png} &
\includegraphics[draft=false,width=0.1\textwidth]{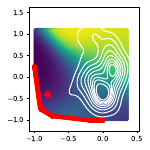}
	&
	\\ \hline
	Romance & +0 & +1200 & +2000 &\\ 
		\multirow{2}{*}{		\includegraphics[draft=false,width=0.2\textwidth]{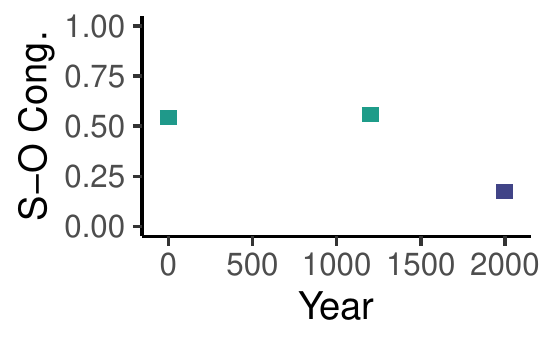}	}& \textit{Latin} & \textit{Old French} & \textit{French} & \\
		& \includegraphics[draft=false,width=0.1\textwidth]{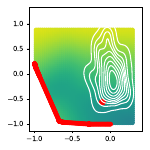} 
	& \includegraphics[draft=false,width=0.1\textwidth]{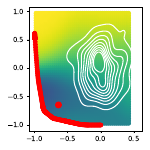} 
	& \includegraphics[draft=false,width=0.1\textwidth]{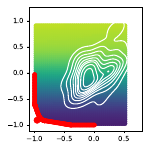} &
	\\
	&& +1400 & +2000 &\\ 
		\multirow{2}{*}{		\includegraphics[draft=false,width=0.2\textwidth]{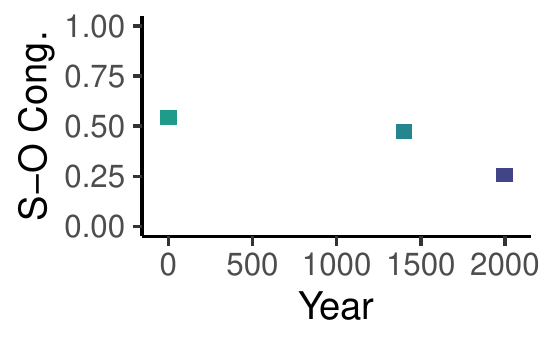}	}	&& \textit{Medieval Spanish} & \textit{Spanish}  &\\
	&
	& \includegraphics[draft=false,width=0.1\textwidth]{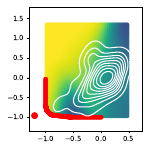} 
	& \includegraphics[draft=false,width=0.1\textwidth]{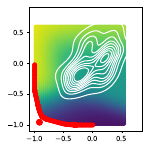}  &
	\\ 
	&& +1400 & +2000 &\\ 
		\multirow{2}{*}{		\includegraphics[draft=false,width=0.2\textwidth]{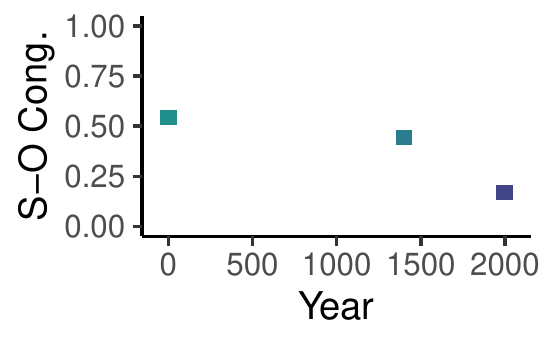}	}	&  & \textit{Medieval Portuguese} & \textit{Portuguese}  &\\

	&& \includegraphics[draft=false,width=0.1\textwidth]{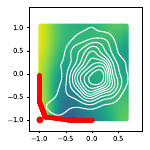} 
	& \includegraphics[draft=false,width=0.1\textwidth]{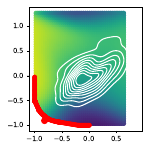}  &
	\\  \hline
	Icelandic   & 1100-1600 & 1600-1900 & 1900-2000  &\\
		\multirow{2}{*}{	\includegraphics[draft=false,width=0.2\textwidth]{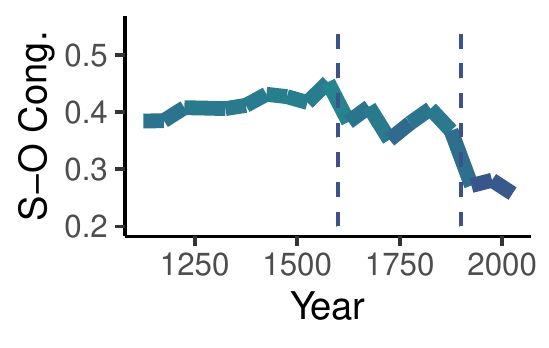}}& Before change & During change & After change &\\	
	& \includegraphics[draft=false,width=0.1\textwidth]{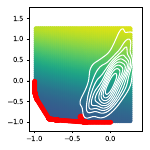}
	& \includegraphics[draft=false,width=0.1\textwidth]{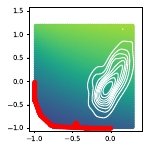}
	& \includegraphics[draft=false,width=0.1\textwidth]{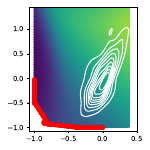} &
	\\ \hline
	Sinitic &  -300 & +2000 & +2000  &\\
			\multirow{2}{*}{		\includegraphics[draft=false,width=0.2\textwidth]{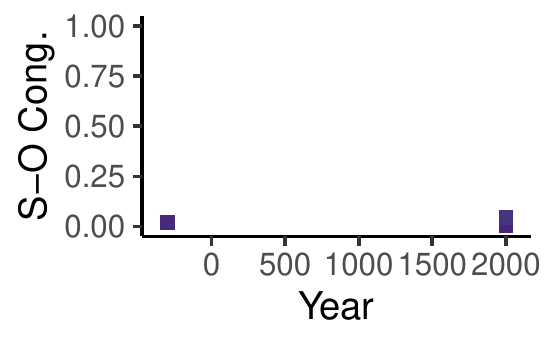}	}	& \textit{Classical Chinese} & \textit{Chinese (Mandarin)}	& \textit{Cantonese} & \\
		& \includegraphics[draft=false,width=0.1\textwidth]{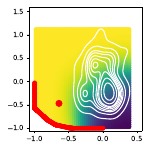}
	& \includegraphics[draft=false,width=0.1\textwidth]{process9_frontier18_Viz.py_Chinese_2.8_smoothed_scaled_noBounds.png-1.png}
	& \includegraphics[draft=false,width=0.1\textwidth]{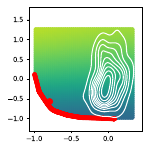} &
	\\ \hline
	Greek & -700 (poetry) & -600 -- -350 & -350 -- +200 &+2000 \\
			\multirow{2}{*}{		\includegraphics[draft=false,width=0.2\textwidth]{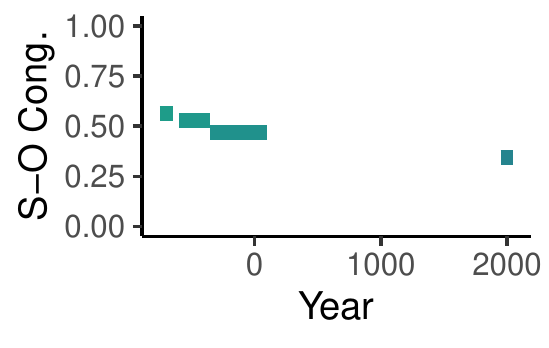}	}	& \textit{Archaic Greek}\footnote{This corpus consists of poetry (Homer and Hesiod), potentially explaining the high efficiency in IL compared both to later stages, and to typologically similar ancient Indo-European languages like Latin and Sanskrit. See also caption of Figure~\ref{fig:historical-corpora}.}  & \textit{Classical Greek} & \textit{Koine Greek} & \textit{Greek} \\
	& \includegraphics[draft=false,width=0.1\textwidth]{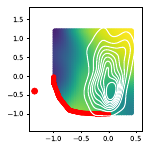}
	& \includegraphics[draft=false,width=0.1\textwidth]{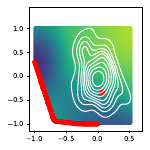}
	& \includegraphics[draft=false,width=0.1\textwidth]{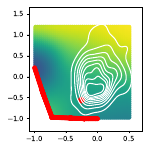}
	& \includegraphics[draft=false,width=0.1\textwidth]{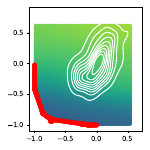}
	\\ \hline
	Hindi/Urdu & -700 & +2000 & +2000  & \\
			\multirow{2}{*}{		\includegraphics[draft=false,width=0.2\textwidth]{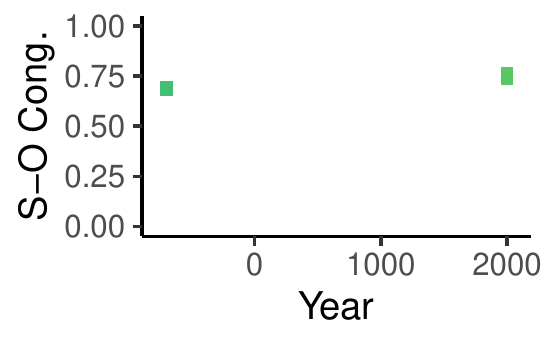}	}	& \textit{Sanskrit} & \textit{Hindi} & \textit{Urdu} & \\
	& \includegraphics[draft=false,width=0.1\textwidth]{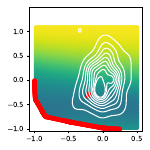}
	& \includegraphics[draft=false,width=0.1\textwidth]{process9_frontier18_Viz.py_Hindi_2.8_smoothed_scaled_noBounds.png-1.png}
	& \includegraphics[draft=false,width=0.1\textwidth]{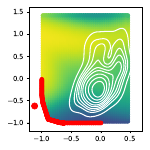}
	&
	\\ \hline
	East Slavic & +1200 & +2000 & +2000 & +2000 \\
		\multirow{2}{*}{			\includegraphics[draft=false,width=0.2\textwidth]{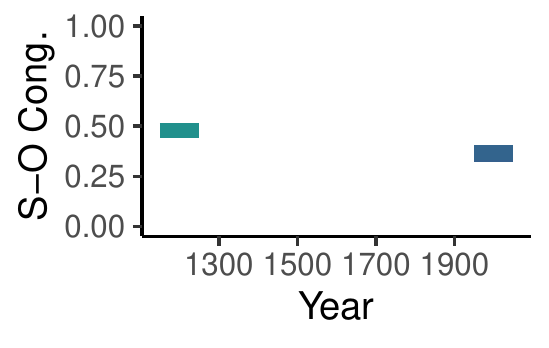}}		& \textit{Old East Slavic} & \textit{Russian} & \textit{Belarusian} & \textit{Ukrainian} \\
	& \includegraphics[draft=false,width=0.1\textwidth]{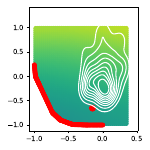}
	& \includegraphics[draft=false,width=0.1\textwidth]{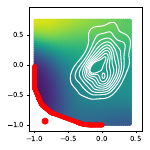}
	& \includegraphics[draft=false,width=0.1\textwidth]{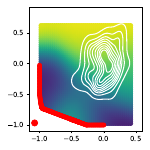}
	& \includegraphics[draft=false,width=0.1\textwidth]{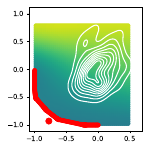}
	\\ \hline
	East South Slavic & +850 & +2000 &&\\
		\multirow{2}{*}{			\includegraphics[draft=false,width=0.2\textwidth]{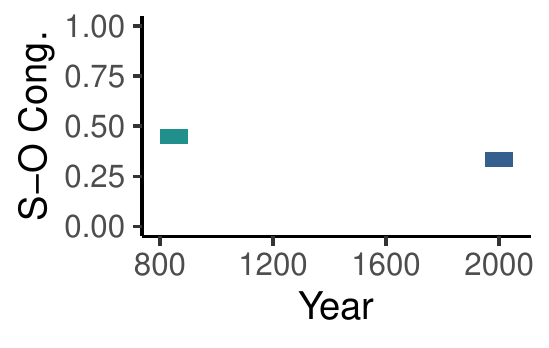}	}	& \textit{Old Church Slavonic} & \textit{Bulgarian} &&\\
	& \includegraphics[draft=false,width=0.1\textwidth]{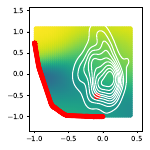}
	& \includegraphics[draft=false,width=0.1\textwidth]{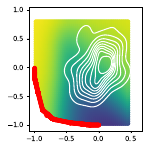} &&
\\ \hline
\end{longtable}

\newpage\section{Other Historical Datasets}\label{sec:other-historical}

\begin{figure}
	\centering
\begin{tabular}{|lll|lllll}\hline
	Old English (ISWOC, $\approx$ 900 AD) 
	& \includegraphics[draft=false,width=0.1\textwidth]{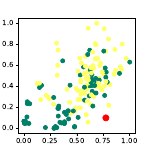} 
	& \includegraphics[draft=false,width=0.1\textwidth]{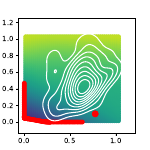} \\ \hline
	Medieval Spanish (ISWOC, $\approx$ 1400 AD) 
	& \includegraphics[draft=false,width=0.1\textwidth]{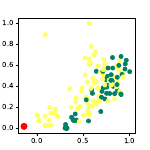} 
	& \includegraphics[draft=false,width=0.1\textwidth]{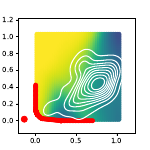} \\ \hline
	Medieval Portuguese (ISWOC, $\approx$ 1400 AD) 
	& \includegraphics[draft=false,width=0.1\textwidth]{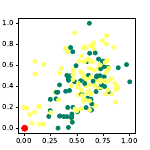} 
	& \includegraphics[draft=false,width=0.1\textwidth]{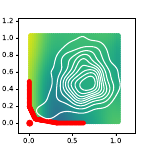} \\ \hline
%
	\hline
	Icelandic, 1100--1600 (before word order change) 
	& \includegraphics[draft=false,width=0.1\textwidth]{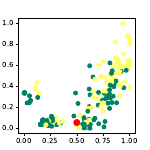} 
	& \includegraphics[draft=false,width=0.1\textwidth]{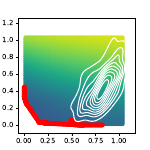} \\ \hline
	Icelandic, 1600--1900 (during word order change) 
	& \includegraphics[draft=false,width=0.1\textwidth]{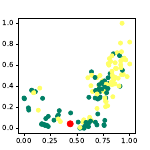} 
	& \includegraphics[draft=false,width=0.1\textwidth]{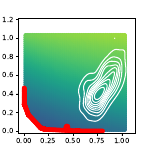} \\ \hline
	Icelandic, 1900--2020 (after word order change) 
	& \includegraphics[draft=false,width=0.1\textwidth]{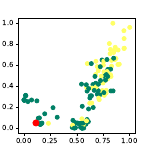} 
	& \includegraphics[draft=false,width=0.1\textwidth]{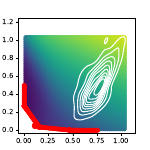} \\\hline
%
	\hline
	Archaic Greek (poetry, Homer and Hesiod, $\approx$ 700 BC) 
	& \includegraphics[draft=false,width=0.1\textwidth]{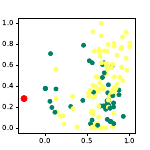} 
	& \includegraphics[draft=false,width=0.1\textwidth]{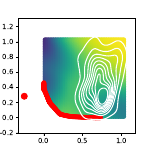} \\
	\hline
	Classical Greek (600--350 BC)
	& \includegraphics[draft=false,width=0.1\textwidth]{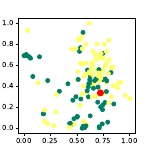} 
	& \includegraphics[draft=false,width=0.1\textwidth]{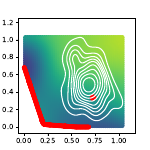} \\
	\hline
	Koine Greek (350 BC--200 AD)
	& \includegraphics[draft=false,width=0.1\textwidth]{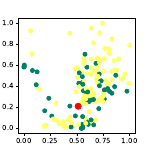} 
	& \includegraphics[draft=false,width=0.1\textwidth]{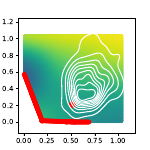} \\
	\hline
\end{tabular}
	\caption{
		Additional historical corpora, in other dependency grammar formalisms, or obtained by splitting UD treebanks into distinct epochs.
	First, we considered the treebanks in the \textbf{ISWOC collection} \citep{bech2014iswoc}, covering Old English, Medieval Spanish, and Medieval Portuguese. These corpora are annotated in a dependendency grammar format, though with differences from the Universal Dependencies formalism.
	Second, we split the \textbf{Icelandic} data, which spans almost a millenium, into three phases.
	We conducted the split based on a documented word-order change, whereby SOV was partly replaced by SVO order, between the 16th and 19th centuries \citep{Hrarsdttir2001WordOC} (see text for details).
	Finally, we split the \textbf{Ancient Greek} data, based on three conventional phases. What stands out is that Archaic Greek appears highly efficient on IL, in contrast both to later forms of Ancient Greek and related early Indo-European languages. We attribute this to the fact that the Archaic Greek subset consists entirely of poetry. The presence of meter might increase local predictability, though we leave an investigation of possible interactions of information locality and meter to future research.
}\label{fig:historical-corpora}

\end{figure}

In Figure~\ref{fig:historical-corpora}, we provide details on the historical datasets obtained from outside the Universal Dependencies Project, and by splitting UD corpora into multiple epochs. 
We plot per-dataset results as in Section~\ref{sec:per-lang}.

\paragraph{Periodization of Ancient Greek}
As shown in Figure~\ref{fig:historical-corpora}, we split Ancient Greek into three conventional phases: Archaic Greek (covering data from Homer and Hesiod, about 700BC), Classical Greek ($\approx$ 600-350 BC, represented e.g. by Herodotus and Sophocles, both $\approx$ 450BC), and Koine Greek ($\approx$ 350 BC--200 AD, represented e.g. by the New Testament and Diodorus Siculus).

\begin{figure}
	\centering
\includegraphics[draft=false,width=0.35\textwidth]{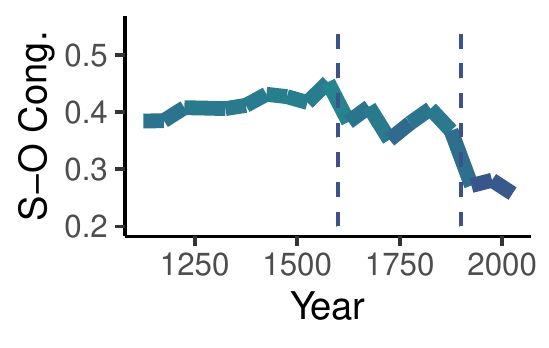} 
	\caption{Attested subject-object position congruence in Icelandic. The vertical bars denote the boundaries between the three periods for which we compute the Pareto frontier. See text (Section~\ref{sec:other-historical}) for details.}\label{sec:icelandic-trajectory}
\end{figure}

\paragraph{Periodization of Icelandic}
In Icelandic, continuous corpus data is available from the 12th century onwards \citep{arnardottir-etal-2020-universal}.
While the grammatical structure of Icelandic remained largely constant during this time, Icelandic witnessed a word order change where previously common OV order became much rarer. \citet[][p. 59]{Hrarsdttir2001WordOC} states ``\textit{OV word order seems to have been as frequent as VO word order in texts until the seventeenth century, but the frequency of OV-orders drops 30--40\% in texts dating from the seventeenth and eighteenth centuries. In the nineteenth century texts studied, the frequency of OV word order has dropped to an average of 24.8\%.}''
Figure~\ref{sec:icelandic-trajectory} shows the trajectory of attested subject-object position congruence, binning all texts in the dataset by half-centuries (e.g., 1200--1250, 1250--1300, etc.).
Subject-object position congruence appears to drop in the 17th century, and only reaches its current low level in the 20th century.
We thus grouped the data into three periods, 1100--1600, 1600--1900, 1900--today.
The first period largely coincides with the conventional period of Old Icelandic, which is conventionally taken to have ended about 1540 with the translation of the New Testament \citep{Morck2004}.

\newpage
\section{Role of Modality}

\begin{table}
	\centering
\begin{longtable}{lrrrr}
Language & Number of Sentences  \\ \hline
English SWBD & 110,504\\
French-Spoken & 2,789\\
Norwegian-NynorskLIA & 5,250\\
Slovenian-SST & 3,188\\
TuebaJS (Japanese) & 17,753\\
\end{longtable}
	\caption{Corpus sizes for spoken corpora.}\label{tab:spoken-sizes}
\end{table}

\begin{figure}
	\centering
\includegraphics[draft=false,width=0.3\textwidth]{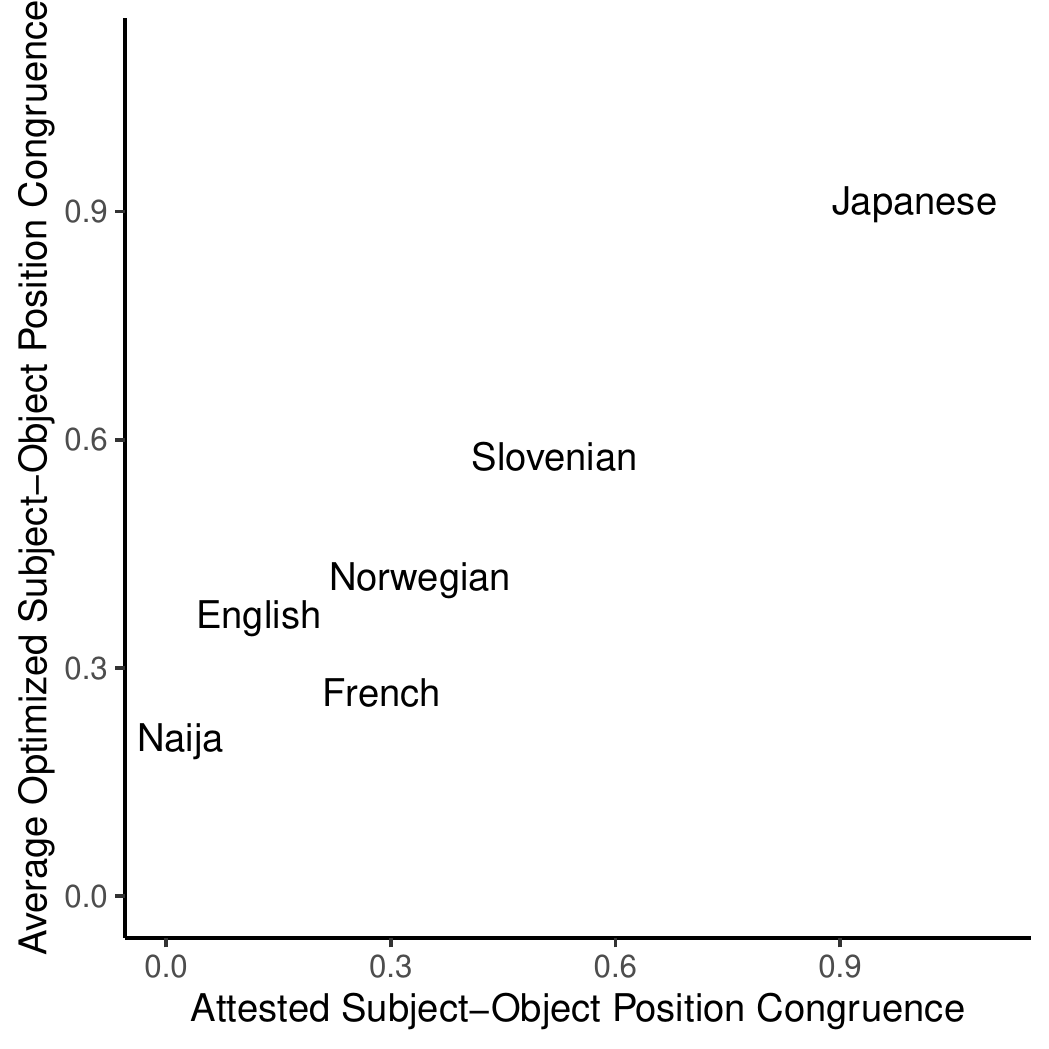} 
	\caption{Attested and average optimized subject-object position congruence for six datasets of spoken text.
	Attested and average optimized congruence are correlated ($R=0.96$, $p=0.002$; $\rho=0.94$, $p=0.02$).
	}\label{sec:congruence-spoken}
\end{figure}

Here, we consider the effect of text modality on usage patterns.
Most corpora in the Universal Dependencies collection reflect written text.
We identified six datasets of spoken languages in the Universal Dependencies format or other dependency grammar formalisms.
The Naija corpus consists entirely of spoken text.
For Slovenian, French, and Norwegian, there are sub-corpora reflecting spoken text.
We further considered the Tueba J/S corpus of spontenous dialogue in Japanese \citep{hall2006conll}
For English, we used an automated conversion \citep{schuster2018sentences} of the Switchboard section of the Penn Treebank \citep{marcus-building-1993} to Universal Dependencies.
Corpus sizes are shown in Table~\ref{tab:spoken-sizes}.

We compare attested and average optimized subject-object position congruence on these six datasets in Figure~\ref{sec:congruence-spoken}.
Results confirm that usage patterns as observed in spoken corpora also support the proposal of coadaptation between usage patterns and word order.

\newpage\section{Effects of Corpus Size}

\begin{figure}
	\centering
\includegraphics[draft=false,width=0.75\textwidth]{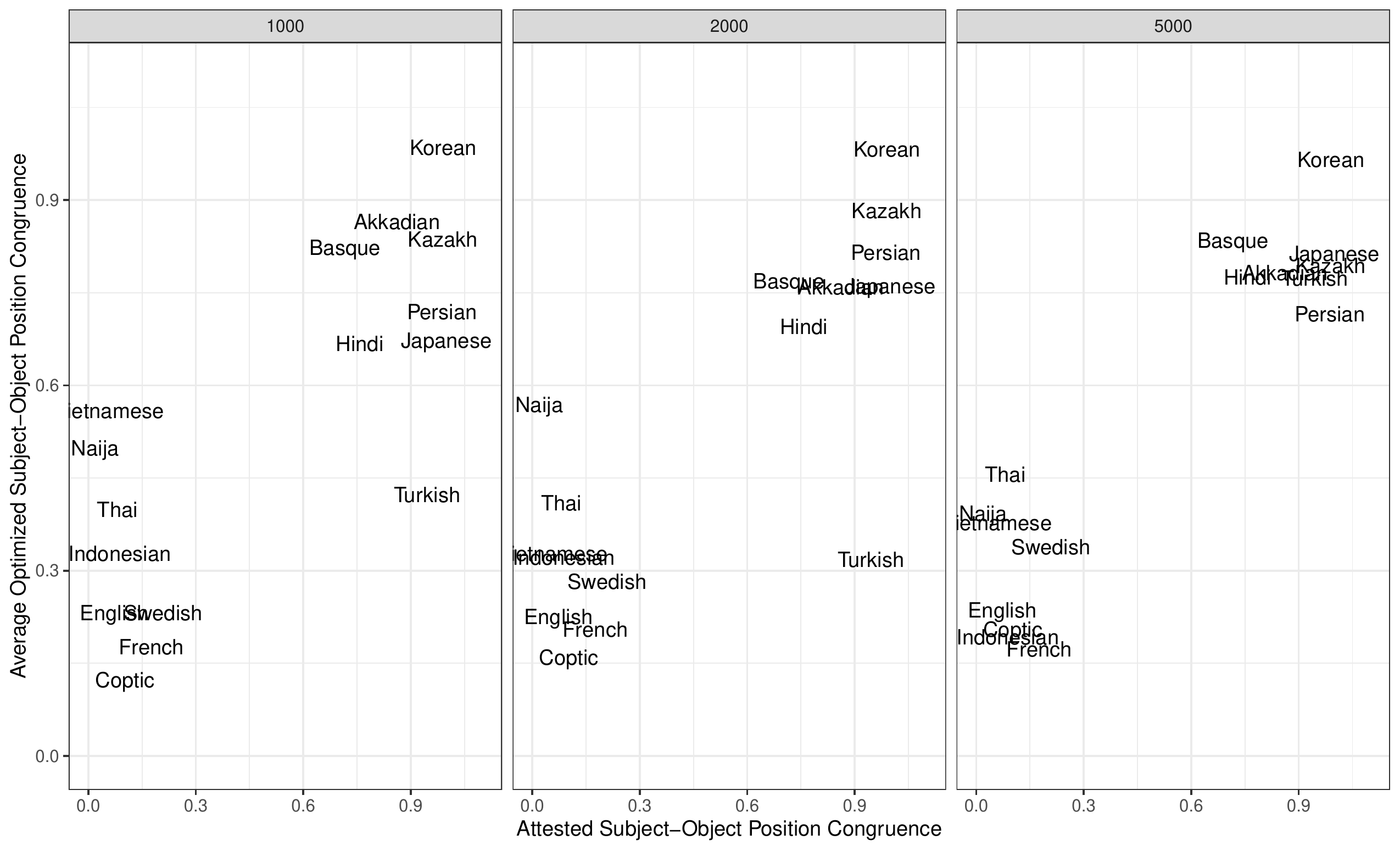} 
	\caption{Comparing attested and average optimized subject-object position congruence in subsampled subsets (1K, 2K, 5K sentences), in 16 languages.}\label{sec:congruence-size}
\end{figure}

The datasets available for different langusages differ substantially in their sizes. While some languages have tens of thousands of sentences available, corpora for other languages are substantially smaller. This raises the question whether estimates of the efficiency plane provide sufficiently reliable signal when corpora are small.
To evaluate this, we selected 8 langages with very high subject-object position congruence, and 8 languages with very low subject-object position congruence.
For each language, we randomly selected subsets of 1,000, 2,000, and of 5,000 sentences, and estimated average optimized subject-object position congruence along the Pareto frontier.
Results are shown in Figure~\ref{sec:congruence-size}.
Results suggest that estimates at 1,000 sentences may be noisier, but they are nonetheless highly correlated with estimates at 5,000 sentences ($R=0.89$, $p<0.00001$).
In order to account for errors due to finiteness of corpora, we also considered a version of the Ornstein-Uhlenbeck process incorporating measurement noise, finding that it continues to support our conclusions, and in fact estimates stronger correlations than our main analysis (Section~\ref{sec:phylo-results}).

\section{Comparison to Fitted Grammars}\label{sec:fitted}
\begin{figure}
	\centering
	\begin{tabular}{ccccccccc}
		&\ \ \ \ 		Observed Orderings & \ \ \ \ Fitted Grammars \\
		&		\includegraphics[draft=false,width=0.35\textwidth]{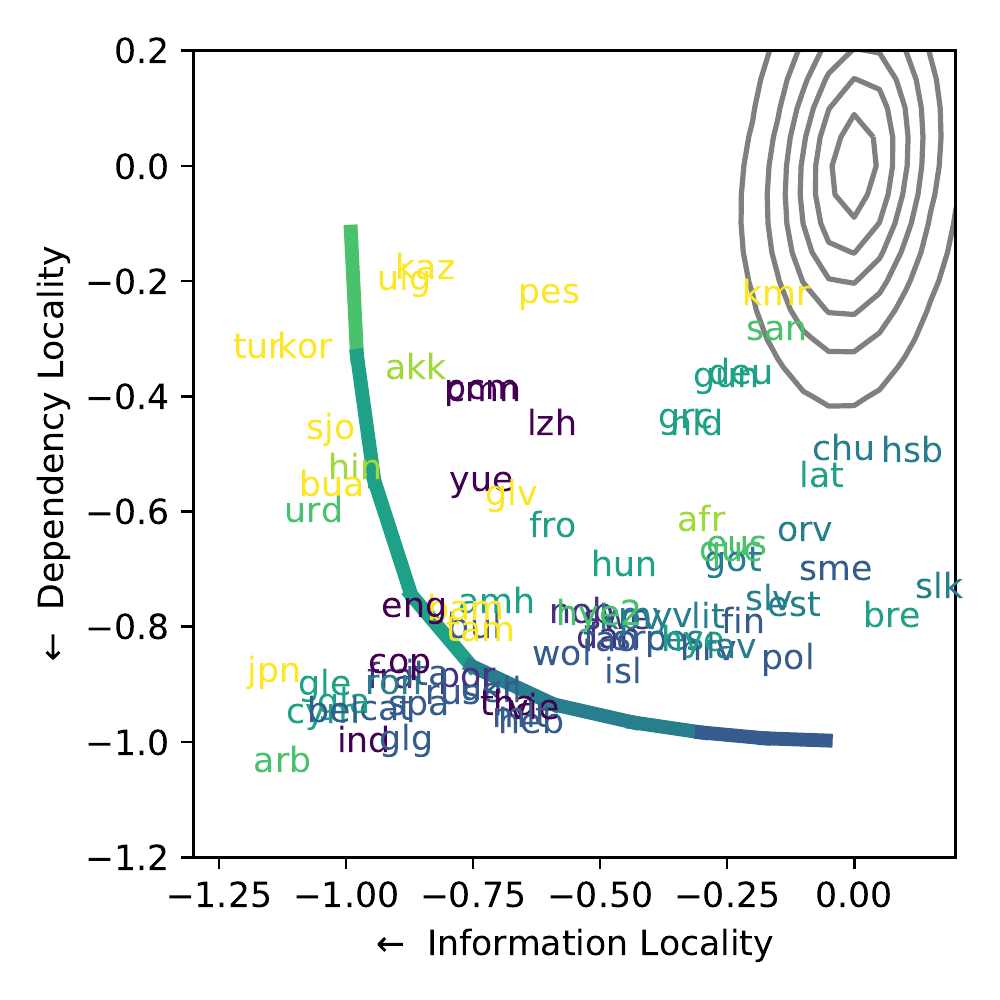} &
	\includegraphics[draft=false,width=0.33\textwidth]{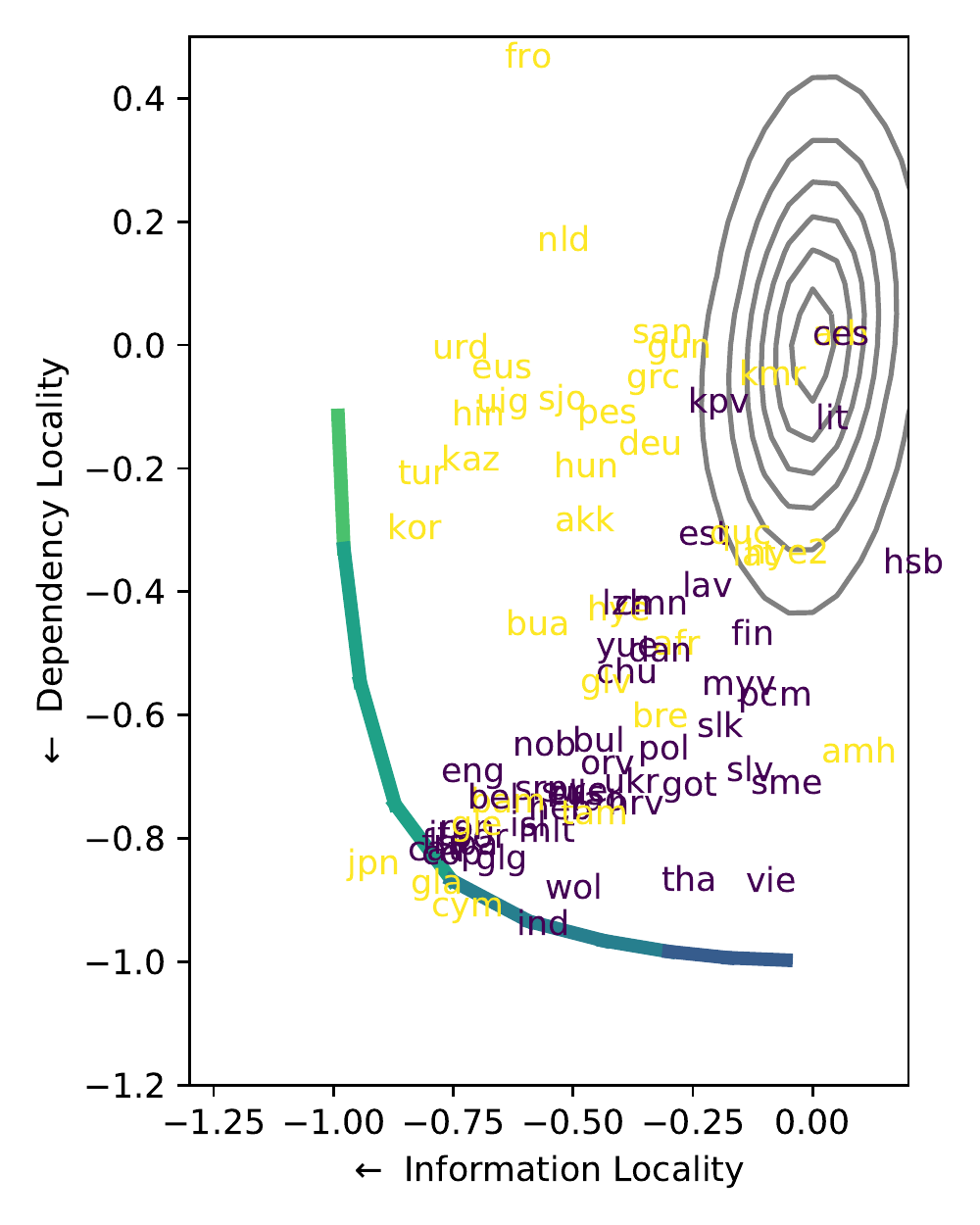} 
		\\
	\end{tabular}
	\caption{Comparing efficiency of the attested orderings with ordering grammars fitted to the attested orderings for each language. The fitted grammars are subject to exactly the same representational constraints as the baseline and approximately optimized grammars; in particular, they are a deterministic fucntion of the sentences and the syntactic relations between them. Similar to the observed orderings, fitted orderings tend to be more efficient than the baseline orderings, inhabiting the region between the baselines and the Pareto frontier. Note that, due to their design, the subject-object position congruence of fitted grammars is either 0 or 1. Observed orderings tend to be even more efficient than fitted grammars, suggesting that human languages use word order flexibility to achieve higher efficiency.}\label{sec:fitted-il-dl}
\end{figure}

Here, we compare to results obtained when representing the grammar of real languages using ordering grammars subject to the same representational constraints as the baseline and approximately optimized grammars, in order to assess the role of word order flexibility beyond the constraints of the ordering grammar formalism in efficiency optimization.
For each language, we used the hill-climbing method also used for optimizing grammars for efficiency to find a grammar which fits the observed orderings, in the sense that it maximizes the fraction of pairs of dependents of the same head that are ordered in the same order as in the attested order. 
We then evaluated these for DL and IL.
Results are shown in Figure~\ref{sec:fitted-il-dl}.

\bibliography{literature}
\bibliographystyle{plainnat}

\end{document}